\documentclass{article}

\usepackage{microtype}
\usepackage{graphicx}
\usepackage{subfigure}
\usepackage{booktabs} % for professional tables
\usepackage{hyperref}
\usepackage{enumitem}
\usepackage[numbers]{natbib}
\usepackage{algorithm}
\usepackage{algorithmic}
\usepackage{amsmath}
\usepackage{amssymb}
\usepackage{mathtools}
\usepackage{amsthm}
\newlength{\myindent}
\newcommand{\INDSTATE}[1][1]{\STATE\hspace{#1\myindent}}
\newcommand{\specialcell}[2]{#1 {\scriptsize $\pm$ #2}}
\usepackage[capitalize,noabbrev]{cleveref}

\theoremstyle{plain}

\theoremstyle{definition}

\theoremstyle{remark}

\newtheorem{hypothesis}{Hypothesis}

\usepackage[accepted]{mlsys2025}

\mlsystitlerunning{PLayer-FL: Personalized Layer-wise FL}
\begin{document}
\twocolumn[
\mlsystitle{PLayer-FL: A Principled Approach to Personalized Layer-wise Cross-Silo Federated Learning}

\begin{mlsysauthorlist}
\mlsysauthor{Ahmed Elhussein}{dbmi,nygc}
\mlsysauthor{Florent Pollet}{dbmi,nygc}
\mlsysauthor{Gamze G\"{u}rsoy}{dbmi,nygc,cs}
\end{mlsysauthorlist}

\mlsysaffiliation{dbmi}{Department of Biomedical Informatics, Columbia University, NYC, USA}
\mlsysaffiliation{nygc}{New York Genome Center, NYC, USA}
\mlsysaffiliation{cs}{Department of Computer Science, Columbia University, NYC, USA}

\mlsyscorrespondingauthor{Gamze G\"{u}rsoy}{gamze.gursoy@columbia.edu}

\mlsyskeywords{Federated Learning}

\vskip 0.3in

\begin{abstract}
Federated learning (FL) with non-IID data often degrades client performance below local training baselines. Partial FL addresses this by federating only early layers that learn transferable features, but existing methods rely on ad-hoc, architecture-specific heuristics. We first conduct a systematic analysis of layer-wise generalization dynamics in FL, revealing an early-emerging transition between generalizable (safe-to-federate) and task-specific (should-remain-local) layers. Building on this, we introduce Principled Layer-wise Federated Learning (PLayer-FL), which aims to deliver the benefits of federation more robustly. PLayer-FL computes a novel federation-sensitivity metric efficiently after a single training epoch to choose the optimal split point for a given task. Inspired by model pruning, the metric quantifies each layer’s robustness to aggregation and highlights where federation shifts from beneficial to detrimental. We show that this metric correlates strongly with established generalization measures across diverse architectures. Crucially, experiments demonstrate that PLayer-FL achieves consistently competitive performance across a wide range of tasks while distributing gains more equitably and reducing client-side regressions relative to baselines.
\end{abstract}
]


\section*{Software and data} 
\vspace{-1mm}
Code for reproducing this paper can be found at \url{https://github.com/G2Lab/PLayer-FL}.

\section{Introduction}
Federated Learning (FL) is a distributed learning framework that enables collaborative model training without data-sharing. Early FL algorithms such as Federated Averaging (FedAvg) focused on developing a unified global model \cite{mcmahan2017communication}. However, when data among clients is non-independently identically distributed (non-IID), the performance of a global model diminishes. This largely stems from the fact that clients, each optimizing their local models using non-IID datasets, often generate model updates that diverge from each other \cite{zhao2018federated}. A weighted averaging of these updates then causes the global model to settle outside of local minima, which reduces the performance. To address this, personalized FL methods have been developed that allow client models to diverge, typically by guiding the training of a local model with a global model \cite{smith2017federated, dinh2021fedu, t2020pfedme}. In certain non-IID settings, Personalized FL algorithms have improved performance, yet they do not consistently outperform FedAvg. Moreover, their benefits are not always equitably distributed among all participating clients \cite{li2022federated, divi2021new}. That is, only a subset of clients might experience increased performance compared to a model trained only on their local data.  

Partial FL, in which only a subset of layers are federated, is a form of personalized FL \citep{arivazhagan2019federated, oh2021fedbabu,collins2021exploiting}. Partial FL is based on an implicit assumption that the greatest divergence in model weights primarily occurs in the model's final layers. By federating early layers and only locally updating later layers, partial FL retains the benefits of FL even with non-IID data. This approach is informed by insights from representation learning regarding layer function and generalizability. It is well-established that early layers in neural networks typically capture universal features that are more generalizable across tasks \cite{yosinski2014transferable, simonyan2013deep, zeiler2014visualizing, mallat2016understanding}. In contrast, later layers are often more task-specific, thus tend to benefit from localized training or fine-tuning \cite{yosinski2014transferable}. This transition in layer function and generalizability is also observed in the model loss landscape. Early layers, which are linked to feature representation, generally occupy flatter areas of the loss function, suggesting a lower sensitivity to perturbations. \cite{chaudhari2019entropy, jiang2019fantastic, yosinski2014transferable}. However, existing partial FL methods \citep{arivazhagan2019federated, oh2021fedbabu,collins2021exploiting} suffer from two fundamental gaps: first, it has not been systematically established that the layer-wise generalization dynamics that motivate partial sharing actually hold under federated, non-IID training; second, these methods select which layers to federate via architecture-specific heuristics, which limits generalizability across architectures and tasks.

Here, we propose that insights from model pruning can be leveraged to establish a more principled and protective approach to federation assignment in partial FL. Model pruning identifies and removes parameters with minimal negative impact on performance, effectively reducing model complexity without sacrificing accuracy. Similar to representation learning, model pruning offers insights into parameter and layer function that can be leveraged in partial FL. While model pruning involves identifying parameters that can be removed with minimal performance loss -- a different focus from that of partial FL -- the underlying principles still hold relevance. Studies have shown that layer-specific pruning can achieve high model performance despite significant reductions in model size \cite{dong2017learning, chen2018shallowing,fan2021layer}. This process highlights that certain parameters are critical for the model’s output and must be retained, while others may be redundant \cite{liu2018rethinking}. Further, the degree of pruning in a layer often correlates with its feature representation \cite{chen2018shallowing}. Crucially, pruning methods often use computationally simple, architecture-agnostic metrics, such as first-order approximations of parameter importance, to quantify this sensitivity.

We posit that a similar approach can identify layers in FL that are robust to aggregation across non-IID datasets. Building on this, we propose Principled Layer-wise-FL (PLayer-FL), a partial FL approach using a novel federation sensitivity metric inspired by these pruning principles. This metric employs a first-order approximation to quantify each layer's robustness to aggregation after just one training epoch, identifying layers suitable for federation while minimizing the risk of negative transfer. Unlike heuristic methods, this allows PLayer-FL to adapt the federation strategy based on early training dynamics across diverse architectures. We validate our approach across various model architectures and real-world non-IID datasets, demonstrating not only consistent performance improvements but also the fairness and incentivization guarantees essential for robust, sustainable collaboration.

\section{Contribution and significance}
In this study, we make several contributions toward robust, equitable FL with non-IID data:
\begin{itemize}[leftmargin=*]
    \item \textbf{Early identification of layer-wise generalization patterns federated training.} We show that models with \emph{identical initialization but independent training on non-IID data} exhibit consistent generalization patterns across corresponding layers, even after just one epoch. This enables early, reliable identification of which layers benefit from federation versus which would suffer from negative transfer, minimizing exposure to harmful training dynamics.
    \item \textbf{Federation sensitivity: A efficient metric for robust collaboration.} We introduce federation sensitivity, a novel metric that estimates each layer's robustness to aggregation and \emph{identifies a transition point where federation shifts from beneficial to potentially harmful}. This metric is architecture-agnostic, correlates strongly with established generalization measures, emerges from the first epoch, and imposes minimal computational overhead -- enabling principled deployment without requiring extensive tuning.
    \item \textbf{PLayer-FL: Principled partial FL.} We develop PLayer-FL, which uses federation sensitivity to determine which layers to federate after only one epoch of training. PLayer-FL achieves \emph{stable and competitive} performance across diverse non-IID datasets: it achieves the best performance, fairness and incentivization rankings across benchmarks, ensuring equitable benefits and protecting all clients from performance degradation relative to local training.
\end{itemize}

\section{Background}
\subsection{Federated Learning}
In each training round $k$, the server broadcasts global model parameters $\Theta^{k-1}$ to clients $c \in C$. These clients update their local model to produce $\Theta_c^k$ which are sent back to the server. The central server then aggregates these parameters and rebroadcasts the updated global model. This is repeated until convergence. In FedAvg, the aggregation is defined as $\Theta^k = \sum_{c=1}^n \alpha_c{\Theta_c^k}$, where $\alpha_c$ is the weight of each client determined by sample size. 

\subsubsection{Cross-Silo FL}
\label{background:cross_fl}
FL has two main settings, cross-silo and cross-device, each with distinct use cases and challenges \cite{kairouz2021advances}. This work concentrates on cross-silo FL, a setting common in sectors such as healthcare to overcome data-sharing constraints. In healthcare, cross-silo FL has advanced research on poorly understood diseases by enabling training on larger datasets \cite{pati2022federated, dayan2021federated}. Cross-silo FL involves a few clients with medium to large datasets. A key challenge is non-IID data as institutions often have different populations, data collection practices, and operational protocols \cite{pati2021federated, terrail2022flamby}. Further, since clients frequently possess enough data to train adequately performant models on their own, FL must surpass local training to motivate participation \cite{huang2022cross}. In contrast, cross-device FL manages numerous edge devices with limited data, prioritizing communication efficiency. While most FL methods target cross-device scenarios, fewer address cross-silo FL's unique challenges \cite{huang2022cross, terrail2022flamby}, highlighting the need for specialized approaches.

\subsection{Model pruning}
Model pruning reduces the size of a model by removing parameters that have limited impact on the output \cite{reed1993pruning}. Simple, yet effective metrics include weight magnitude or the scaling parameter in batch normalization \cite{han2015learning, liu2017learning}. Other methods leverage a first or second-order approximation of the change in loss following the removal of parameter $w_s$ to quantify its importance, denoted as $\mathcal{I}_s(w)$  \cite{lecun1989optimal, molchanov2019importance}. The importance of a parameter set, $W_\mathcal{S}$, where $w_s \in W_\mathcal{S}$, is defined as \cite{molchanov2019importance}:
\begin{equation}
\label{eqn:first_order_approx}
    \mathcal{I}_\mathcal{S}(W) \triangleq \sum_{s\in \mathcal{S}} \mathcal{I}_s(w)= \sum_{s \in \mathcal{S}} (g_sw_s) ^2
\end{equation}
where $g_s$ is the gradient with respect to the parameter $w_s$. Importantly, many of these simple methods have been shown to be reliably effective across diverse model architectures. 

\subsection{Model generalization}
Many studies in representation learning have sought to understand when and how models generalize. One method is to examine the loss landscape. Parameters located in flatter regions of the loss landscape exhibit reduced sensitivity to perturbations and are more generalizable. Multiple studies have also made the link that generalizable layers identify low-level features that are common across tasks \cite{yosinski2014transferable}. To estimate the characteristics of the loss landscape, first and second-order gradient approximations are used.  \citet{jiang2019fantastic} showed that gradient variance of a layer, even after one epoch, correlates with its generalizability. \citet{chaudhari2019entropy} used the eigenspectrum of the Hessian, which reflects the steepness of the loss function, as a measure of generalizability. They showed that biasing model optimization towards wide valleys improves a model's sensitivity to perturbation.  

Another approach to measuring model generalization is comparing representational similarity across models \cite{li2015convergent, morcos2018insights}. Typically, the same model is randomly initialized and trained on the same dataset a number of times. The representation of a set of samples after each layer is then compared across all models \cite{kornblith2019similarity,  williams2021generalized}. One key insight is that generalizing networks create similar representations across samples, while overfit, memorizing networks, trained on random labels, do not. \cite{morcos2018insights}. 

\textbf{Relation to our work:} We apply these techniques to address a related but distinct question more relevant for FL:

\emph{To what extent do identically initialized models trained on non-IID data resemble each other?} 

In other words, how much model divergence do we observe in FL and can this be reliably estimated? A natural question to ask is why not directly compare the weights. However, this approach has several limitations. First, there is not a straightforward, effective metric for capturing weight divergence. Common metrics such as L2-norm and cosine similarity are not invariant to linear transformation and may lack the sensitivity to detect subtle, yet significant differences in weights. Second, unlike gradients, differences in weights may not always emerge early in training. Early emergence helps to limit computational cost and mitigates the negative impacts of model training on non-IID data.
 
\section{Related works}
\textbf{Global FL} aims to generate a global model for all clients. In FedAvg, a global model is updated using a weighted average of local updates from each client \cite{mcmahan2017communication}. While this simple procedure has achieved remarkable success, it struggles when clients have non-IID data \cite{li2020federated}. Some attempts to solve this problem while retaining a global model include sharing a subset of public data \cite{zhao2018federated}, augmenting the data to be more IID \cite{duan2019astraea}, and adding a regularization term to keep local updates close to the global model \cite{li2020federated}. While these methods mitigate some of the loss in performance, they still only train a single global model that is not optimized for each individual client.

\textbf{Personalized FL} aims to learn a model for each client that is informed by a global model. One approach is to cluster similar clients together and train a separate model for each cluster \cite{briggs2020federated, mansour2020three}. However, identifying clusters can be challenging. Another related approach is multi-task learning, where clients with similar data can share model parameters more extensively \cite{smith2017federated, dinh2021fedu}. There are also regularization-based approaches that train a local model but adapt the loss function to ensure updates remain close to the global model \cite{li2021ditto, t2020pfedme}. Additional hyperparameters add complexity to the training process. In local adaptation, clients fine-tune the global model locally after convergence \cite{yu2020salvaging}. This works well when the global model is already near the optimal local model for each client which may not hold with non-IID data.

\textbf{Partial FL} is a type of personalized FL that selectively federates subsets of the model while leaving the remainder to train locally. In practice, early layers are usually federated and later layers retained for local adaptation. FedPer trains the full model locally but only federates the early layers \cite{arivazhagan2019federated}; FedBABU trains and federates early layers while keeping the final layers fixed until convergence \cite{oh2021fedbabu}; and FedRep decouples a shared body from a local head, jointly learning a body for low-dimensional representations while reserving the head for local training \cite{collins2021exploiting}. However, two gaps remain: (1) it has not been systematically established that the layer-wise generalization dynamics motivating these designs actually hold under federated, non-IID training; and (2) the choice of which layers to federate is typically pre-determined or based on architecture-specific heuristics, which limits generalizability beyond CNNs. Other methods such as FedLP and FedLAMA select layers dynamically during training \cite{zhu2023fedlp,lee2023layer}, but they primarily target communication efficiency and aim to match FedAvg performance rather than address non-IID data. pFedLA proposes personalized layer-wise aggregation via server-side hypernetworks that learn client- and layer-specific aggregation weights \cite{ma2022layer}, but this adds considerable complexity. Finally, pFedHR reassembles models on the server after grouping functionally related layers \cite{wang2024towards}, yet it requires sharing data, which is often infeasible in cross-silo FL.

\section{Layer-wise generalization dynamics in Federated Learning}
\label{sec:gen_sim}

A fundamental challenge in partial FL is determining which layers benefit from federation versus which would be harmed by aggregation across non-IID data. To address this, we investigate whether identically initialized models trained on non-IID data exhibit predictable, layer-specific patterns in their generalization capabilities.

We examine two training scenarios: (1) independent training, where each client trains completely locally (representing maximum divergence), and (2) federated training, where models undergo periodic aggregation (representing typical FL divergence). By comparing these scenarios, we can identify the extent to which federation either mitigates or exacerbates client divergence at different layers. Figures \ref{fig:grad_var}--\ref{fig:layer_imp} present results from the first epoch. Figures \ref{sfig:grad_var_best}--\ref{sfig:layer_imp_best_fl} present final model results, demonstrating that the layer-wise generalization patterns observed early in training persist throughout the entire training process. Definitions for all metrics are provided in Section \ref{metric_defn}.

\textbf{Key finding.} Prior work shows that randomly initialized models on IID data follow similar training trajectories \cite{wang2018towards}. We study the complementary setting: identically initialized models trained on non-IID client subsets. A layer-wise analysis reveals a stable pattern across architectures and datasets: \textbf{early layers converge to similar, broadly transferable solutions despite non-IID data, whereas later layers diverge markedly}. This echoes transfer-learning behavior but is established here specifically for FL.

\textbf{Identifying a transition point.} Crucially, we observe a consistent \emph{transition point} in each architecture, a boundary that separates layers functioning as generalizable feature extractors (suitable for federation) from layers serving as task-specific feature utilizers (requiring local training). This transition marks where the benefits of federation shift to potential harm through negative transfer. Remarkably, this transition point emerges after just one epoch and remains stable throughout training, enabling early, reliable decisions about which layers to federate without prolonged exposure to harmful aggregation effects.

This empirical finding motivates our federation sensitivity metric: if layer-wise generalization patterns are consistent and emerge early, we can develop a principled, data-driven approach to identify robust collaboration boundaries for each task rather than relying on architecture-specific heuristics.

\subsection{Layer generalization}
\begin{figure*}[ht!]
\begin{center}
\makebox[\linewidth][c]{  % Center the four subfigures in a line
\begin{minipage}{.22\linewidth}  % Adjust the width to 25% for each subfigure
  \centering
  \raisebox{-\height}{\includegraphics[width=\linewidth]{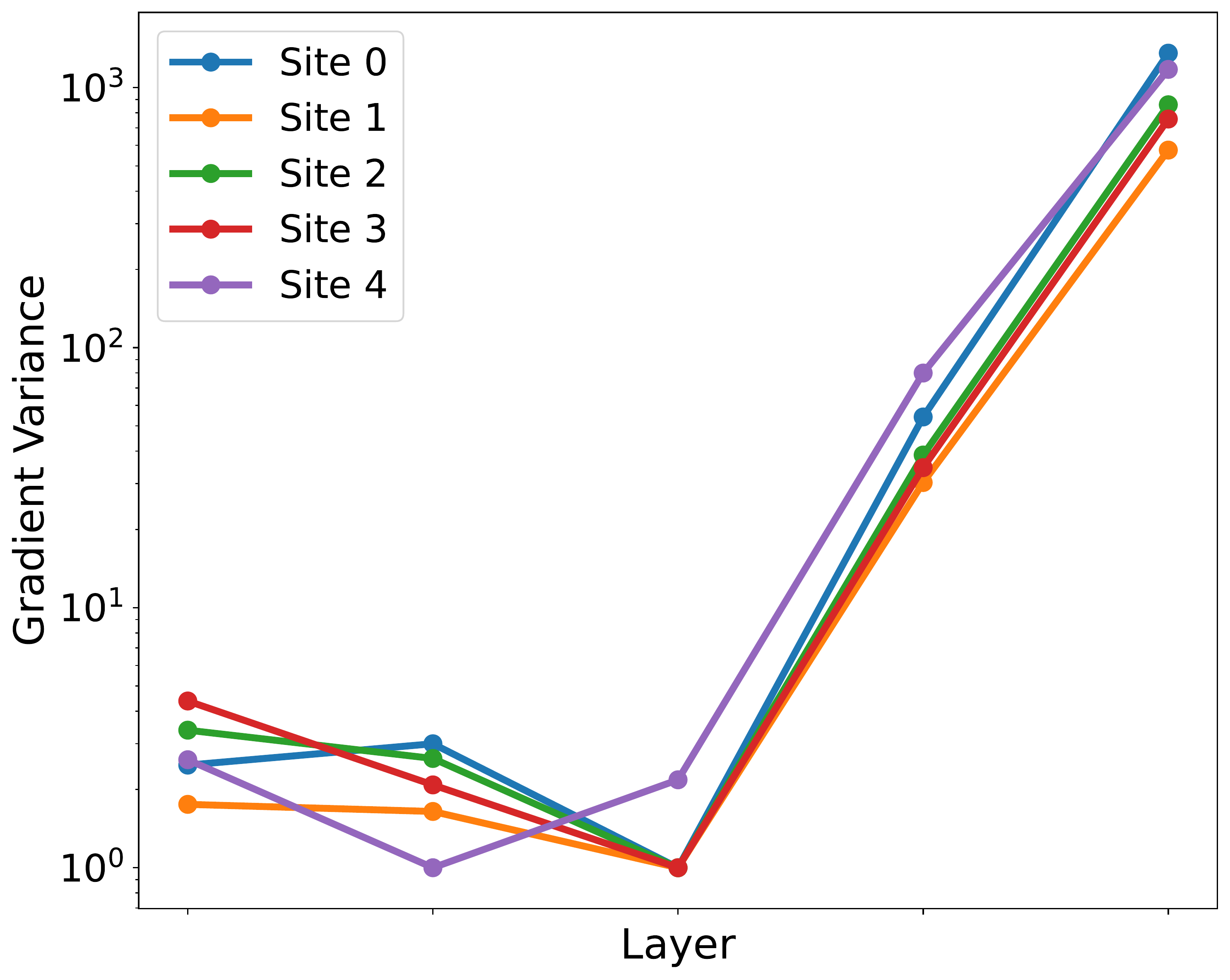}}
  \makebox[\linewidth][c]{\scriptsize\textit{(A) FashionMNIST}}  % Subfigure label
\end{minipage}%
\begin{minipage}{.22\linewidth}  % Adjust the width to 25% for each subfigure
  \centering
  \raisebox{-\height}{\includegraphics[width=\linewidth]{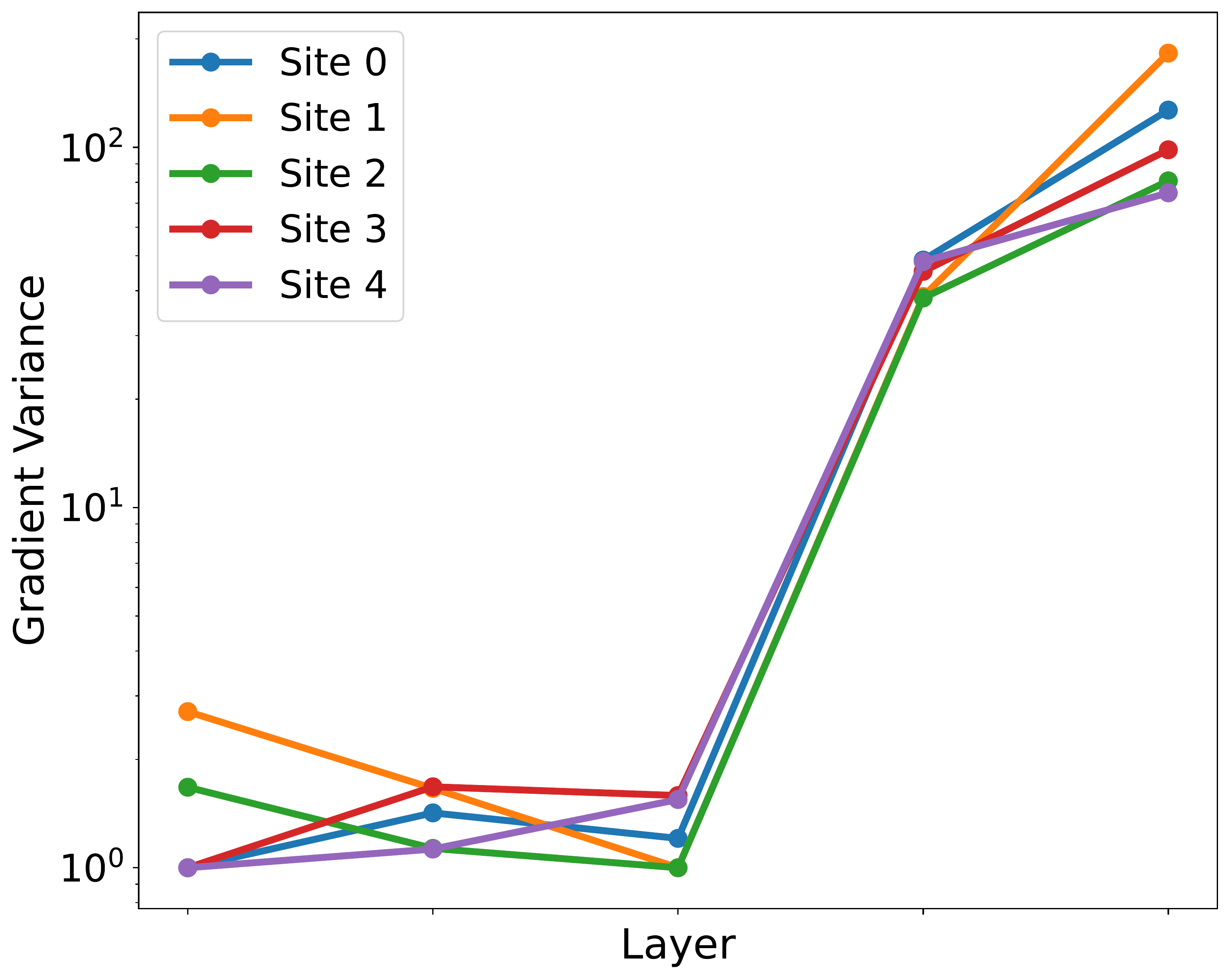}}
  \makebox[\linewidth][c]{\scriptsize\textit{(B) EMNIST}}  % Subfigure label
\end{minipage}%
\begin{minipage}{.22\linewidth}  % Adjust the width to 25% for each subfigure
  \centering
  \raisebox{-\height}{\includegraphics[width=\linewidth]{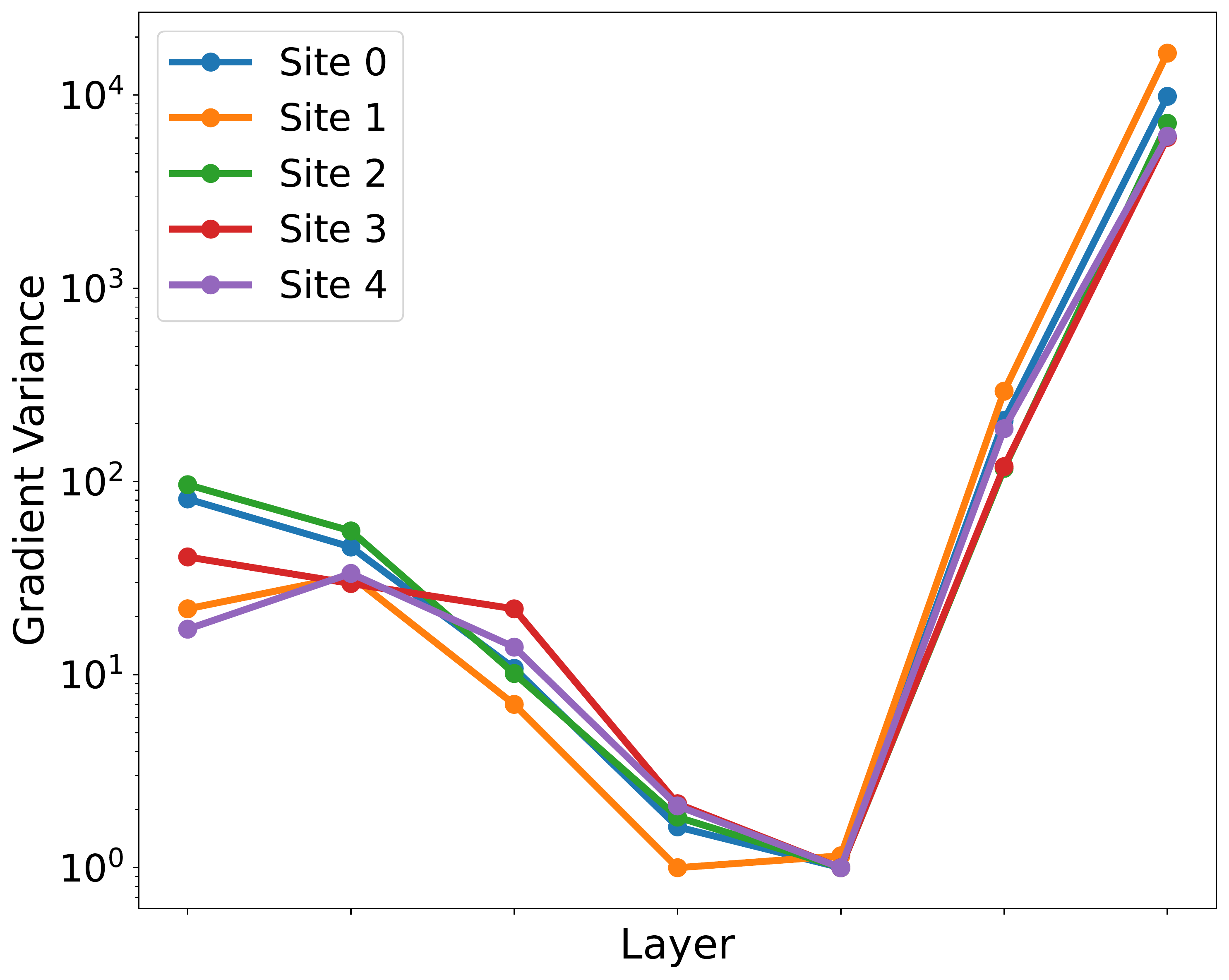}}
  \makebox[\linewidth][c]{\scriptsize\textit{(C) CIFAR-10}}  % Subfigure label
\end{minipage}%
\begin{minipage}{.22\linewidth}  % Adjust the width to 25% for each subfigure
  \centering
  \raisebox{-\height}{\includegraphics[width=\linewidth]{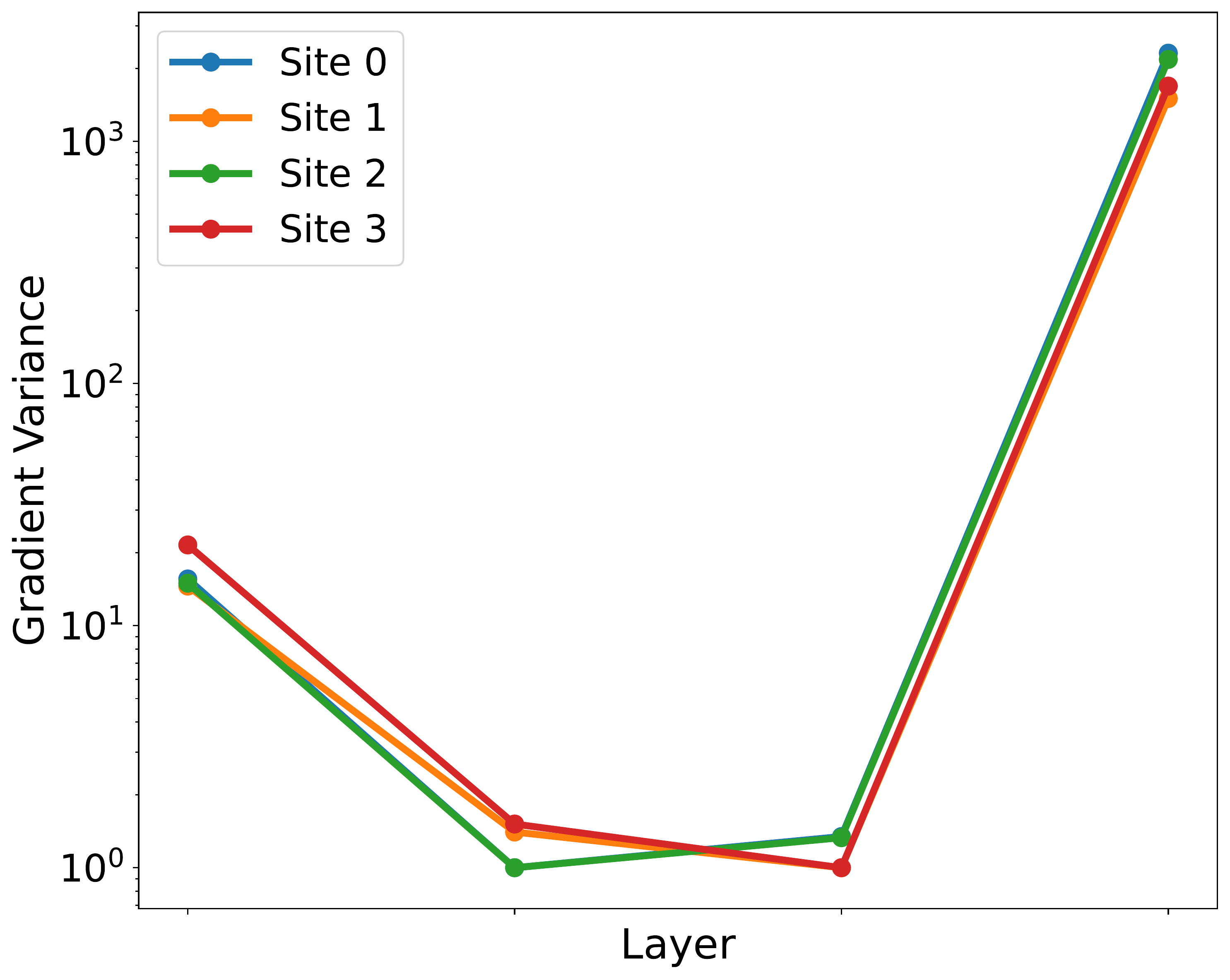}}
  \makebox[\linewidth][c]{\scriptsize\textit{(D) MIMIC-III}}  % Subfigure label
\end{minipage}
}
\caption{\textbf{Layer gradient variance after one epoch}. All models identically initialized and independently trained on non-IID data.}
\label{fig:grad_var}
\end{center}

\end{figure*}

\begin{figure*}[ht]
\begin{center}
\makebox[\linewidth][c]{  % Center the four subfigures in a line
\begin{minipage}{.22\linewidth}  % Adjust the width to 25% for each subfigure
  \centering
  \raisebox{-\height}{\includegraphics[width=\linewidth]{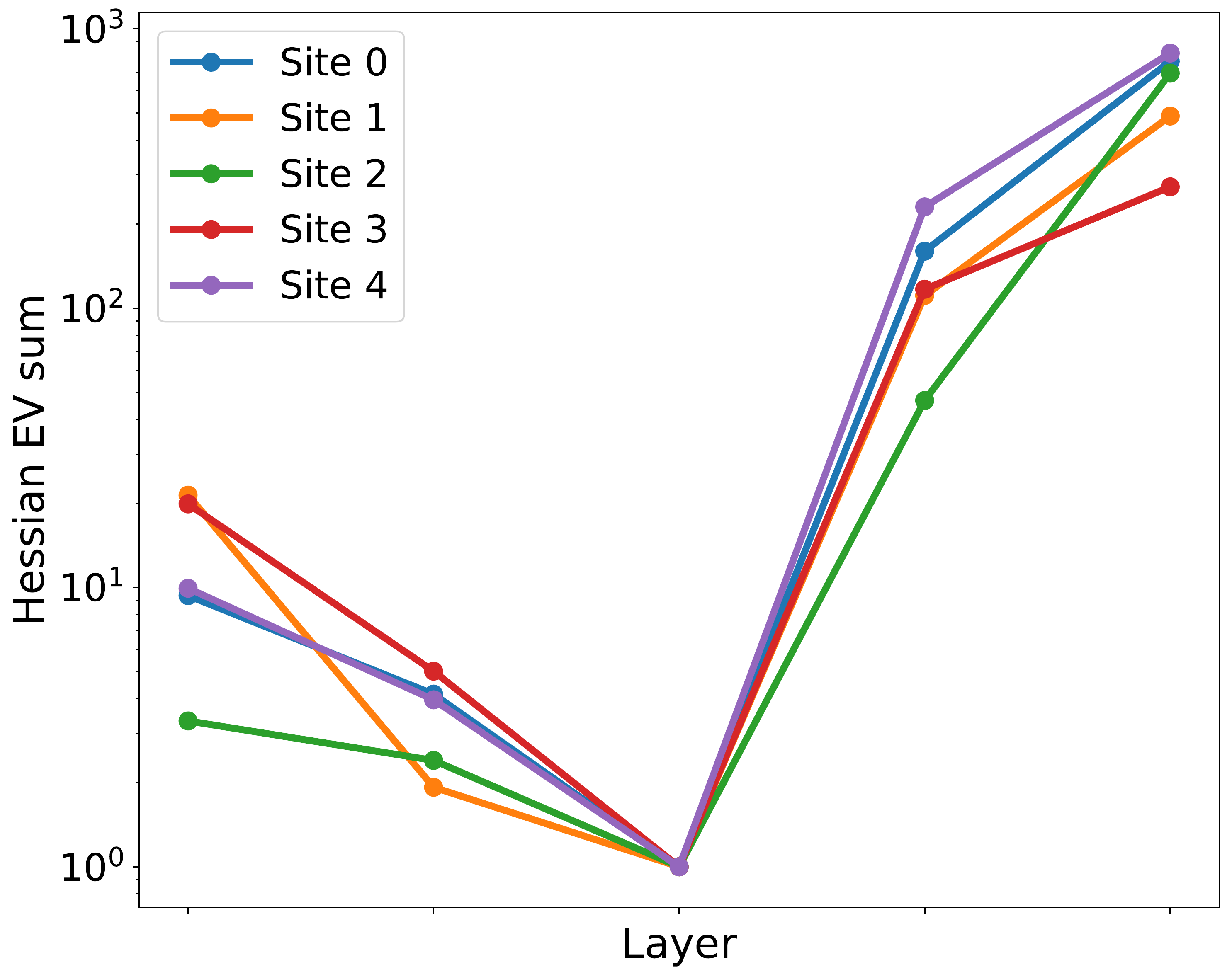}}
  \makebox[\linewidth][c]{\scriptsize\textit{(A) FashionMNIST}}  % Subfigure label
\end{minipage}%
\begin{minipage}{.22\linewidth}  % Adjust the width to 25% for each subfigure
  \centering
  \raisebox{-\height}{\includegraphics[width=\linewidth]{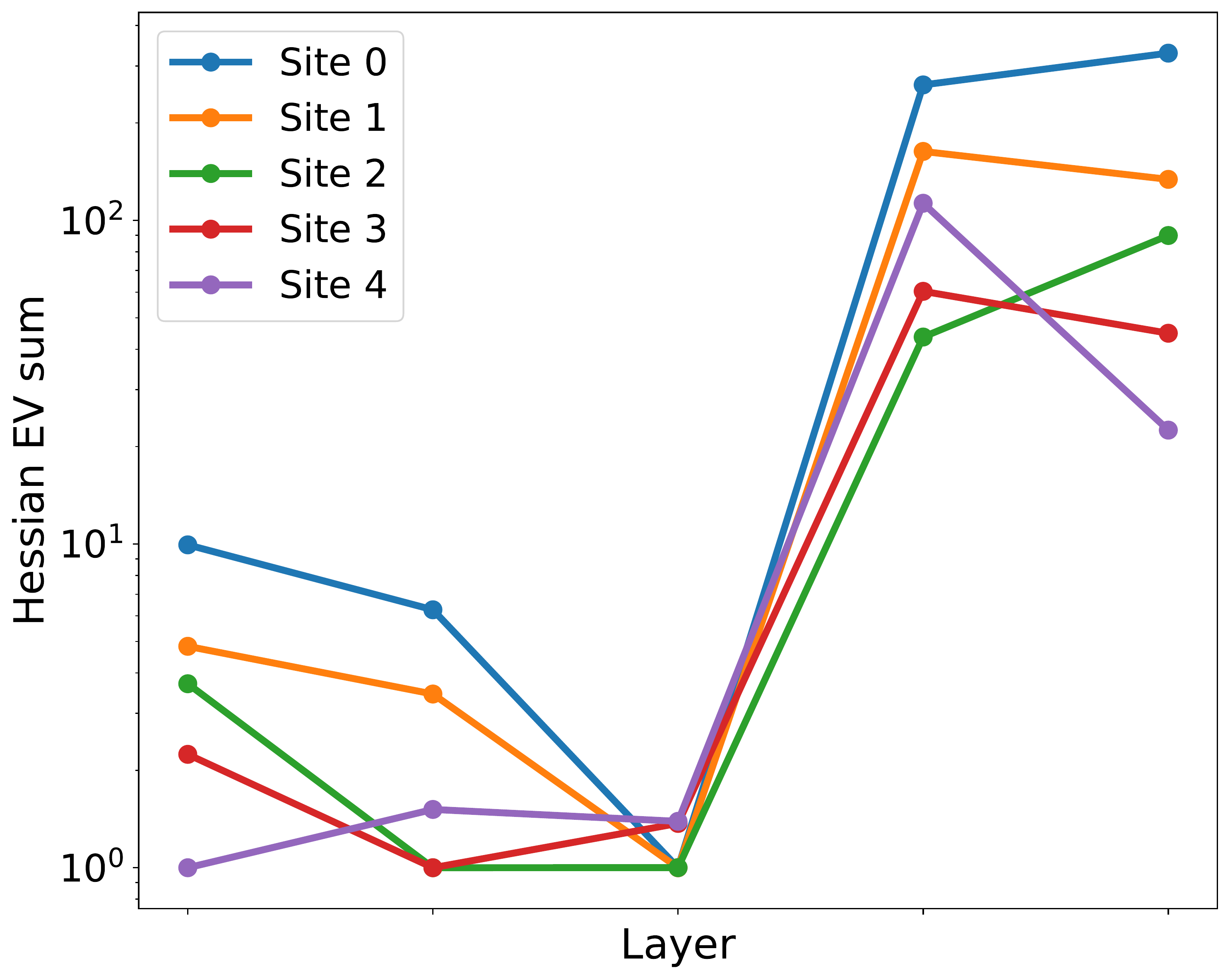}}
  \makebox[\linewidth][c]{\scriptsize\textit{(B) EMNIST}}  % Subfigure label
\end{minipage}%
\begin{minipage}{.22\linewidth}  % Adjust the width to 25% for each subfigure
  \centering
  \raisebox{-\height}{\includegraphics[width=\linewidth]{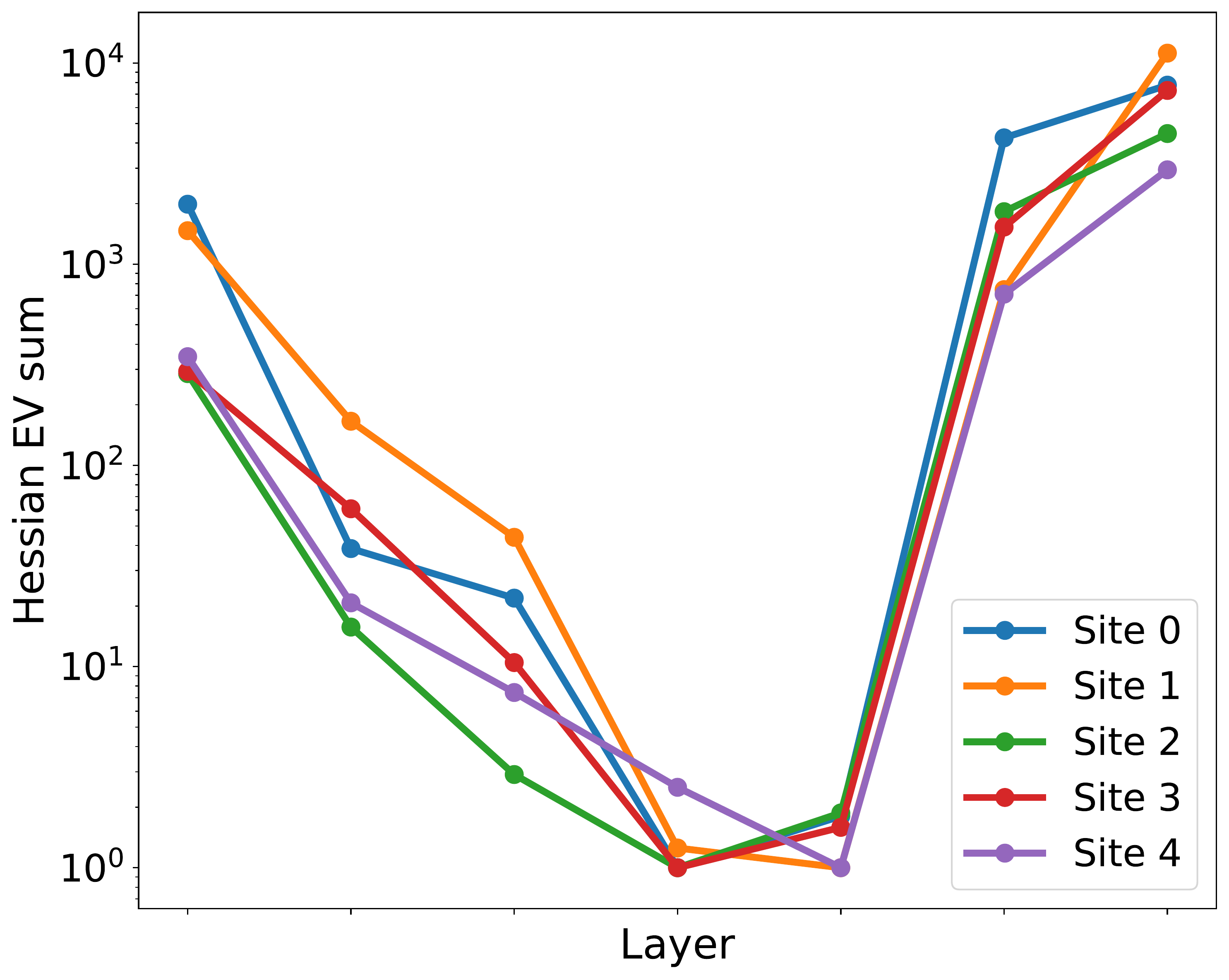}}
  \makebox[\linewidth][c]{\scriptsize\textit{(C) CIFAR-10}}  % Subfigure label
\end{minipage}%
\begin{minipage}{.22\linewidth}  % Adjust the width to 25% for each subfigure
  \centering
  \raisebox{-\height}{\includegraphics[width=\linewidth]{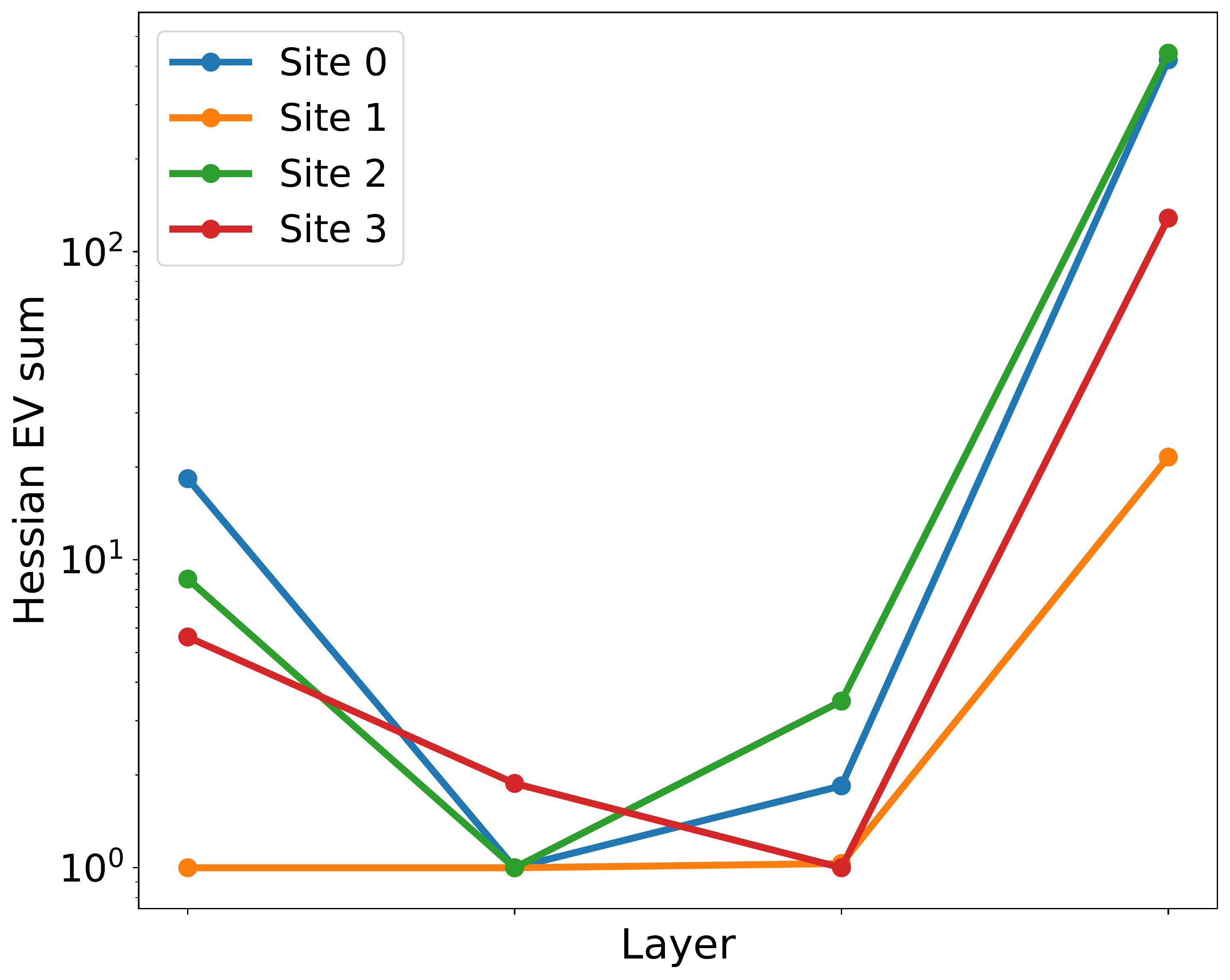}}
  \makebox[\linewidth][c]{\scriptsize\textit{(D) MIMIC-III}}  % Subfigure label
\end{minipage}
}
\caption{\textbf{Hessian eigenvalue sum after one epoch}. Each model was identically initialized and trained on respective non-IID datasets.}
\label{fig:hess_ev_sum}
\end{center}

\end{figure*}

\textbf{Early layers are robust to perturbation -- a protective property for federation.}
Figures~\ref{fig:grad_var} and \ref{fig:hess_ev_sum} report the gradient variance (Eq.~\ref{eqn:grad_var}) and the sum of Hessian eigenvalues (Eq.~\ref{eqn:hess_ev}) after one epoch across four datasets (note that independent and FL training are equivalent after one epoch, prior to any aggregation). To account for layer size, we normalize both metrics by the number of parameters in each layer. Extended analyses at the end of independent and FL training are given in Figures~\ref{sfig:grad_var_first}--\ref{sfig:hess_eig_best_fl}.

A consistent pattern emerges: both metrics rise sharply in the final layers (see log scale), indicating that early layers lie in flatter regions of the loss landscape while later layers reside in sharper, more sensitive regions. This has direct implications for robust federation: layers in flat regions have lower sensitivity to perturbation and greater generalization capacity, whereas layers in sharp regions are more vulnerable and task-specific.

\textbf{Identifying layers for robust collaboration.}
Interpreting federated averaging as a perturbation to locally optimized weights reveals a \emph{transition point}. Early layers remain in locally flat basins even after aggregation, so federating them should not harm client performance. By contrast, later layers exhibit high perturbation sensitivity; aggregating them across non-IID clients risks disrupting task-specific optima and degrading performance.

This interpretation rests on a mild assumption: that, for early layers, the global and local optima are proximate -- \emph{i.e.,} aggregation does not induce a parameter shift large enough to exit the local basin of the loss. Under this condition, averaging early-layer weights is a small perturbation within a flat region and should, at minimum, avoid worsening performance. For later layers, even small perturbations can be detrimental due to sharper minima. We empirically validate the proximity assumption via representational similarity analysis in the next section, showing that early layers converge to similar solutions across clients despite non-IID data.

\vspace{2mm}

\subsection{Validating the transition point: Representational similarity analysis}

Having established that early layers occupy flat loss regions, we now validate the assumption: \textbf{do early layers actually converge to similar solutions across clients despite non-IID training?} If true, this confirms that federating early layers should at least not harm performance.

\begin{figure*}[ht]
\begin{center}
\makebox[\linewidth][c]{  % Center the four subfigures in a line
\begin{minipage}{.22\linewidth}  % Adjust the width to 25% for each subfigure
  \centering
  \raisebox{-\height}{\includegraphics[width=\linewidth]{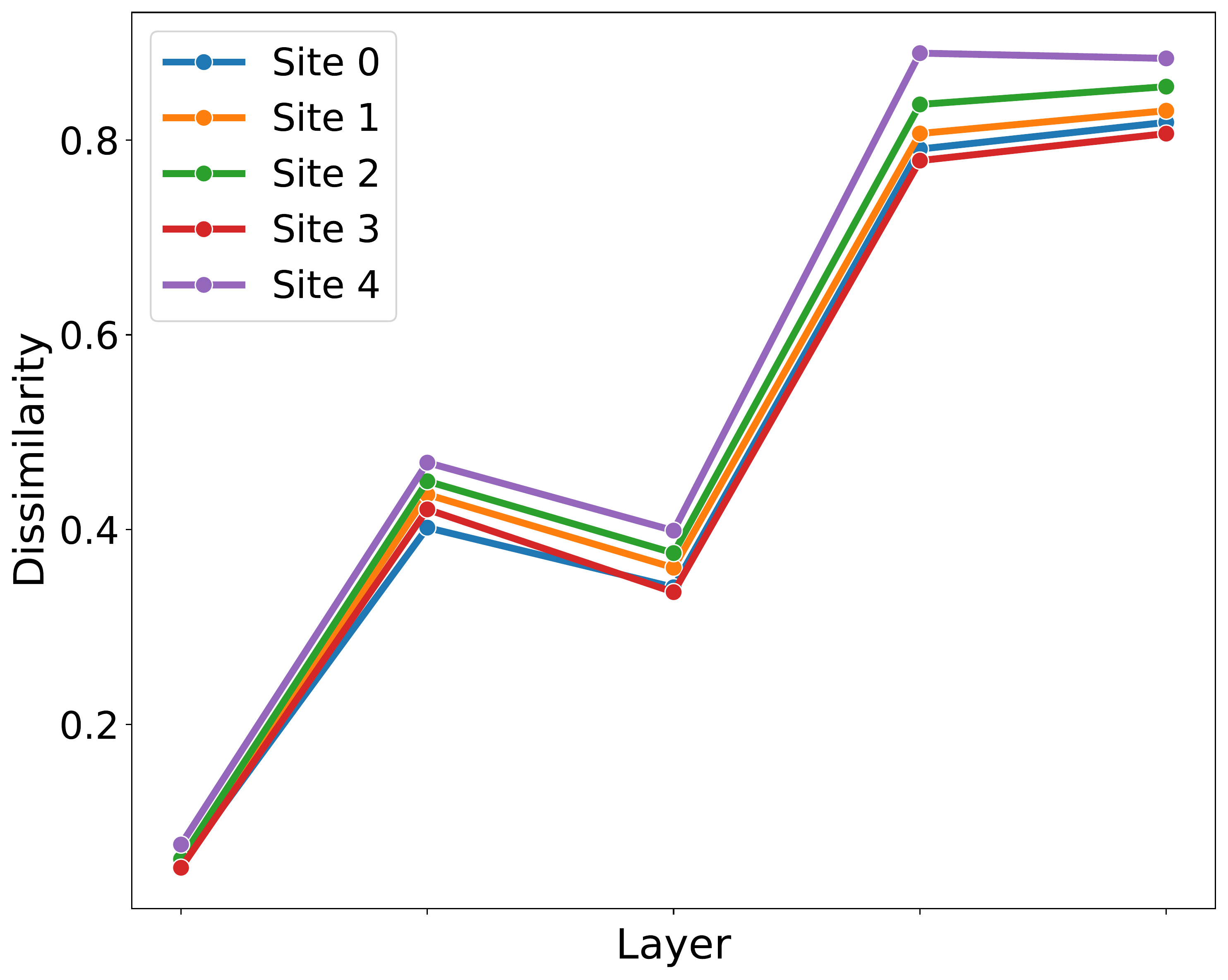}}
  \makebox[\linewidth][c]{\scriptsize\textit{(A) FashionMNIST}}  % Subfigure label
\end{minipage}%
\begin{minipage}{.22\linewidth}  % Adjust the width to 25% for each subfigure
  \centering
  \raisebox{-\height}{\includegraphics[width=\linewidth]{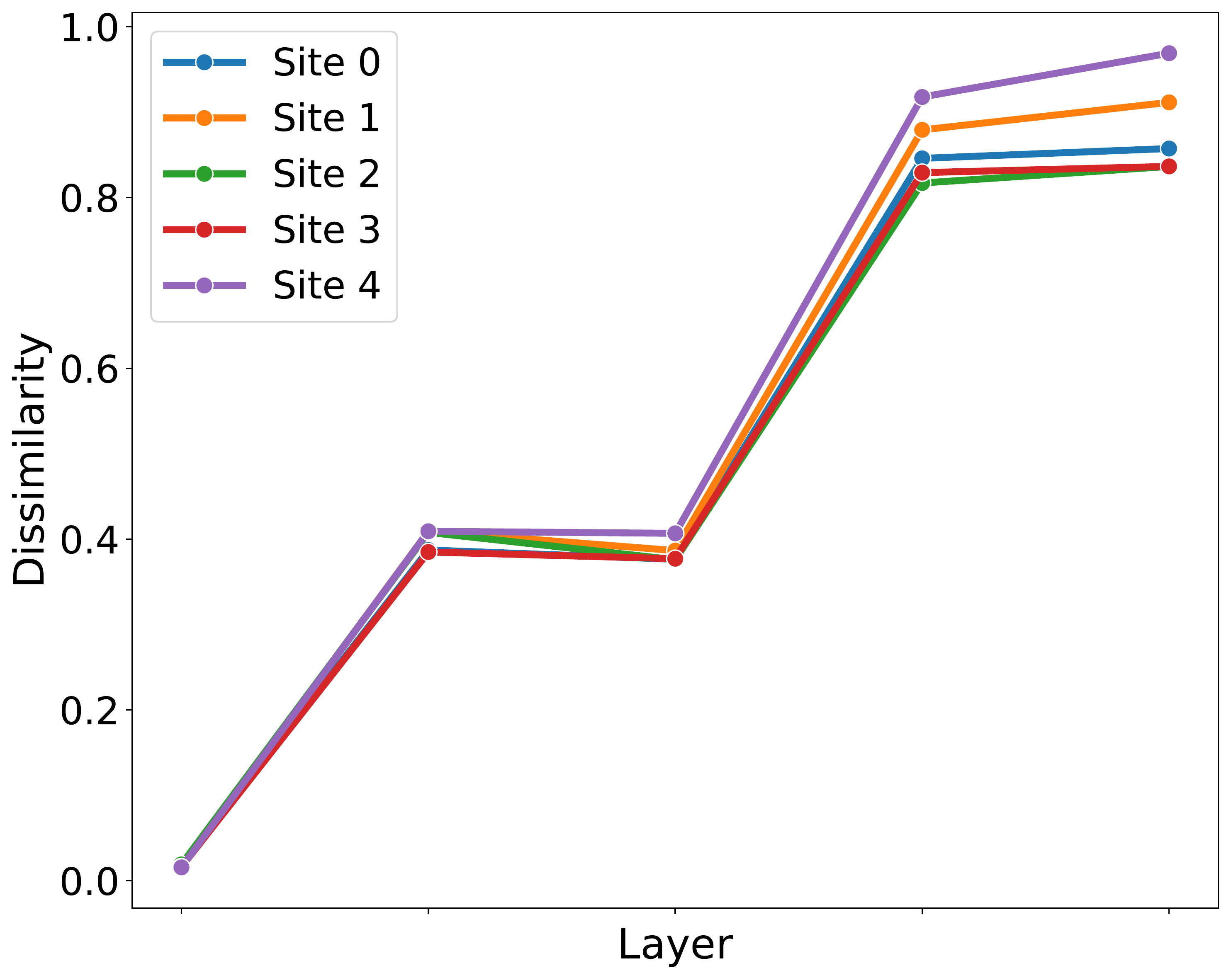}}
  \makebox[\linewidth][c]{\scriptsize\textit{(B) EMNIST}}  % Subfigure label
\end{minipage}%
\begin{minipage}{.22\linewidth}  % Adjust the width to 25% for each subfigure
  \centering
  \raisebox{-\height}{\includegraphics[width=\linewidth]{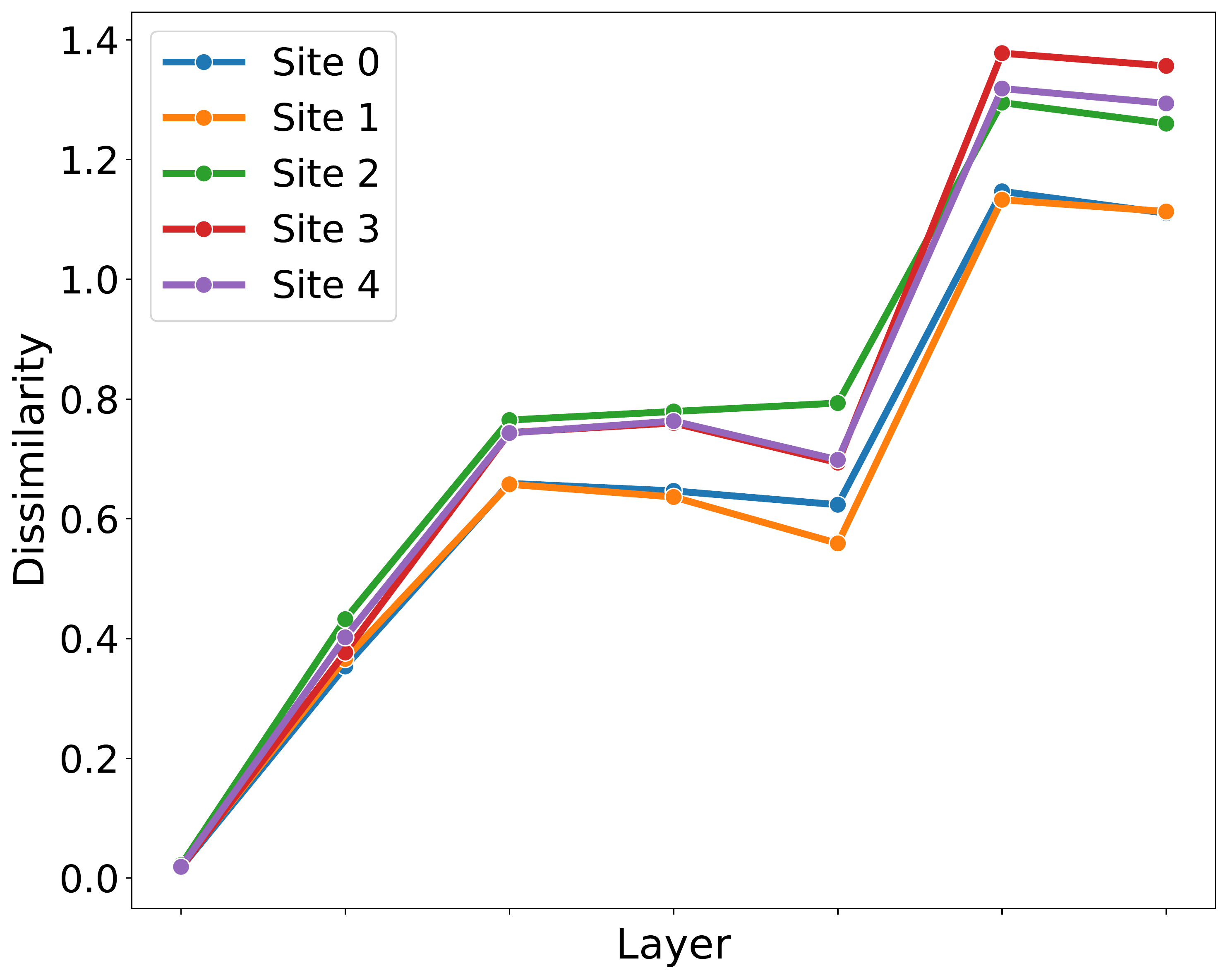}}
  \makebox[\linewidth][c]{\scriptsize\textit{(C) CIFAR-10}}  % Subfigure label
\end{minipage}%
\begin{minipage}{.22\linewidth}  % Adjust the width to 25% for each subfigure
  \centering
  \raisebox{-\height}{\includegraphics[width=\linewidth]{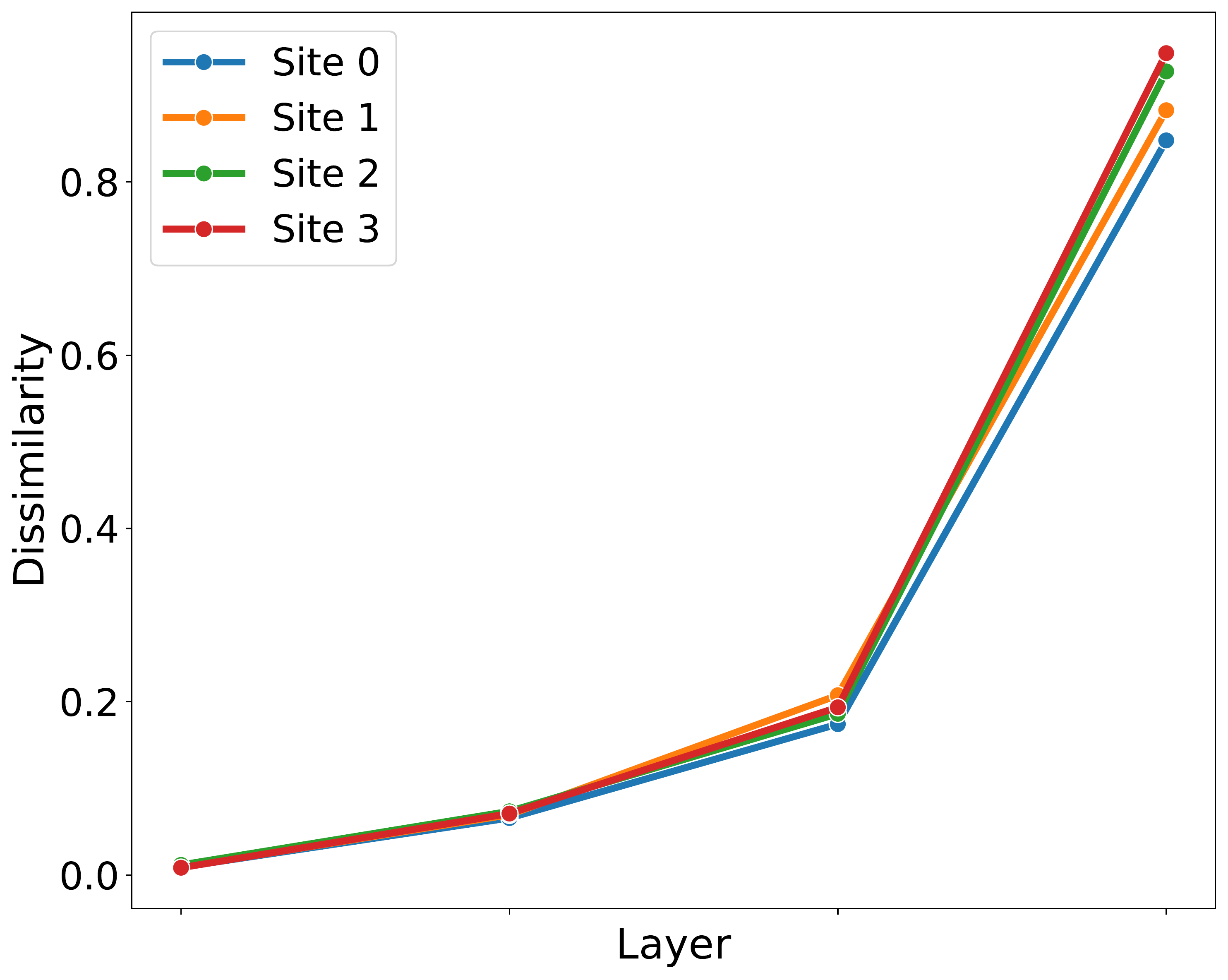}}
  \makebox[\linewidth][c]{\scriptsize\textit{(D) MIMIC-III}}  % Subfigure label
\end{minipage}
}
\caption{\textbf{Model representation similarity by layer}. Models identically initialized and independently trained on non-IID data.}
\label{fig:sample_rep}
\end{center}
\end{figure*}

We measure layer-wise representational similarity across all client models using Centered Kernel Alignment (CKA, Eq. \ref{eqn:sample_rep}) \cite{kornblith2019similarity}, which quantifies how similarly different models transform the same input samples at each layer. Figure \ref{fig:sample_rep} shows the average representational similarity for each model compared to all other models after one epoch of training. Figures \ref{sfig:sample_rep_first} and \ref{sfig:sample_rep_best} provide comparable analyses at the conclusion of training across all datasets.

The results validates our hypothesis: early layers exhibit high representational similarity across all clients (indicating convergence to similar solutions), while later layers show dramatically increasing dissimilarity (Figure \ref{fig:sample_rep}). This transition aligns precisely with the gradient-based metrics from the previous section, confirming that:

\begin{itemize}[leftmargin=*]
    \item \textbf{Early layers:} Converge to similar solutions and occupy flat regions and so are \emph{safe to federate without harming clients}
    \item \textbf{Later layers:} Diverge to client-specific solutions and occupy sharp regions so \emph{federation would disrupt optimization and degrade performance}
\end{itemize}

Critically, this transition point emerges after just one training epoch and becomes more pronounced with continued training, enabling early identification without prolonged exposure to harmful aggregation dynamics.

These findings provide empirical support for our central claim: 
\begin{hypothesis}[Early-layer robustness to federation under non-IID FL]
In non-IID settings, federating early layers -- which converge to similar, generalizable solutions -- does not harm client performance. 
Conversely, federating later layers is likely to degrade performance by disrupting task-specific optimization achieved locally.
\end{hypothesis}

\vspace{3mm}

\subsection{Federation sensitivity: Efficiently identifying the transition point}

Having empirically validated that a transition point exists we now introduce a principled metric to automatically identify this boundary for any architecture and dataset. This addresses a critical gap in partial FL: existing methods rely on architecture-specific heuristics that create deployment risks, while our approach provides a data-driven, parameter-light method for robust federation decisions.

\textbf{Key insight from model pruning.} Our approach builds on the observation that layers with strong cross-client generalization occupy flatter loss regions, a property quantifiable using parameter importance metrics from model pruning. In pruning, parameter importance $\mathcal{I}_p(\theta) = (\theta_p\nabla\theta_p)^2$ estimates each parameter's contribution to the loss using a first-order approximation \cite{molchanov2019importance}. We adapt this concept to the federation context: \emph{if a layer's parameters have low importance (indicating flat loss geometry and robustness to perturbation), that layer is suitable for federation}. Conversely, high parameter importance signals vulnerability to the perturbation induced by aggregation.

\textbf{Federation sensitivity metric.} We define the federation sensitivity of layer $l$, denoted $\mathcal{F}_l(\Theta)$, as:
\begin{equation}
\label{eqn:fo_approx}
    \mathcal{F}_l(\Theta) \triangleq \sum_{k=1}^l \frac{1}{n_k} \sum_{p =1}^{n_k} \mathcal{I}_p(\theta) = \sum_{k=1}^l \frac{1}{n_k} \sum_{p=1}^{n_k} (\theta_p\nabla\theta_p)^2
\end{equation}
where $\Theta$ represents all model parameters, $\theta_p$ is parameter $p$ in layer $k$, $\nabla\theta_p$ is its gradient, and $n_k$ is the number of parameters in layer $k$.

This metric incorporates two key design choices that reflect the constraints and goals of robust partial federation:
\begin{enumerate}[leftmargin=*]
    \item \textbf{Layer-wise normalization.} We divide by $n_k$ to enable fair comparison across layers of different sizes, ensuring that large layers do not artificially dominate the metric. This provides architectural robustness cross diverse model designs.
    \item \textbf{Cumulative aggregation.} We sum contributions from all layers up to $l$ because partial federation has an inherent constraint: selecting any layer for federation necessarily requires including all previous layers. This cumulative formulation naturally identifies the transition point where adding another layer to federation would become harmful.
\end{enumerate}

\textbf{Interpretation:} Federation sensitivity quantifies each layer's vulnerability to aggregation. Low values indicate robustness (suitable for federation); high values indicate vulnerability (would be harmed by federation).

\begin{figure*}[ht]
\begin{center}

% First four subfigures
\makebox[\linewidth][c]{
\begin{minipage}{.24\linewidth}
  \centering
  \raisebox{-\height}{\includegraphics[width=\linewidth]{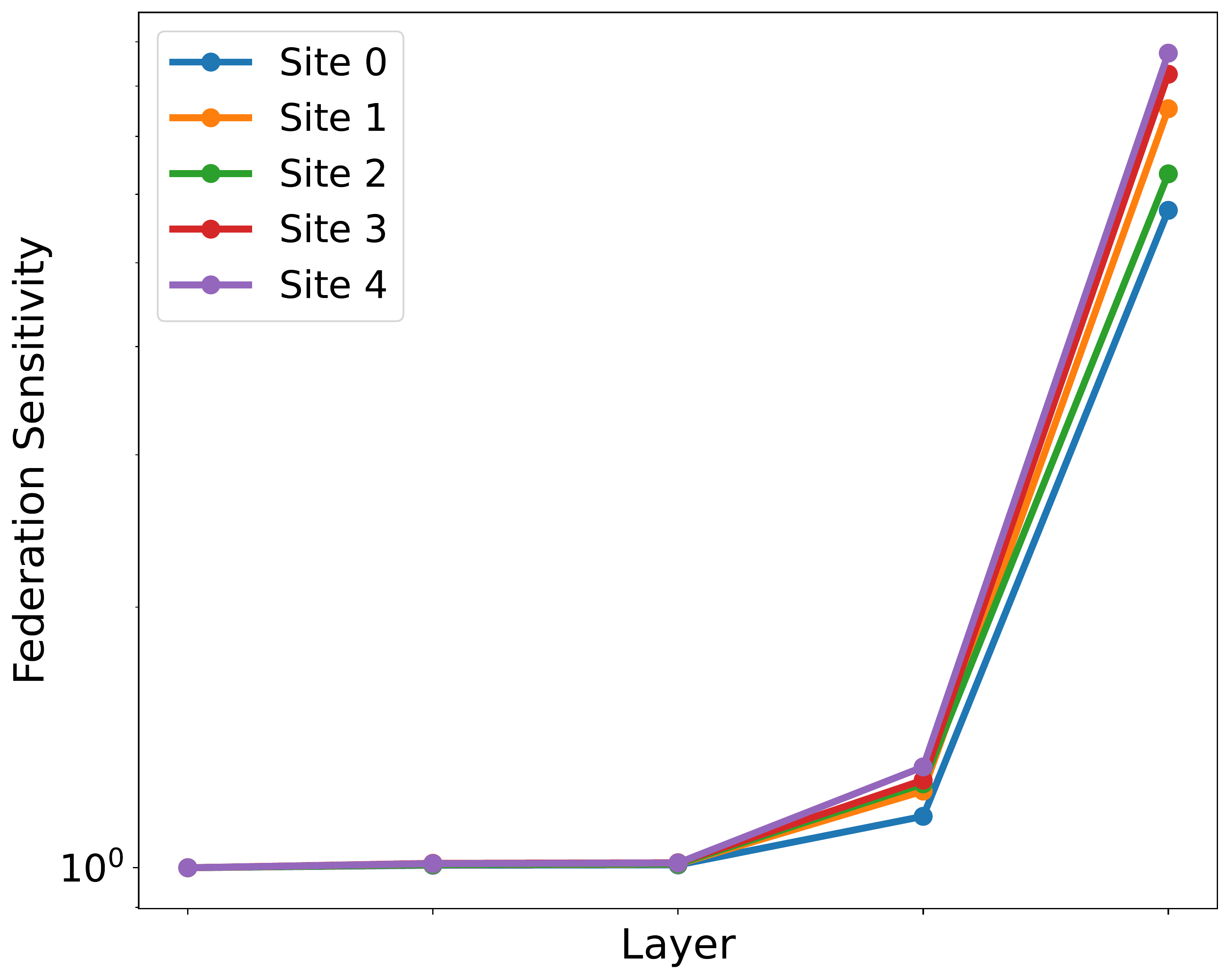}}
  \makebox[\linewidth][c]{\scriptsize\textit{(A) FashionMNIST}}  % Subfigure label
\end{minipage}%
\begin{minipage}{.24\linewidth}
  \centering
  \raisebox{-\height}{\includegraphics[width=\linewidth]{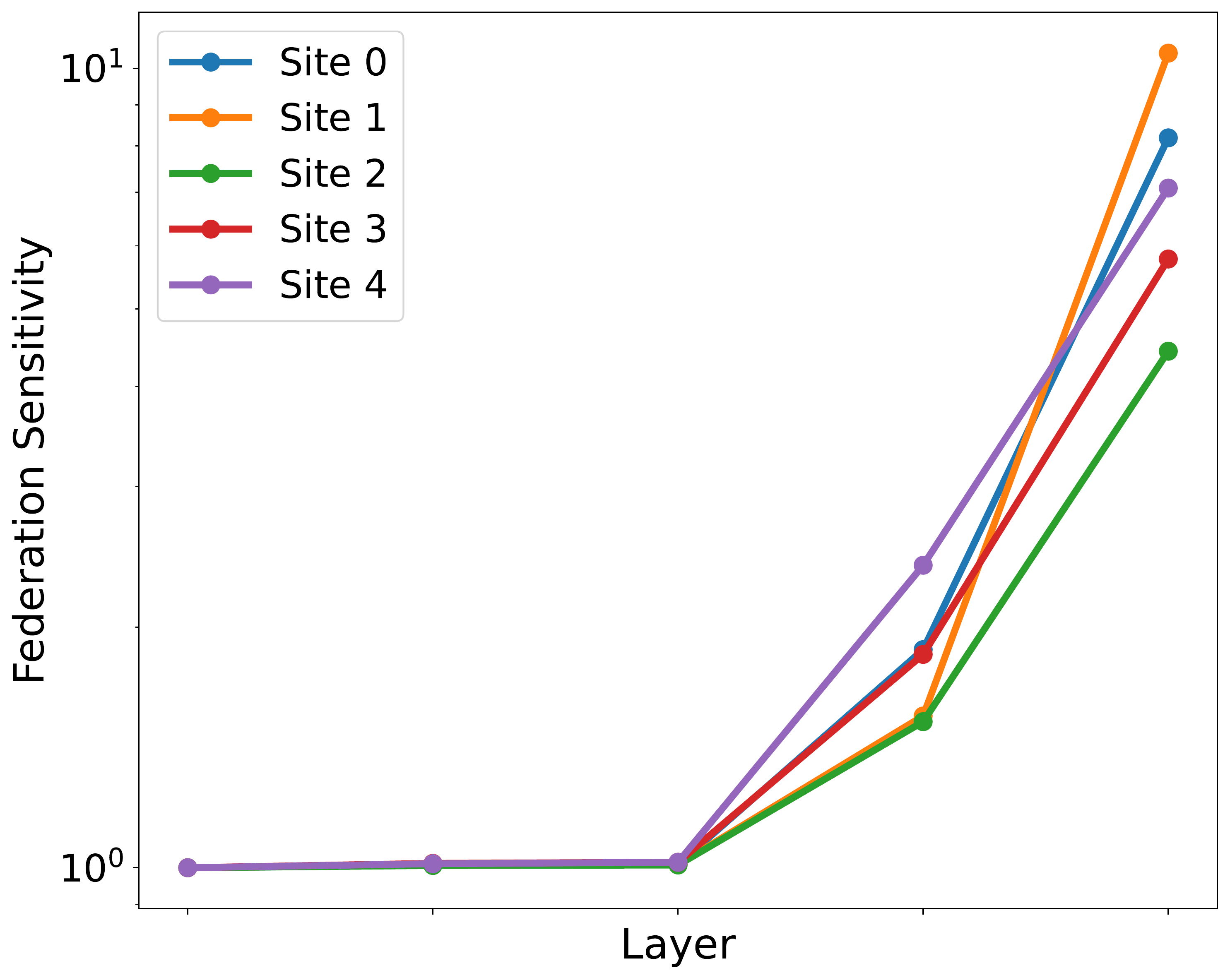}}
  \makebox[\linewidth][c]{\scriptsize\textit{(B) EMNIST}}  % Subfigure label
\end{minipage}%
\begin{minipage}{.24\linewidth}
  \centering
  \raisebox{-\height}{\includegraphics[width=\linewidth]{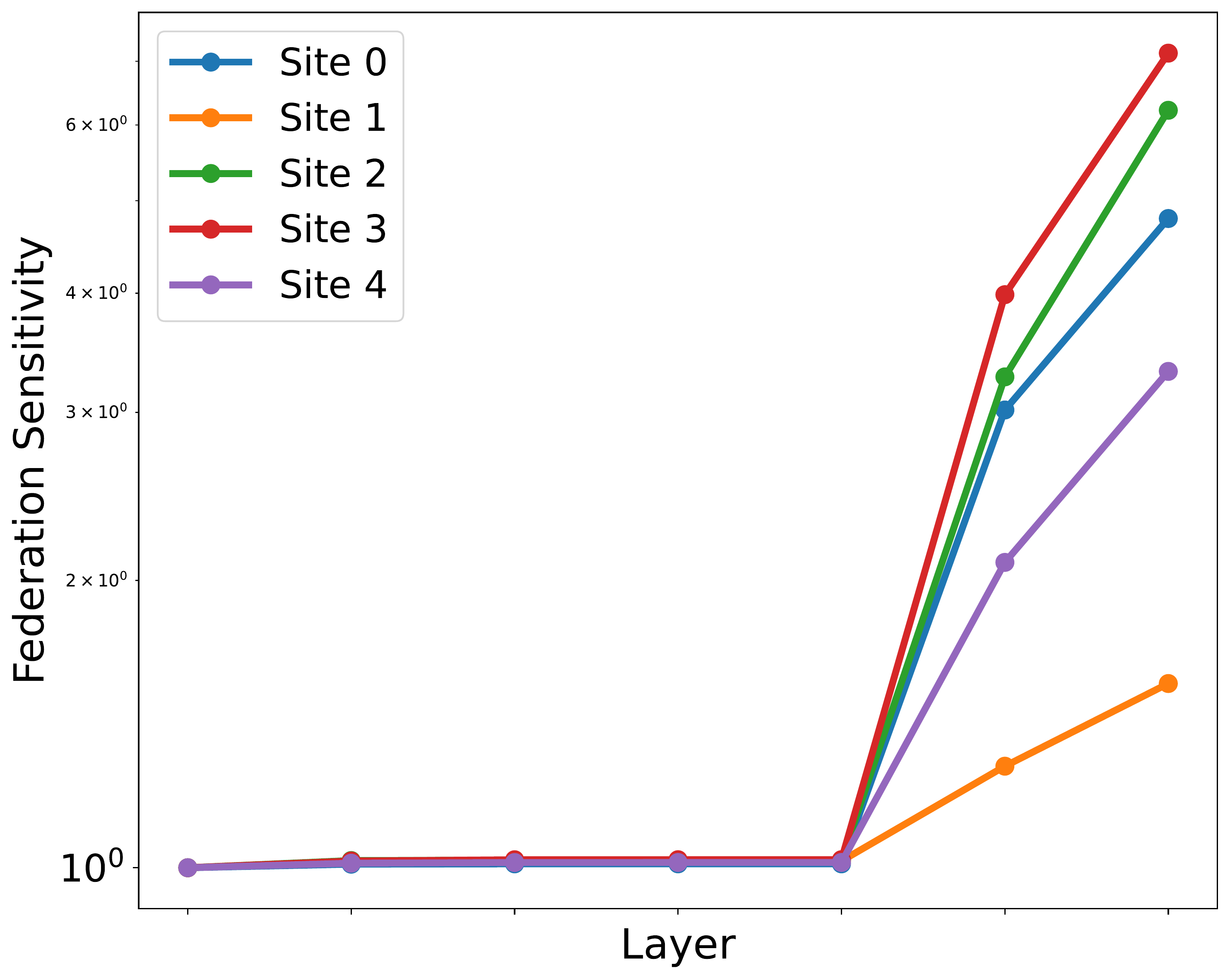}}
  \makebox[\linewidth][c]{\scriptsize\textit{(C) CIFAR-10}}  % Subfigure label
\end{minipage}%
\begin{minipage}{.24\linewidth}
  \centering
  \raisebox{-\height}{\includegraphics[width=\linewidth]{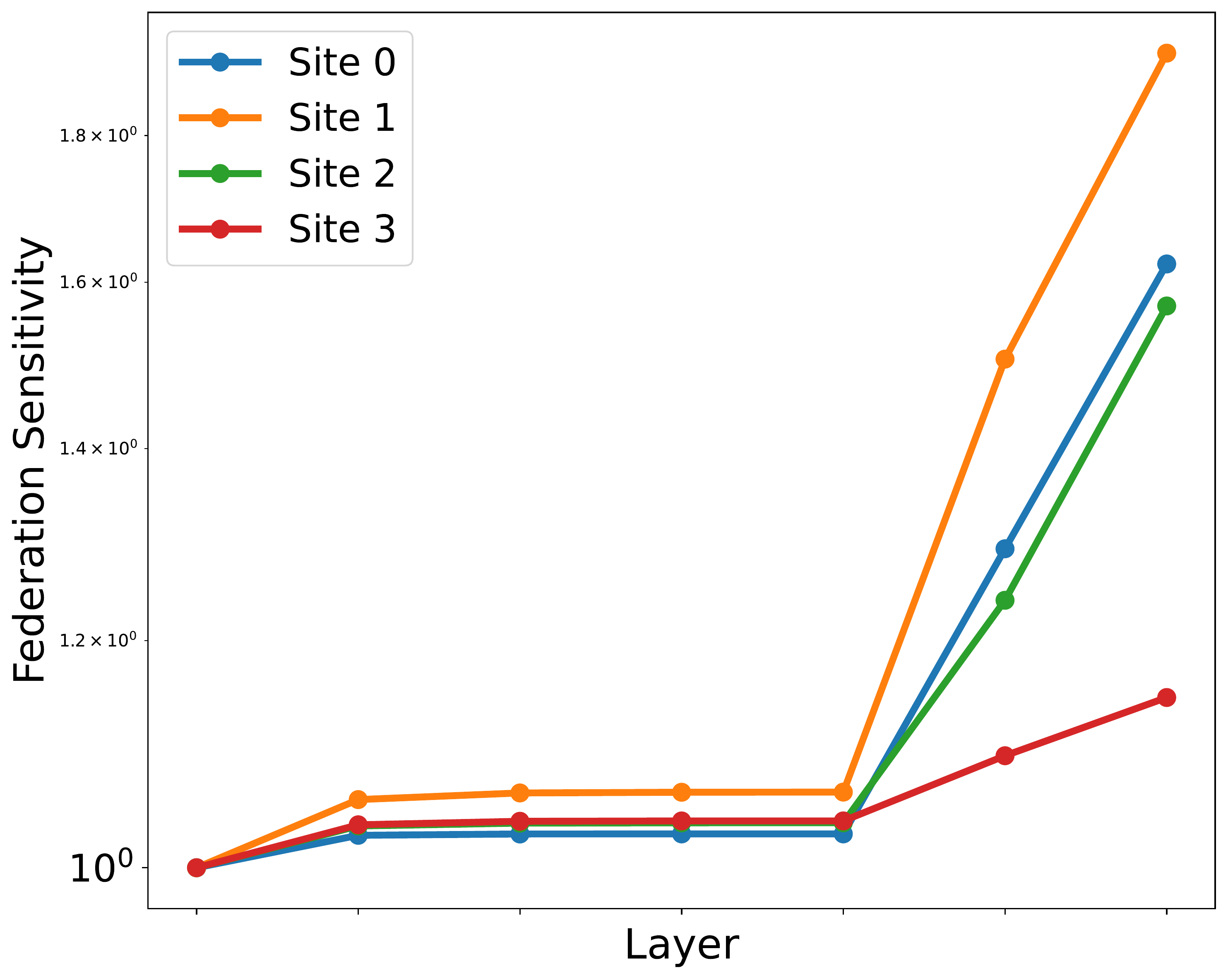}}
  \makebox[\linewidth][c]{\scriptsize\textit{(D) ISIC-2019}}  % Subfigure label
\end{minipage}
}

% Next three subfigures
\makebox[\linewidth][c]{
\begin{minipage}{.24\linewidth}
  \centering
  \raisebox{-\height}{\includegraphics[width=\linewidth]{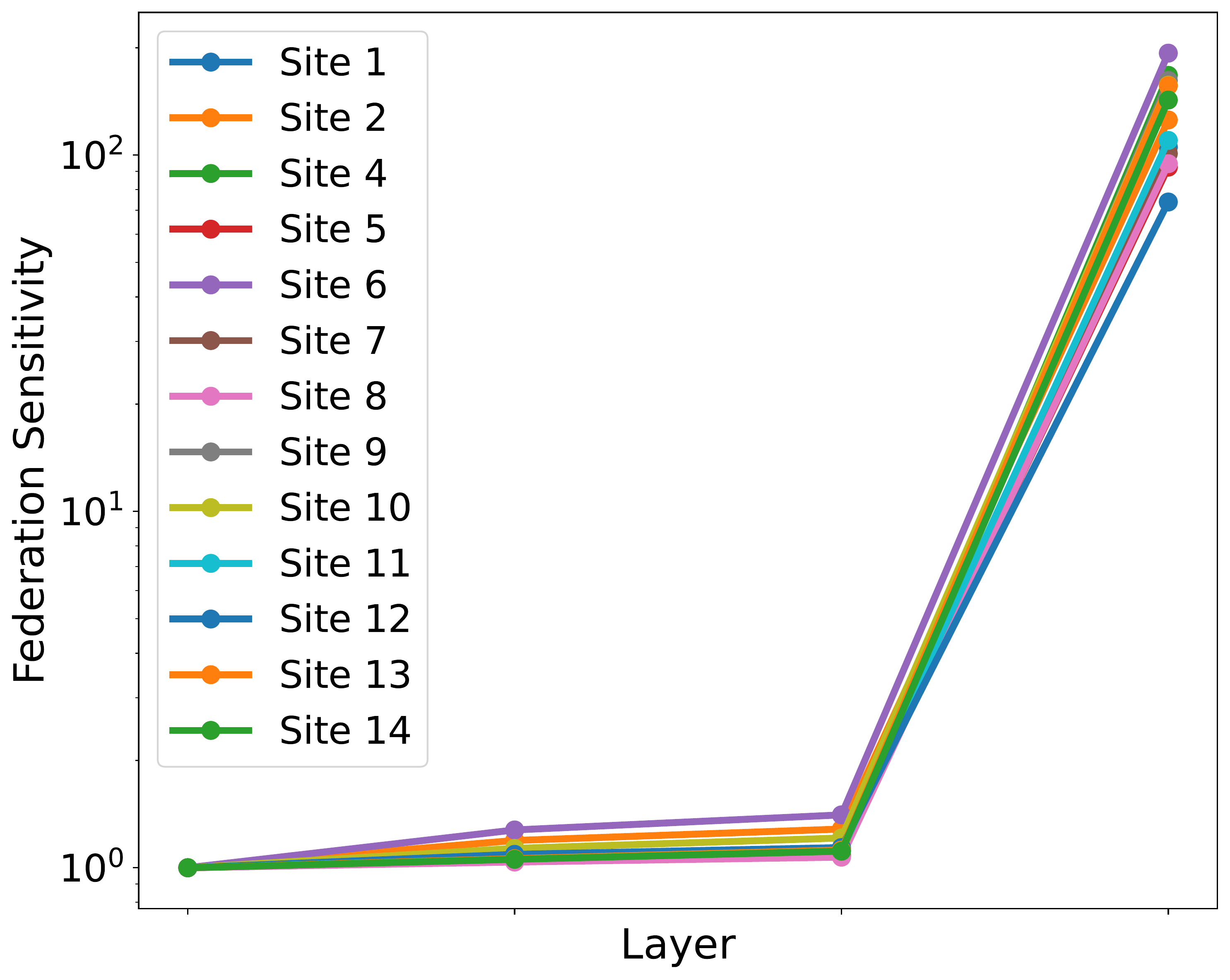}}
  \makebox[\linewidth][c]{\scriptsize\textit{(E) Sent-140}}  % Subfigure label
\end{minipage}%
\begin{minipage}{.24\linewidth}
  \centering
  \raisebox{-\height}{\includegraphics[width=\linewidth]{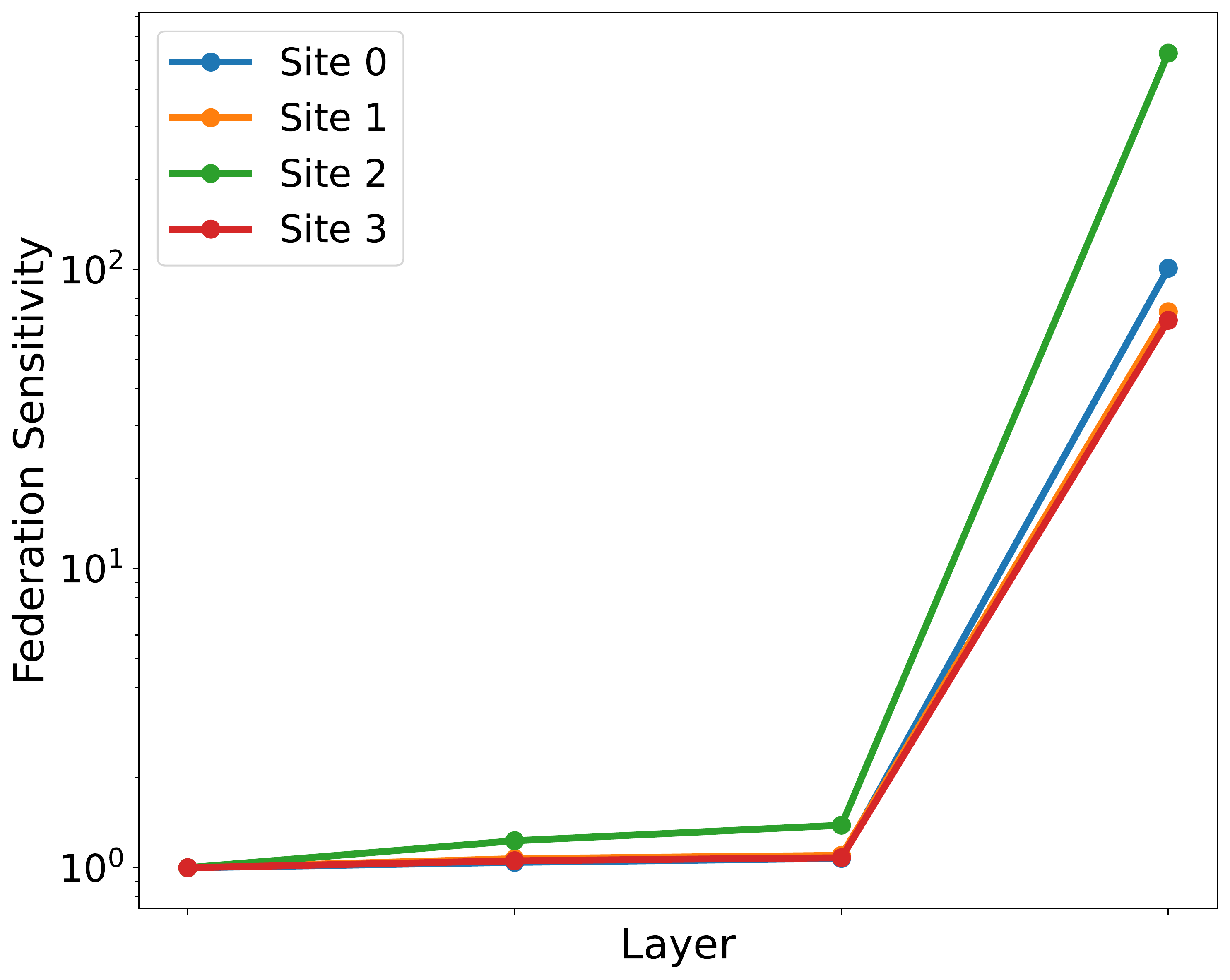}}
  \makebox[\linewidth][c]{\scriptsize\textit{(F) MIMIC-III}}  % Subfigure label
\end{minipage}%
\begin{minipage}{.24\linewidth}
  \centering
  \raisebox{-\height}{\includegraphics[width=\linewidth]{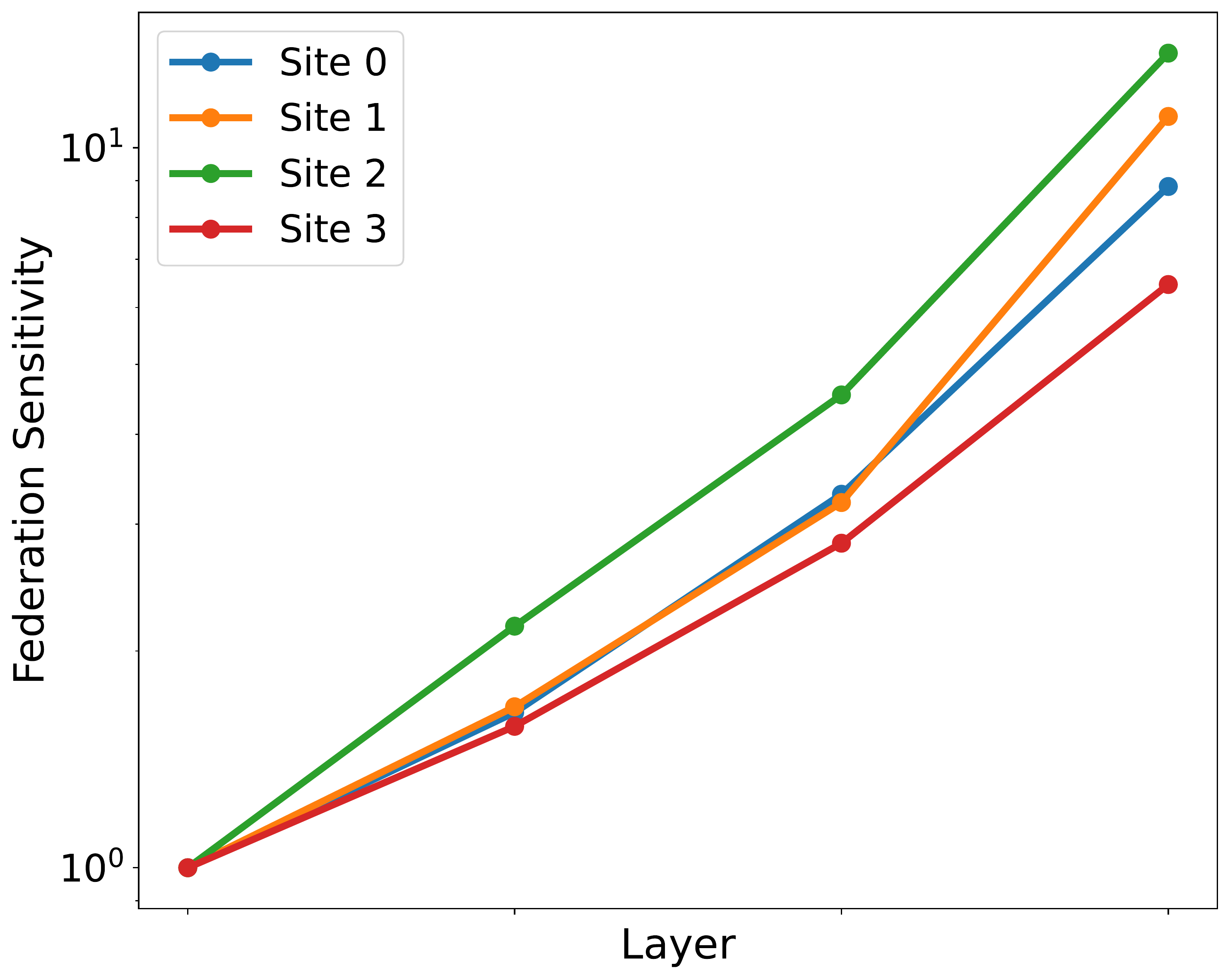}}
  \makebox[\linewidth][c]{\scriptsize\textit{(G) Fed-Heart-Disease}}  % Subfigure label
\end{minipage}
}
\caption{\textbf{Federation sensitivity after one epoch}. All models identically initialized and trained on non-IID data subsets. Values expressed as \% of first layer value}
\label{fig:layer_imp}
\end{center}
\end{figure*}

\textbf{Empirical validation of the transition point.} Figure \ref{fig:layer_imp} shows federation sensitivity after one epoch across all architectures (values normalized as percentage of first-layer sensitivity). Figures \ref{sfig:layer_imp_best} and \ref{sfig:layer_imp_best_fl} show results after full training. A characteristic pattern emerges: \textbf{federation sensitivity remains low in early layers, then spikes sharply in later layers}. 

Critically, this spike aligns with the transitions observed in our generalization metrics (gradient variance, Hessian eigenvalues) and representational similarity analysis (Section \ref{sec:gen_sim}). This convergence of multiple independent metrics provides strong empirical validation that federation sensitivity successfully identifies the transition point between layers suitable for federation and those that would suffer negative transfer.

\textbf{Protective properties enabling deployment.} Federation sensitivity offers several advantages critical for real-world deployment:

\begin{itemize}[leftmargin=*]
    \item \textbf{Stable:} Requires minimal hyperparameter tuning, eliminating a major source of deployment risk and ensuring consistent behavior across settings.
    
    \item \textbf{Computationally efficient:} Uses weights and gradients already computed during training, adding negligible overhead.
    
    \item \textbf{Architecture-agnostic:} Shows consistent patterns across Fully-Connected Networks, CNNs, and Transformers, enabling broad applicability without architecture-specific modifications.
    
    \item \textbf{Early emergence:} Identifies the optimal split point after just one epoch, minimizing exposure to harmful training dynamics while enabling rapid, reliable federation decisions.
\end{itemize}

\section{Proposed Framework}
We propose Principled Layer-wise-FL (PLayer-FL), a partial FL approach that uses federation sensitivity to dynamically determine which layers to federate (Algorithm \ref{alg:PLayer-FL-main}). A key advantage of our method is its computational efficiency -- federation sensitivity is calculated during the first epoch of training using weights and gradients that are already computed as part of the standard training process. To identify the transition point between federated and local layers, we compare the relative federation sensitivity between consecutive layers against a threshold $t$ (eqn. \ref{eqn:layer_select}). While this threshold controls how aggressively the algorithm federates layers, with higher values leading to more federated layers, we found in our experiments that the transition point remains stable across a wide range of threshold values (Figure \ref{fig:layer_imp}).

PLayer-FL stands out by offering a light-weight and principled approach for determining the layers to federate during training specifically to address that challenges of non-IID data. Algorithms \ref{alg:PLayer-FL-main} - \ref{alg:layer-split} present the formal pseudocode. For brevity, we omit procedures C\textsc{lient}U\textsc{pdate} and S\textsc{erver}A\textsc{ggregate} that are identical to FedAvg. Note that PLayer-FL conducts simultaneous updates to both the federated and localized layers. However, only the federated layers are sent to the central server for aggregation. We use the same learning rate for both federated and localized training, but this can be altered. Additionally, as our algorithm's sole purpose is to delineate layers to be federated, it can be used in conjunction with other personalized FL algorithms. 

\begin{algorithm}
   \caption{PLayer-FL: Main Algorithm}
   \label{alg:PLayer-FL-main}
\begin{algorithmic}
   \STATE {\bfseries Input:} Number of rounds $N$, set of clients $C$, threshold $t$
   \STATE {\bfseries Output:} Model parameters $\Theta_{c}$ for $c\in C$
   
   \STATE \textbf{Initialize} global model parameters $\Theta^0$
      \FORALL{clients $c \in C$ in parallel}
         \STATE $\Theta^1_{c} \leftarrow$ C\textsc{lient}U\textsc{pdate}($\Theta^0$)
         \STATE Calculate layer federation sensitivity (Algorithm \ref{alg:fed-sensitivity}):
         \STATE $\mathcal{F}(\Theta_c) \leftarrow$ C\textsc{alculate}F\textsc{ed}S\textsc{ensitivity}($\Theta_{c}$)
      \ENDFOR
      \STATE Determine federated and local layers (Algorithm \ref{alg:layer-split}):
      \STATE\hspace{5mm} $\Theta_{fed}, \Theta_{local}$ $\leftarrow \text{L\textsc{ayer}S\textsc{plit}}(\Theta, \{\mathcal{F}(\Theta_c)\}, t)$
    \STATE Execute FL on selected layers:      
   \FOR{$r = 1$ {\bfseries to} $N$}
      \FORALL{clients $c \in C$}
          \STATE $\Theta_{c}^{r} \leftarrow$ C\textsc{lient}U\textsc{pdate}($\Theta^{r}$)
      \ENDFOR
      \STATE $\Theta_{fed}^{r+1} \leftarrow \text{S\textsc{erver}A\textsc{ggregate}}(\{\Theta_{c,fed}^{r}\}, C)$
   \ENDFOR
\end{algorithmic}
\end{algorithm}
\begin{algorithm}
   \caption{PLayer-FL: C\textsc{alculate}F\textsc{ed}S\textsc{ensitivity}}
   \label{alg:fed-sensitivity}
\begin{algorithmic}
\STATE {\bfseries Input:} Client model parameters $\Theta_c$
\STATE {\bfseries Output:} Layer-wise federation sensitivity array $\mathcal{F}(\Theta_c)$
\STATE Initialize $\mathcal{F}(\Theta_c)$ as an empty array
\STATE Let $L$ be the set of all layers in $\Theta_c$
\STATE {\bfseries for all } {layers $l \in L$} {\bfseries do } 
       \INDSTATE Calculate federation sensitivity, $\mathcal{F}_{l}(\Theta_c)$, for layer $l$ as in equation \ref{eqn:fo_approx}
       \INDSTATE Append $\mathcal{F}_{l}(\Theta_c)$ to $\mathcal{F}(\Theta_c)$
\STATE {\bfseries end for }
\STATE {\bfseries Return} $\mathcal{F}(\Theta_c)$ 
\end{algorithmic}
\end{algorithm}
\begin{algorithm}
   \caption{PLayer-FL: \text{L\textsc{ayer}S\textsc{plit}}}
   \label{alg:layer-split}
\begin{algorithmic}
\STATE {\bfseries Input:} Model parameters $\Theta$, client sensitivities $\{\mathcal{F}(\Theta_c)\}$, threshold $t$
\STATE {\bfseries Output:} Federated layers $\Theta_{fed}$, local layers $\Theta_{local}$
   \STATE Aggregate client federation sensitivities:
   \STATE \hspace{5mm} $\mathcal{F}(\Theta) \leftarrow \sum^C_{c=1} \mathcal{F}(\Theta_c)$
   \STATE Find transition point $p$ where relative sensitivity exceeds threshold $t$:
   \STATE \vspace{-4mm}\begin{equation}  \hspace{-15mm}
       p \leftarrow \min \{p: \mathcal{F}_{p+1}(\Theta_c) \mathcal{F}_p(\Theta) > t \}
       \label{eqn:layer_select}
   \end{equation}
   \vspace{-5mm}
   \STATE Partition model at point $p$:
   \STATE \hspace{5mm} $\Theta_{fed} \leftarrow \Theta_{l \leq p}$, $\Theta_{local} \leftarrow \Theta_{l > p}$
   \STATE {\bfseries Return} $\Theta_{fed}, \Theta_{local}$  
\end{algorithmic}
\end{algorithm}

PLayer-FL has the same computational complexity as FedAvg. This is because, compared to FedAvg, PLayer-FL introduces two procedures, both executed once during training: C\textsc{alculate}F\textsc{ed}S\textsc{ensitivity} (Algorithm \ref{alg:fed-sensitivity}) and L\textsc{ayer}S\textsc{plit} (Algorithm \ref{alg:layer-split}) with computational complexity of $O(P)$, where $P$ is the number of parameters, and $O(L)$, where $L$ is the number of layers, respectively.

\section{Evaluation}
\label{evaluation}
\subsection{Methods}
Our setup focuses on \textbf{cross-silo FL}: we use \emph{highly} non-IID datasets from a limited number of clients, all of whom have sufficient data to train a local model and are available for each round of training (see Section \ref{background:cross_fl} for justification).

\subsubsection{Datasets}
We evaluate performance on FashionMNIST, ExtendedMNIST, CIFAR-10, ISIC2019, Fed-Heart-Disease, Sentiment-140 and MIMIC-III. The datasets cover tabular, imaging, and natural language modalities. The first three datasets do not have a natural partition, thus we introduce label skew using a Dirichlet distribution ($\alpha=0.5$) following \citet{li2022federated}. The other datasets have natural partitions:
\begin{itemize}[leftmargin=*]
    \item Fed-ISIC-2019 \cite{terrail2022flamby}: Dermoscopy skin lesion classification (4 classes); partitioned by hospital.
    \item Fed-Heart-Disease \cite{terrail2022flamby}: Patient heart disease classification (5 classes); partitioned by hospital.
    \item Sent-140 \cite{caldas2018leaf}: Classification of tweet sentiments; partitioned by users, selecting the top 15 users. 
    \item MIMIC-III \cite{johnson2016mimic}: Mortality prediction using hospital admission notes; partitioned by diagnosis. 
\end{itemize}

\subsubsection{Performance}
We evaluate our method against three categories of baselines: (1) standard approaches -- local site training and FedAvg \cite{mcmahan2017communication}, (2) established personalized FL algorithms -- FedProx \cite{li2020federated}, pFedMe \cite{t2020pfedme}, Ditto \cite{li2021ditto}, and Local Adaptation \cite{yu2020salvaging}, and (3) existing partial FL methods -- FedBABU \cite{oh2021fedbabu}, FedLP \cite{zhu2023fedlp}, FedLAMA \cite{lee2023layer}, and pFedLA \cite{ma2022layer}. We also evaluate PLayer-FL-Random, which federate up to a randomly selected layer. This serves as proxy for \textbf{non-principled} partial FL approaches. Full training and evaluation details are in Section \ref{supp:model_training}.

We evaluate performance using F1 score (macro-averaged \emph{i.e.,} treating each label equally, see eqn. \ref{eqn:f1}), accuracy, and test loss. The macro-F1 score serves as our primary metric due to its robustness to class imbalance. Results for accuracy and test loss are presented in Sections \ref{supp:accuracy_results} and \ref{supp:loss_results}. For personalized FL algorithms, we also study per-site metrics. This is important as personalized models that achieve the highest average performance may not necessarily uniformly benefit every site \cite{divi2021new}. To evaluate fairness, we compare the variance in performance across clients (eqn. \ref{eqn:fairness}) \cite{divi2021new}. We also evaluate participation incentive by calculating the percentage of clients where performance surpasses both local site training and FedAvg (eqn. \ref{eqn:incentive}) \cite{cho2022federate}. If our intuition that federating generalizable layers should not negatively impact performance holds true, we anticipate our method to exhibit high fairness and participation incentive. To make conclusions across multiple datasets, we use the Friedman test \cite{demvsar2006statistical}.

\subsection{Results}
\begin{table*}[htbp]
\caption{Macro-averaged F1 Score and average rank. In \textbf{bold} is top-performing model. Friedman rank test p-value $<5\times10^{-3}$}
\label{f1-table}
\begin{center}
\setlength{\tabcolsep}{5.3pt}
\begin{tabular}{lcccccccc}
\toprule
Algorithm & FMNIST & EMNIST & CIFAR & ISIC & Heart & Sentiment & mimic & Mean Rank \\
\midrule
Local & \specialcell{75.9}{1.6} & \specialcell{56.9}{1.4} & \specialcell{61.6}{3.2} & \textbf{\specialcell{53.1}{1.3}} & \textbf{\specialcell{42.0}{0.8}} & \specialcell{57.4}{1.0} & \specialcell{58.6}{1.5} & 6.9 \\
FedAvg & \specialcell{75.6}{1.8} & \specialcell{64.2}{1.1} & \specialcell{65.4}{4.5} & \specialcell{43.3}{0.7} & \specialcell{40.4}{1.0} & \specialcell{52.6}{0.3} & \specialcell{63.0}{1.7} & 6.0 \\
FedProx & \specialcell{76.4}{1.8} & \specialcell{65.3}{1.9} & \specialcell{65.8}{4.9} & \specialcell{40.6}{2.6} & \specialcell{37.2}{1.6} & \specialcell{52.6}{0.3} & \specialcell{63.2}{0.8} & 6.3 \\
pFedMe & \specialcell{77.3}{1.8} & \specialcell{66.2}{0.7} & \specialcell{48.9}{2.5} & \specialcell{44.9}{2.6} & \specialcell{42.5}{0.8} & \specialcell{58.9}{0.4} & \specialcell{60.9}{0.4} & 5.3 \\
Ditto & \specialcell{77.8}{1.2} & \specialcell{61.6}{0.8} & \specialcell{48.7}{1.9} & \specialcell{43.4}{0.7} & \specialcell{40.7}{1.0} & \specialcell{58.9}{0.9} & \specialcell{61.0}{0.5} & 6.0 \\
LocalAdaptation & \specialcell{76.9}{1.5} & \specialcell{63.8}{1.8} & \specialcell{65.5}{3.3} & \specialcell{43.5}{1.9} & \specialcell{39.8}{0.9} & \specialcell{53.2}{0.6} & \specialcell{62.6}{1.6} & 6.4 \\
FedBABU & \specialcell{77.2}{1.0} & \textbf{\specialcell{66.1}{0.2}} & \textbf{\specialcell{67.7}{4.0}} & \specialcell{49.1}{3.5} & \specialcell{40.3}{0.8} & \specialcell{58.3}{0.2} & \specialcell{62.7}{0.4} & 4.1 \\
FedLP & \specialcell{77.2}{1.3} & \specialcell{64.1}{0.9} & \specialcell{63.7}{5.1} & \specialcell{40.2}{2.1} & \specialcell{40.1}{1.0} & \specialcell{54.4}{1.1} & \specialcell{62.1}{1.3} & 6.7 \\
FedLama & \specialcell{71.6}{1.7} & \specialcell{56.9}{1.3} & \specialcell{41.8}{3.0} & \specialcell{36.0}{1.1} & \specialcell{39.7}{0.8} & \specialcell{50.3}{0.4} & \textbf{\specialcell{66.7}{1.0}} & 10.1 \\
pFedLA & \specialcell{47.1}{0.9} & \specialcell{07.1}{0.0} & \specialcell{10.5}{1.6} & \specialcell{28.4}{1.9} & \specialcell{40.1}{0.8} & \specialcell{50.3}{0.4} & \specialcell{65.0}{2.6} & 11.4 \\
\midrule
PLayer-FL & \textbf{\specialcell{79.3}{1.4}} & \specialcell{62.2}{1.2} & \specialcell{67.1}{3.7} & \specialcell{52.1}{1.5} & \specialcell{41.9}{1.0} & \textbf{\specialcell{59.6}{1.2}} & \specialcell{63.0}{0.7} & \textbf{2.6} \\
PLayer-FL-Random & \specialcell{76.5}{2.3} & \specialcell{57.1}{1.1} & \specialcell{65.7}{2.3} & \specialcell{52.1}{1.5} & \specialcell{41.3}{0.9} & \specialcell{55.7}{0.9} & \specialcell{61.0}{1.4} & 6.1 \\
\bottomrule
\end{tabular}
\end{center}
\end{table*}

\textbf{PLayer-FL achieves robust and equitable FL collaboration.} Our results demonstrate that PLayer-FL delivers consistently competitive performance while ensuring equitable outcomes and strong participation incentives.

\textbf{Consistent and competitive performance across diverse settings.} Table \ref{f1-table} presents F1 scores and average ranks across all datasets (accuracy and loss in Tables \ref{supp:acc-table} and \ref{supp:loss-table} show consistent conclusions). PLayer-FL achieves the highest mean rank with statistical significance ($p<5\times10^{-3}$), demonstrating robust performance across various architectures and realistic non-IID conditions. Notably, in line with typical observations in cross-silo FL settings where data is highly non-IID \citep{huang2022cross}, local site training is competitive when compared to FL algorithms, as each site can effectively train its own model.

Importantly, while PLayer-FL does not always achieve the top individual score, it consistently ranks in the top three across all benchmarks, a pattern that reflects a critical protective property: \emph{stability without hyperparameter tuning}. Unlike optimization-based personalized FL methods that require careful tuning of regularization weights or learning rates (which can harm performance when misconfigured), PLayer-FL's parameter-light design eliminates this deployment risk. By relying on generalization dynamics observable in the first epoch rather than task-specific hyperparameters, PLayer-FL provides \emph{stable, predictable performance across diverse settings}.

\textbf{Fairness: Equitable benefits across all clients.} Table \ref{rank-table} shows that PLayer-FL achieves the \emph{best average fairness rank} across all benchmarks (full results in Table \ref{supp:fairness-table}), with statistical significance ($p<5\times10^{-3}$). This demonstrates that PLayer-FL distributes performance improvements equitably rather than benefiting only a subset of clients, a property that ensures no client is systematically disadvantaged. This fairness is not incidental but emerges directly from our principled approach: by federating only layers where clients naturally converge, we avoid the harmful averaging of divergent, client-specific representations that causes inequitable outcomes in standard FL.

\textbf{Incentivization: Protecting clients from performance degradation.} Beyond fairness, PLayer-FL achieves the \emph{best average incentivization rank} (Table \ref{rank-table}, full results in Table \ref{supp:incentive-table}, $p=0.043$), meaning that clients are protected from performing worse than local training. This is essential for sustainable collaboration: institutions will only participate if federation does not harm their individual performance. By identifying and respecting the transition point, PLayer-FL ensures that collaboration provides value to all participants -- addressing a fundamental barrier to FL adoption in cross-silo settings.

\textbf{Validation of the transition point approach.} These results provide comprehensive empirical validation of our core hypothesis: \emph{federating early, generalizable layers while keeping later, task-specific layers local provides robust protective guarantees}. The superiority of PLayer-FL over PLayer-FL-Random further demonstrates that our federation sensitivity metric successfully identifies the transition point -- achieving advantages beyond those of arbitrary partial federation. Together, these findings establish that principled identification of a transition point enables FL that is simultaneously effective, equitable, and protective.

\begin{table}[htbp]
\centering
\caption{Fairness and incentivization average ranks, Friedman rank test p-value $<5\times10^{-3}$ and $=0.043$, respectively}
\label{rank-table}
\setlength{\tabcolsep}{2.7pt}
\begin{tabular}{lcc}
\toprule
Algorithm & Fairness & Incentivization \\ 
\midrule
FedProx           & 7.7 & 5.4 \\
pFedMe            & 4.3 & 5.1 \\
Ditto             & 5.0 &  5.6 \\
LocalAdaptation   & 6.4 & 5.2 \\
FedBABU           & 4.4 &  4.4 \\
FedLP             & 6.1 &  6.2 \\
FedLAMA           & 5.6 & 7.3 \\
pFedLA            & 5.1 &  7.7 \\
\midrule
PLayer-FL          & \textbf{3.8} &  \textbf{3.4} \\
PLayer-FL-Random       & 6.5 &  4.8 \\
\bottomrule
\end{tabular}
\end{table}

\section{Conclusion}
In this work, we present PLayer-FL, a principled partial FL algorithm that uses a novel federation sensitivity metric to determine which layers to federate and which to locally train. We show that this metric correlates with known measures of generalizability and model similarity and that it emerges after only one epoch of training. As such, it can be readily incorporated into the training algorithm with minimal computational overhead. Over a set of rigorous experiments that include benchmark and real-world datasets, we show that PLayer-FL outperforms existing algorithms in cross-silo settings. We also show that it produces fairer outcomes and is more likely to incentivize participation in FL.

\textbf{Limitations}: Our work demonstrates strong empirical evidence for the federation sensitivity metric across diverse model architectures but currently lacks theoretical guarantees. However, the success of similar metrics in related fields and the established understanding of generalizable versus task-specific layers reinforces the validity and utility of our approach. Developing formal theoretical foundations presents a promising direction for future research. Additionally, our study primarily focuses on cross-silo FL, where clients have sufficient data to calculate federation sensitivity, rather than the cross-device setting. However, this focus is deliberate, as cross-silo FL represents a significant sub-field with critical applications in industries like healthcare. These settings face unique challenges, making the development of tailored method essential \cite{terrail2022flamby, huang2022cross, kairouz2021advances, pati2021federated}.

\bibliography{bibliography}
\bibliographystyle{mlsys2025}

%%%%%%%%%%%%%%%%%%%%%%%%%%%%%%%%%%%%%%%%%%%%%%%%%%%%%%%%%%%%%%%%%%%%%%%%%%%%%%%
%%%%%%%%%%%%%%%%%%%%%%%%%%%%%%%%%%%%%%%%%%%%%%%%%%%%%%%%%%%%%%%%%%%%%%%%%%%%%%%
% APPENDIX
%%%%%%%%%%%%%%%%%%%%%%%%%%%%%%%%%%%%%%%%%%%%%%%%%%%%%%%%%%%%%%%%%%%%%%%%%%%%%%%
%%%%%%%%%%%%%%%%%%%%%%%%%%%%%%%%%%%%%%%%%%%%%%%%%%%%%%%%%%%%%%%%%%%%%%%%%%%%%%%
\newpage
\appendix
\onecolumn
\renewcommand{\thetable}{A.\arabic{table}}
\renewcommand{\thefigure}{A.\arabic{figure}}
\renewcommand{\theequation}{A.\arabic{equation}} 
\renewcommand{\thesection}{A.\arabic{section}}
\setcounter{section}{0}
\setcounter{table}{0}
\setcounter{figure}{0}
\setcounter{equation}{0}

\section{Metric definitions}
\label{metric_defn}
\subsection{Gradient variance}
Gradient variance is proposed by \citet{jiang2019fantastic} and is defined as:
\begin{equation}
\label{eqn:grad_var}
    \text{Var}(\nabla\theta_i):=\frac{1}{n}\sum_{j=1}^n\left( \nabla\theta_i^j - \overline{\nabla\theta_i}  \right)^T\left( \nabla\theta_i^j - \overline{\nabla\theta_i}  \right)
\end{equation}
where $\theta_i^j$ is parameter $j$ in layer $i$, $ \nabla\theta_i^j$ is the gradient with respect to that parameter and $\overline{\nabla\theta_i}$ is the mean gradient of all parameters in layer $i$.
\subsection{Hessian eigenvalue sum}
Hessian eigenvalue sum is proposed by \citet{chaudhari2019entropy}. The Hessian at layer $i$, denoted $H_i$, is a square matrix of second-order partial derivatives with respect to parameters at layer $i$. Each entry in $H_i$ is defined as:
\begin{equation*}
    (H_i)_{jk} = \frac{\partial^2L}{\partial\theta_i^j\partial\theta_i^k}
\end{equation*}
where $\theta_i^j$ and $\theta_i^k$ are parameters $j$ and $k$ in layer $i$, $L$ is the loss and $\partial.$ is the partial derivative with respect to that parameter. The sum of the eigenvalues is then calculated as:
\begin{equation}
\label{eqn:hess_ev}
    \text{Tr}(H_i) = \sum_{p=1}^n \lambda_i^p
\end{equation}
where $\lambda_i^p$ is the $p^{th}$ eigenvalue of $H_i$.

\subsection{Sample representation similarity}
Sample representation similarity is calculated via Centered Kernel Alignment (CKA) which is proposed by \citet{kornblith2019similarity}. Its calculated as:
\begin{equation}
\label{eqn:sample_rep}
    \frac{||Y_i^TX_i||_F^2}{||X_i^TX_i||_F^2||Y_i^TY_i||_F^2}
\end{equation}
where $X_i$ and $Y_i$ are sample representations after layer $i$ coming from two distinct models.

\subsection{F1 score}
Multi-class  macro-averaged F1 score is calculated as
\begin{equation}
\label{eqn:f1}
\text{F1}_{\text{macro}} = \frac{1}{N}\sum_{i=1}^N 2 \cdot \frac{\text{precision}_i \cdot \text{recall}_i}{\text{precision}_i + \text{recall}_i}
\end{equation}
where $N$ is the number of classes, and precision$_i$ and recall$_i$ are the precision and recall for class $i$. This equation gives equal weight to all classes which is useful in imbalanced datasets like ours.

\subsection{Algorithm Fairness}
Fairness is calculated as described in \cite{divi2021new} - the variance in individual client performances on their local test set:
\begin{equation}
\label{eqn:fairness}
\text{Fairness} = \frac{1}{C}\sum_{c=1}^C \left( P_c - \overline{P_c}\right)^2
\end{equation}
where $C$ is the number of clients, $P_c$ is the performance of client $c$ and $\overline{P_c}$ is the average performance of all clients for a given personalized algorithm.

\subsection{Algorithm Incentivization}
Incentivization is defined as in \cite{cho2022federate} - the percentage of clients that outperform their local site model or global model from FedAvg:
\begin{equation}
\label{eqn:incentive}
\text{Incentivization} = \frac{1}{C}\sum_{c=1}^C \mathbb{I} \{ P_c > \max(S_c, G_c) \}
\end{equation}
where $C$ is the number of clients, $P_c$ is performance of the personalized model in client $c$, $S_c$ is the performance of the local site model in client $c$, and $G_c$ is the performance of the global Fed Avg model in client $c$
\section{Datasets}
Table \ref{supp:dataset-table} provides a description of the datasets which include: 4 image, 1 tabular and 2 NLP tasks. For FashionMNIST, EMNIST, and CIFAR-10 we create non-IID datasets by via label skew using a a dirichlet distribution ($\alpha = 0.5)$. The remaining datasets have a natural partition that we adhere to. For Sent-140 we select the top 15 users by tweet count. 

\begin{table}[ht]
\caption{Dataset descriptions}
\label{supp:dataset-table}
\begin{center}
\begin{small}
\begin{tabular}{l p{4.5cm} l p{4.5cm}}
\toprule
Dataset & Description & Classes & Partition \\
\midrule
FashionMNIST    & Fashion images & 10 & Label skew $Dir \sim (0.5)$  \\
EMNIST & Handwritten digit and character images  & 62 &  Label skew $Dir \sim (0.5)$ \\
CIFAR-10    & Color images & 10 & Label skew $Dir \sim (0.5)$ \\
ISIC-2019    & Skin lesion images taken via dermoscopy & 4 &  Data collected from 4 hospitals     \\
Fed-Heart-Disease    & Tabular data of patients with heart disease & 5 & Data collected from 5 hospitals \\
Sent-140     & Sentiment classification of tweets & 2 & Data collected from 15 users \\
MIMIC-III      & Mortality prediction using patient admission note & 2  & Data grouped by patient admitting diagnosis  \\
\bottomrule
\end{tabular}
\end{small}
\end{center}
\end{table}

\section{Model architectures and Transition point}
Table \ref{supp:layer-table} shows the model architecture and transition point identified by the federation sensitivity metric for each dataset. Of note, 

\begin{table}[ht]
\centering
\caption{Model architectures and transition points identified by federation sensitivity metric.}
\label{supp:layer-table}
\begin{small}
\begin{tabular}{lcc}
\toprule
Dataset & Architecture & Transition Point \\ 
\midrule
FashionMNIST & \{3$\times$Conv-2$\times$FC\} & Conv$_3$ \\
EMNIST & \{3$\times$Conv-2$\times$FC\} & Conv$_3$ \\
CIFAR-10 & \{5$\times$Conv-2$\times$FC\} & Conv$_5$ \\
ISIC-2019 & \{5$\times$Conv-2$\times$FC\} & Conv$_5$ \\
Sent-140 & \{Position$\oplus$Token Emb-Attention$\oplus$Residual-FC$\oplus$Projection-FC\} & FC$_1\oplus$Projection  \\
MIMIC-III & \{Position$\oplus$Token  Emb-Attention$\oplus$Residual-FC$\oplus$Projection-FC\} & FC$_1\oplus$Projection \\
Fed-Heart & \{4$\times$ FC\} & FC$_2$ \\
\bottomrule
\end{tabular}
\\
\footnotesize{*FC = Fully connected layer}
\end{small}
\end{table}

\section{Model training}
\label{supp:model_training}
Table \ref{supp:lr} presents the learning rates utilized for each algorithm and Table \ref{supp:hyperparams} presents the learning rate grid explored, in addition to the loss function, the number of training epochs, and the count of independent training runs. With the exception of pFedME, the AdamW optimizer was used for all algorithms. For pFedME, we adopted the specific optimizer presented by the original authors, which integrates Moreau envelopes into the training process \cite{t2020pfedme}. To account for this, we multiply the learning rate grid tested by 100 as early testing showed the pFedME optimizer demonstrated improved performance with higher learning rates. All experiments were ran on 1 Tesla V100  16GB node.

\begin{table}[ht!]
\caption{Learning rates used}
\label{supp:lr}

\begin{center}
\begin{small}
\begin{tabular}{lccccccc}
\toprule
Algorithm & FMNIST & EMNIST & CIFAR & ISIC & Heart & Sent-140 & MIMIC-III \\ 
\midrule
Local client &  $1\cdot 10^{-3}$ & $1\cdot 10^{-3}$ & $1\cdot 10^{-3}$ & $1\cdot 10^{-3}$ & $5\cdot 10^{-2}$ & $5\cdot 10^{-4}$ & $8\cdot 10^{-5}$  \\
FedAvg &  $5\cdot 10^{-4}$ & $1\cdot 10^{-3}$ & $1\cdot 10^{-3}$ & $1\cdot 10^{-3}$ & $1\cdot 10^{-1}$ & $1\cdot 10^{-3}$ & $5\cdot 10^{-4}$ \\
FedProx &  $5\cdot 10^{-4}$ & $5\cdot 10^{-3}$ & $1\cdot 10^{-3}$ & $1\cdot 10^{-3}$ & $5\cdot 10^{-2}$ & $1\cdot 10^{-3}$ & $5\cdot 10^{-4}$ \\
pFedMe & $5\cdot 10^{-2}$ & $5\cdot 10^{-2}$ & $5\cdot 10^{-2}$ & $5\cdot 10^{-3}$ & $1\cdot 10^{-1}$ & $1\cdot 10^{-2}$ & $1\cdot 10^{-3}$ \\
Ditto &  $5\cdot 10^{-4}$ & $1\cdot 10^{-3}$ & $1\cdot 10^{-3}$ & $1\cdot 10^{-3}$ & $5\cdot 10^{-2}$ & $1\cdot 10^{-3}$ & $5\cdot 10^{-4}$ \\
LocalAdaptation &  $5\cdot 10^{-4}$ & $1\cdot 10^{-3}$ & $1\cdot 10^{-3}$ & $1\cdot 10^{-3}$ & $1\cdot 10^{-2}$ & $1\cdot 10^{-3}$ & $5\cdot 10^{-4}$ \\
BABU &  $1\cdot 10^{-3}$ & $1\cdot 10^{-3}$ & $5\cdot 10^{-4}$ & $1\cdot 10^{-3}$ & $1\cdot 10^{-1}$ & $8\cdot 10^{-5}$ & $5\cdot 10^{-4}$ \\
PLayer-FL &  $1\cdot 10^{-3}$ & $1\cdot 10^{-3}$ & $1\cdot 10^{-3}$ & $5\cdot 10^{-4}$ & $1\cdot 10^{-2}$ & $8\cdot 10^{-5}$ & $3\cdot 10^{-4}$ \\
PLayer-FL-1 & $1\cdot 10^{-3}$ & $5\cdot 10^{-4}$ & $1\cdot 10^{-3}$ & $5\cdot 10^{-4}$ & $5\cdot 10^{-2}$ & $8\cdot 10^{-5}$ & $8\cdot 10^{-5}$ \\
PLayer-FL+1 & $1\cdot 10^{-3}$ & $1\cdot 10^{-3}$ & $5\cdot 10^{-4}$ & $1\cdot 10^{-3}$ & $5\cdot 10^{-2}$ & $1\cdot 10^{-4}$ & $5\cdot 10^{-4}$ \\
\bottomrule
\end{tabular}
\end{small}
\end{center}
\end{table}

\begin{table}
\centering
\caption{Learning rate grid, loss function, number of epochs and number of runs for each dataset}
\label{supp:hyperparams}
\begin{small}
\begin{tabular}{l p{6cm} ccc}
\toprule
Dataset & Learning rate grid & Loss & Epochs & Runs\\ 
\midrule
FashionMNIST & $1\cdot10^{-3}$, $5\cdot10^{-4}$, $1\cdot10^{-4}$, $8\cdot10^{-5}$ & Cross Entropy & 75 & 10\\
EMNIST & $5\cdot10^{-3}$, $1\cdot10^{-3}$, $5\cdot10^{-4}$, $1\cdot10^{-4}$, $8\cdot10^{-5}$ &Cross Entropy& 75 & 10\\
CIFAR-10 & $5\cdot10^{-3}$, $1\cdot10^{-3}$, $5\cdot10^{-4}$, $1\cdot10^{-4}$ & Cross Entropy& 50 & 10\\
ISIC-2019 & $1\cdot10^{-3}$, $5\cdot10^{-3}$, $1\cdot10^{-4}$ & Multiclass Focal & 50 & 3\\
Sent-140 & $1\cdot10^{-3}$, $5\cdot10^{-4}$, $1\cdot10^{-4}$, $8\cdot10^{-5}$ & Cross Entropy & 75 & 20\\
Heart & $5\cdot10^{-1}$, $1\cdot10^{-1}$, $5\cdot10^{-2}$, $1\cdot10^{-2}$, $5\cdot10^{-3}$ & Multiclass Focal& 50 & 50 \\
MIMIC-III & $5\cdot10^{-4}$, $1\cdot10^{-4}$, $3\cdot10^{-4}$, $8\cdot10^{-5}$ & Multiclass Focal& 50 & 10\\
\bottomrule
\end{tabular}
\end{small}
\end{table}

\section{Comparing models trained on non-IID data}

\subsection{Gradient variance}
\begin{figure}[ht]

\begin{center}

% First four subfigures
\makebox[\linewidth][c]{
\begin{minipage}{.24\linewidth}
  \centering
  \raisebox{-\height}{\includegraphics[width=\linewidth]{figures/FMNIST_Gradient_Variance_first-1.png}}
  \makebox[\linewidth][c]{\scriptsize\textit{(A) FashionMNIST}}  % Subfigure label
\end{minipage}%
\begin{minipage}{.24\linewidth}
  \centering
  \raisebox{-\height}{\includegraphics[width=\linewidth]{figures/EMNIST_Gradient_Variance_first-1.png}}
  \makebox[\linewidth][c]{\scriptsize\textit{(B) EMNIST}}  % Subfigure label
\end{minipage}%
\begin{minipage}{.24\linewidth}
  \centering
  \raisebox{-\height}{\includegraphics[width=\linewidth]{figures/CIFAR_Gradient_Variance_first-1.png}}
  \makebox[\linewidth][c]{\scriptsize\textit{(C) CIFAR-10}}  % Subfigure label
\end{minipage}%
\begin{minipage}{.24\linewidth}
  \centering
  \raisebox{-\height}{\includegraphics[width=\linewidth]{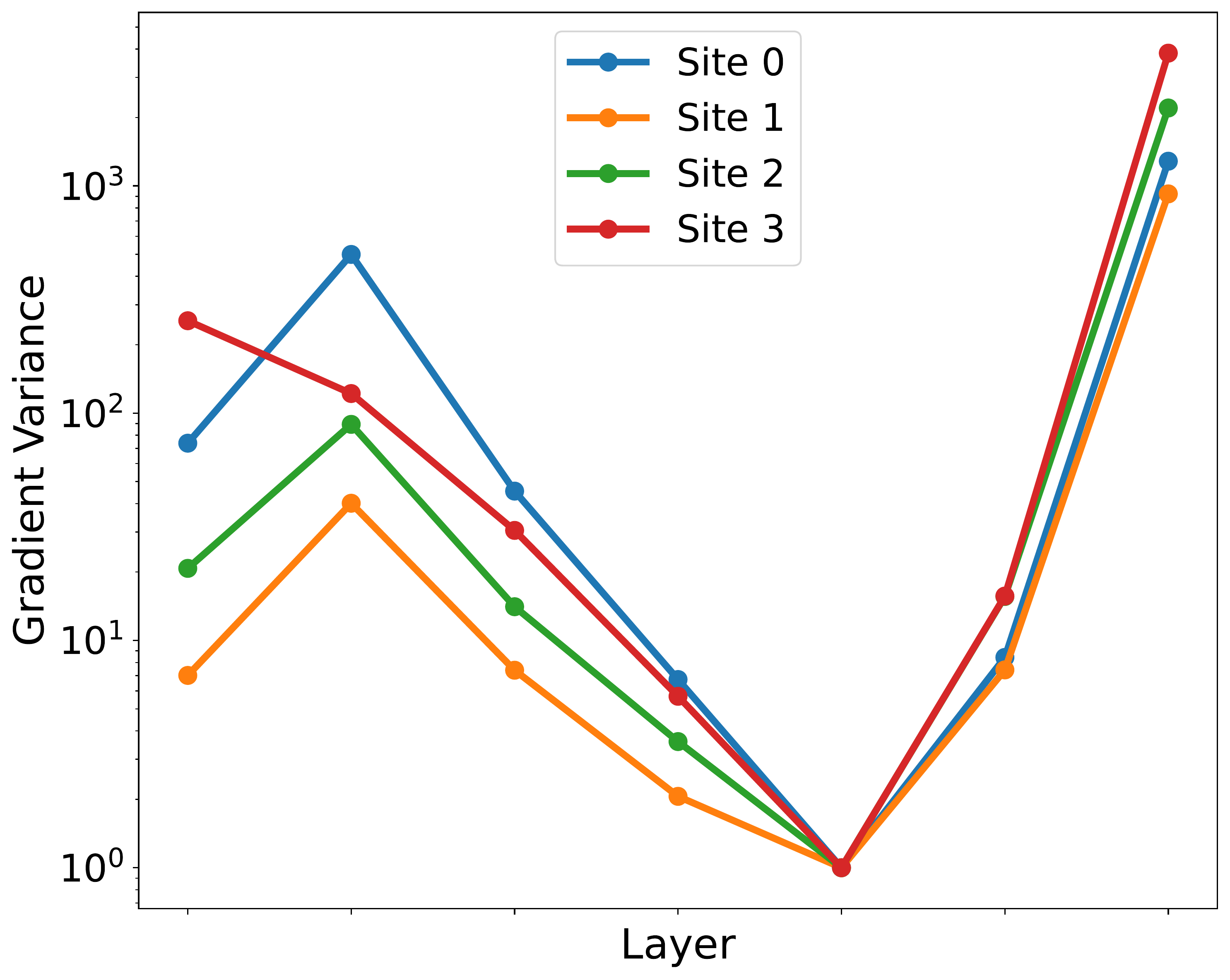}}
  \makebox[\linewidth][c]{\scriptsize\textit{(D) ISIC-2019}}  % Subfigure label
\end{minipage}
}
\makebox[\linewidth][c]{
\begin{minipage}{.24\linewidth}
  \centering
  \raisebox{-\height}{\includegraphics[width=\linewidth]{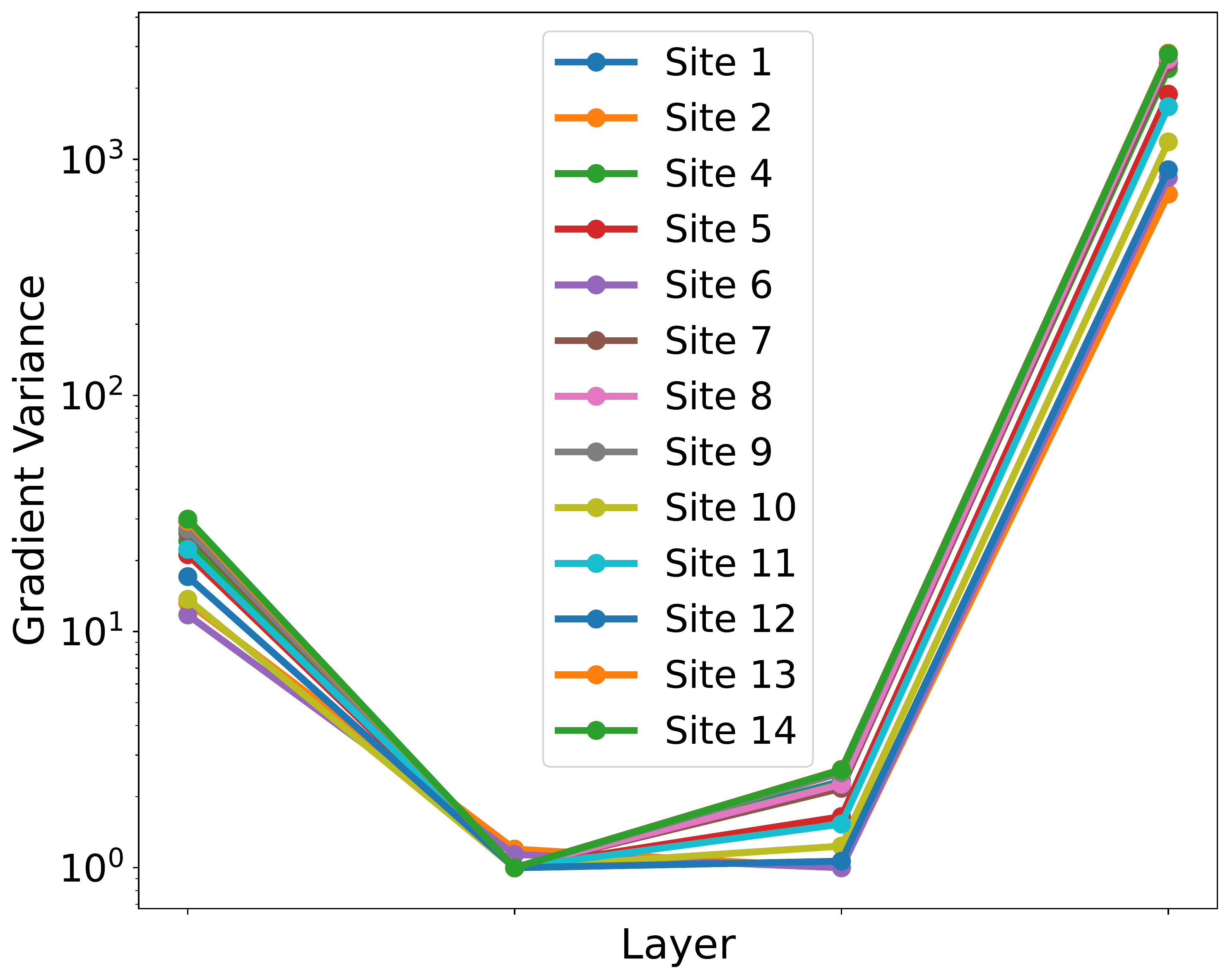}}
  \makebox[\linewidth][c]{\scriptsize\textit{(E) Sent-140}}  % Subfigure label
\end{minipage}%
\begin{minipage}{.24\linewidth}
  \centering
  \raisebox{-\height}{\includegraphics[width=\linewidth]{figures/mimic_Gradient_Variance_first-1.png}}
  \makebox[\linewidth][c]{\scriptsize\textit{(F) MIMIC-III}}  % Subfigure label
\end{minipage}%
\begin{minipage}{.24\linewidth}
  \centering
  \raisebox{-\height}{\includegraphics[width=\linewidth]{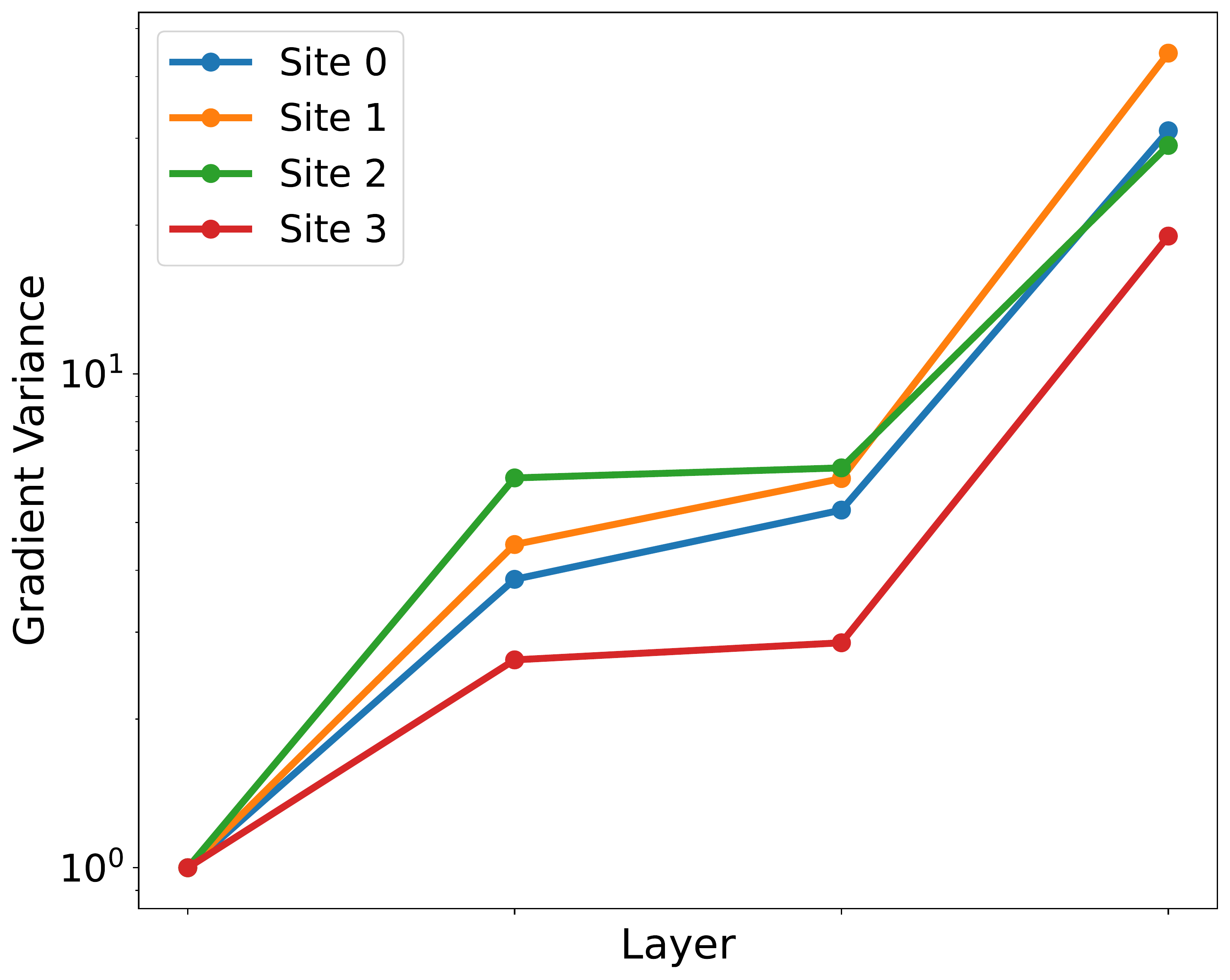}}
  \makebox[\linewidth][c]{\scriptsize\textit{(G) Fed-Heart-Disease}}  % Subfigure label
\end{minipage}
}

\caption{\textbf{Layer gradient variance after one epoch}. All models identically initialized and independently trained on non-IID data.}
\label{sfig:grad_var_first}
\end{center}

\end{figure}

\begin{figure}[ht]

\begin{center}

% First four subfigures
\makebox[\linewidth][c]{
\begin{minipage}{.24\linewidth}
  \centering
  \raisebox{-\height}{\includegraphics[width=\linewidth]{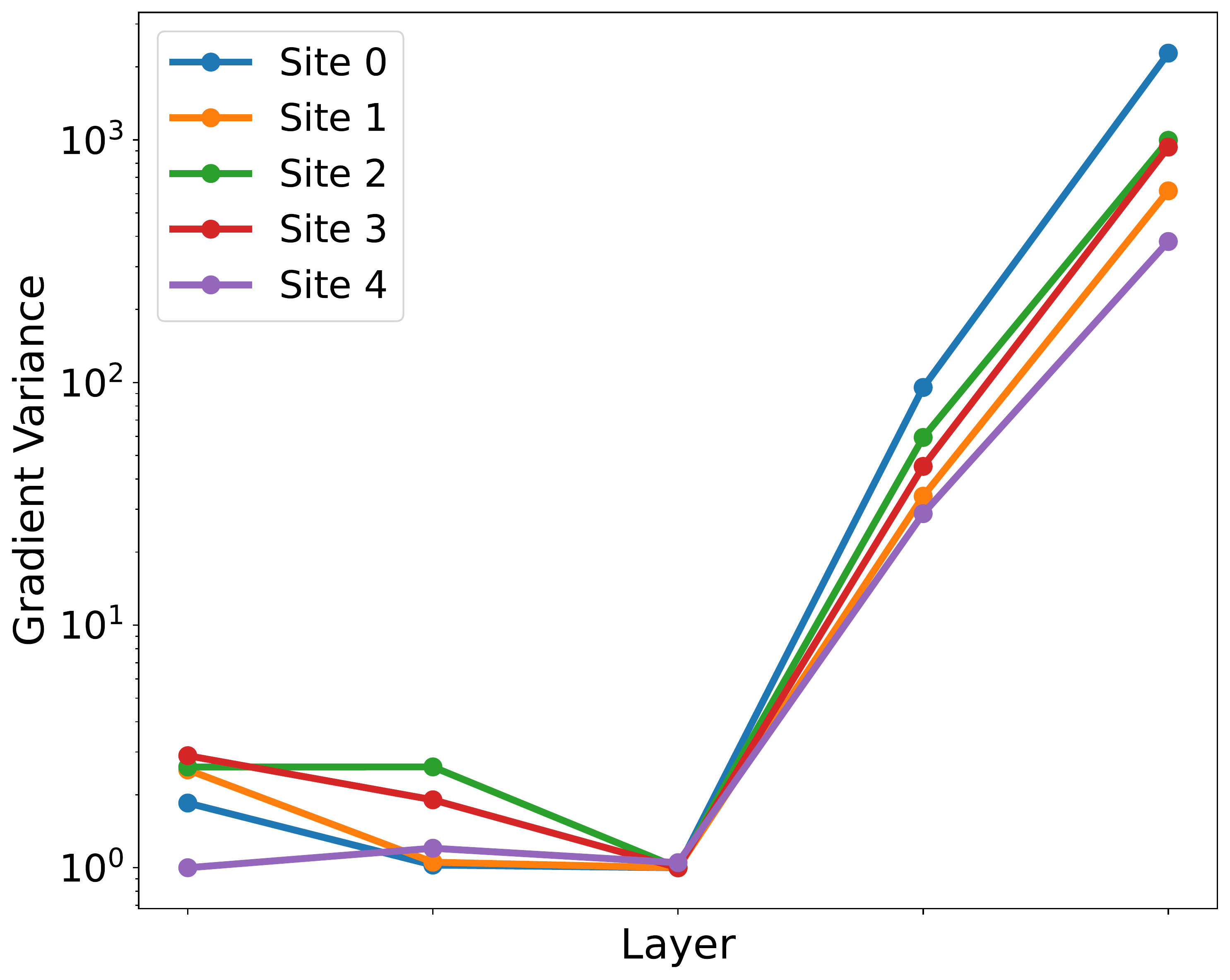}}
  \makebox[\linewidth][c]{\scriptsize\textit{(A) FashionMNIST}}  % Subfigure label
\end{minipage}%
\begin{minipage}{.24\linewidth}
  \centering
  \raisebox{-\height}{\includegraphics[width=\linewidth]{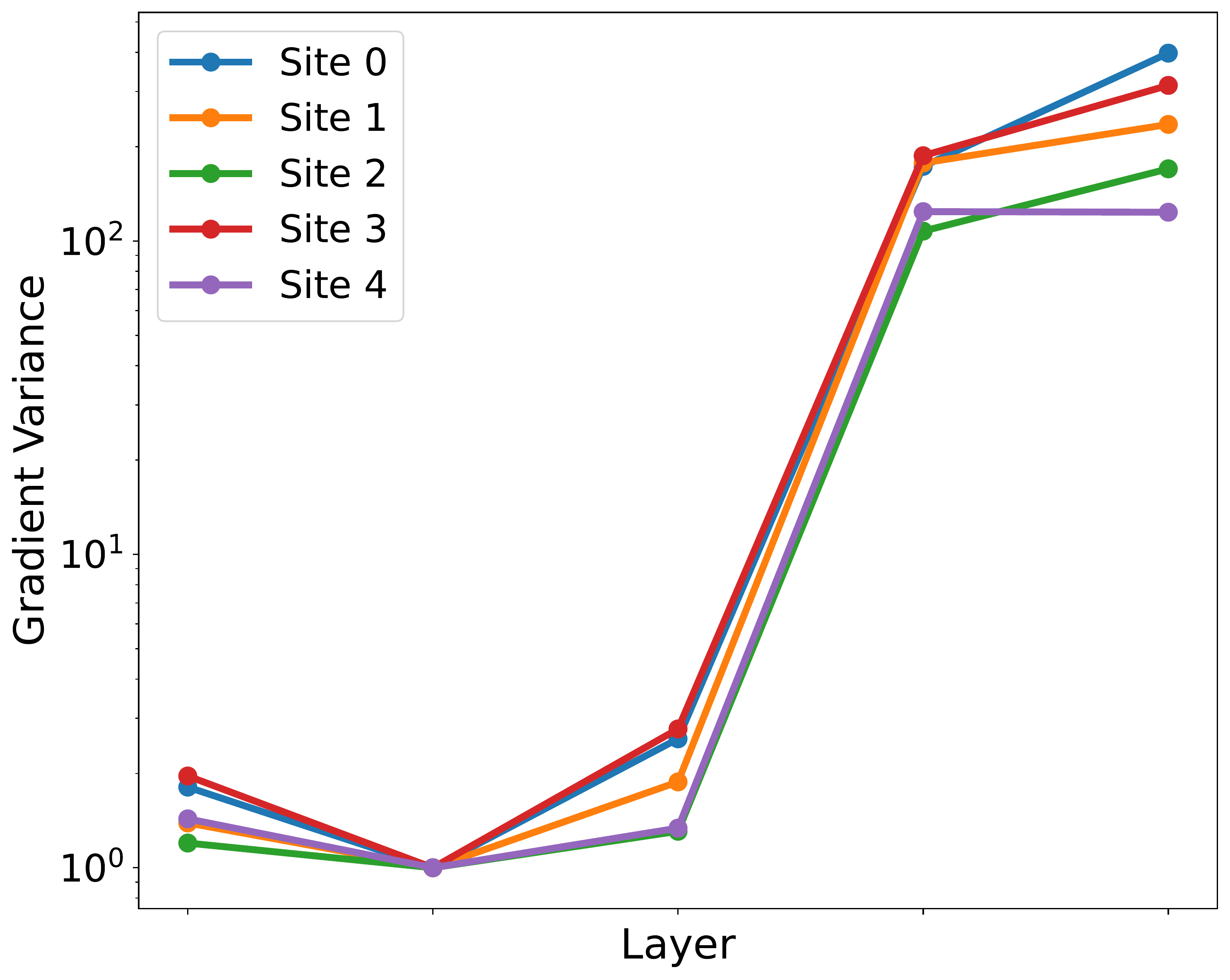}}
  \makebox[\linewidth][c]{\scriptsize\textit{(B) EMNIST}}  % Subfigure label
\end{minipage}%
\begin{minipage}{.24\linewidth}
  \centering
  \raisebox{-\height}{\includegraphics[width=\linewidth]{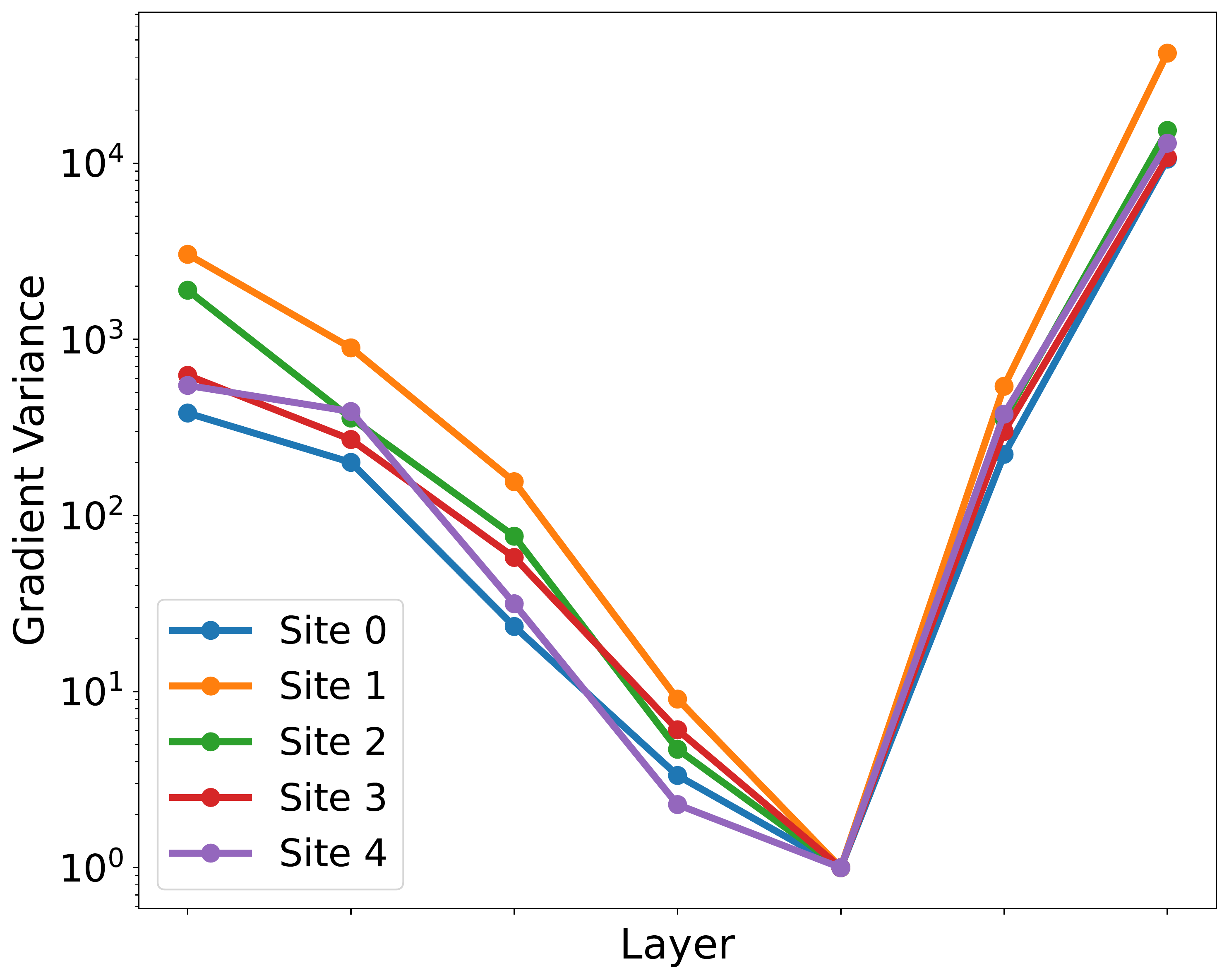}}
  \makebox[\linewidth][c]{\scriptsize\textit{(C) CIFAR-10}}  % Subfigure label
\end{minipage}%
\begin{minipage}{.24\linewidth}
  \centering
  \raisebox{-\height}{\includegraphics[width=\linewidth]{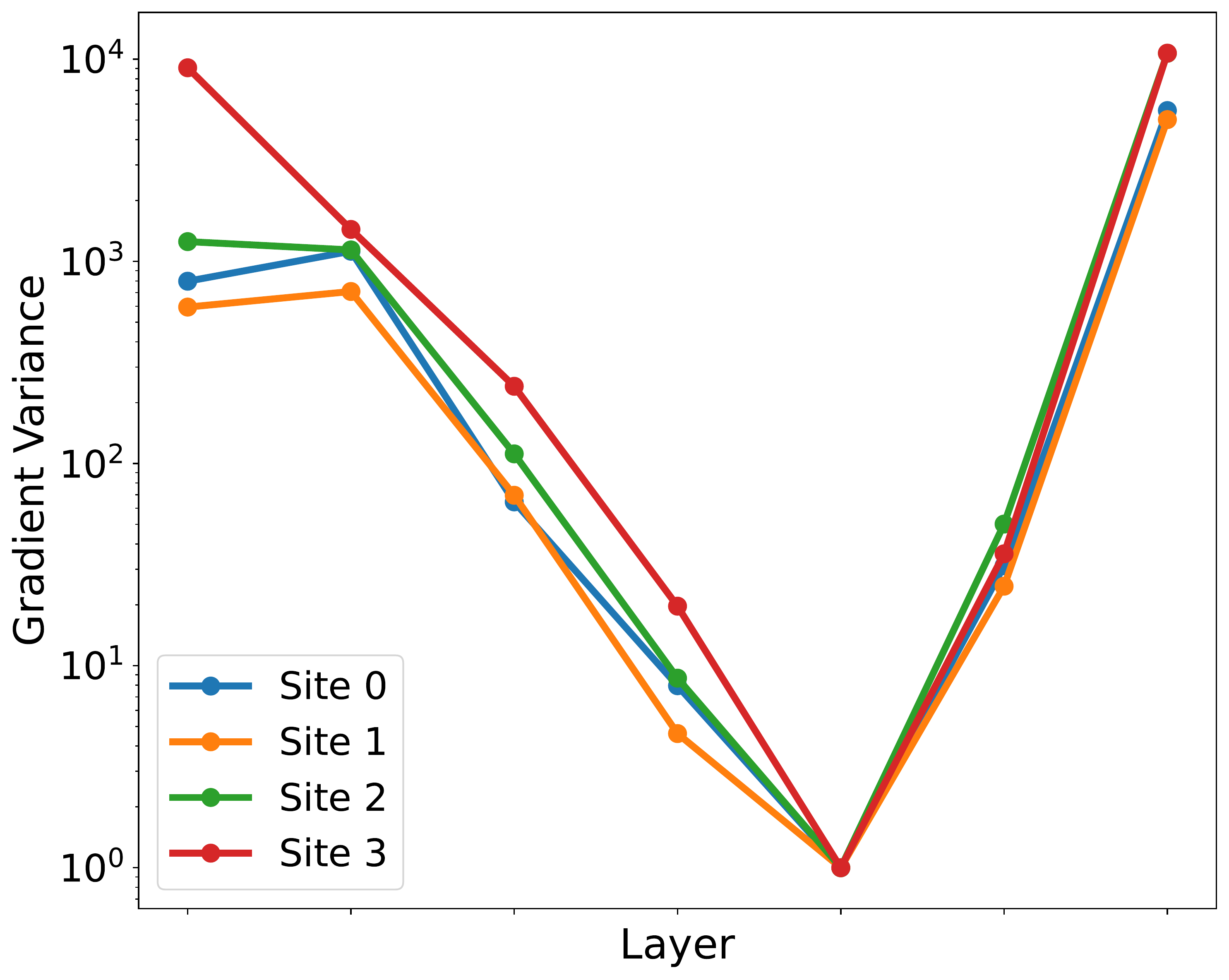}}
  \makebox[\linewidth][c]{\scriptsize\textit{(D) ISIC-2019}}  % Subfigure label
\end{minipage}
}
\makebox[\linewidth][c]{
\begin{minipage}{.24\linewidth}
  \centering
  \raisebox{-\height}{\includegraphics[width=\linewidth]{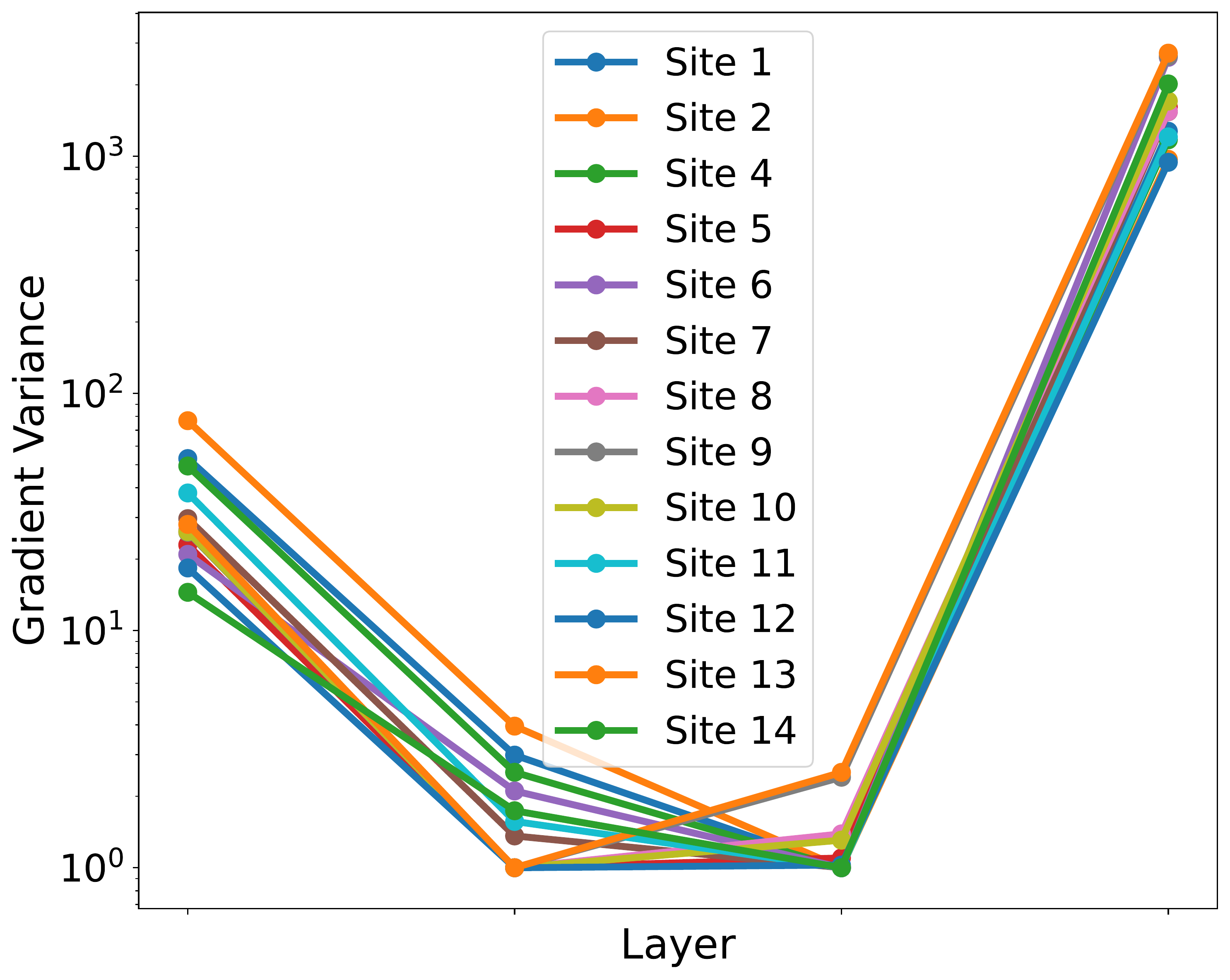}}
  \makebox[\linewidth][c]{\scriptsize\textit{(E) Sent-140}}  % Subfigure label
\end{minipage}%
\begin{minipage}{.24\linewidth}
  \centering
  \raisebox{-\height}{\includegraphics[width=\linewidth]{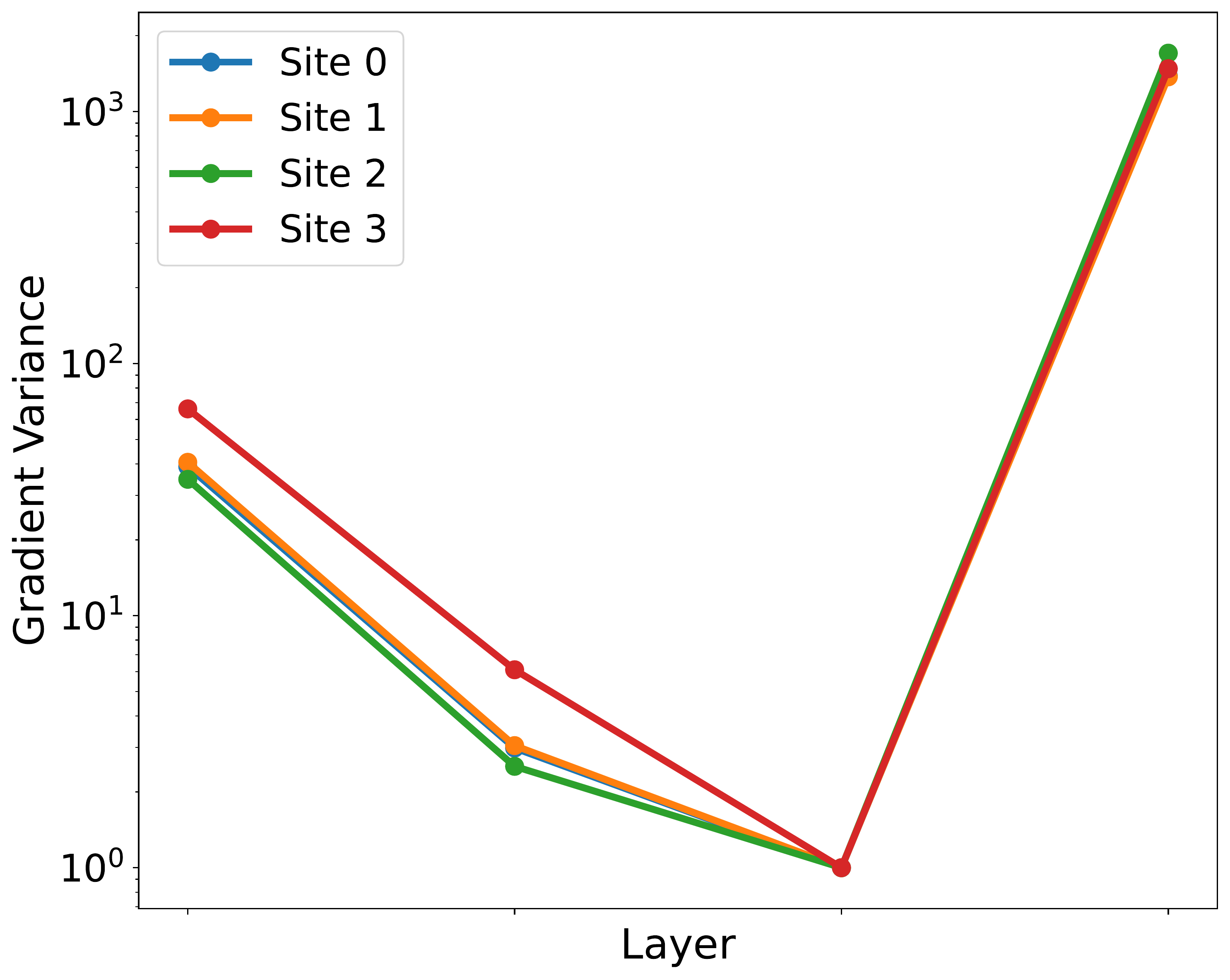}}
  \makebox[\linewidth][c]{\scriptsize\textit{(F) MIMIC-III}}  % Subfigure label
\end{minipage}%
\begin{minipage}{.24\linewidth}
  \centering
  \raisebox{-\height}{\includegraphics[width=\linewidth]{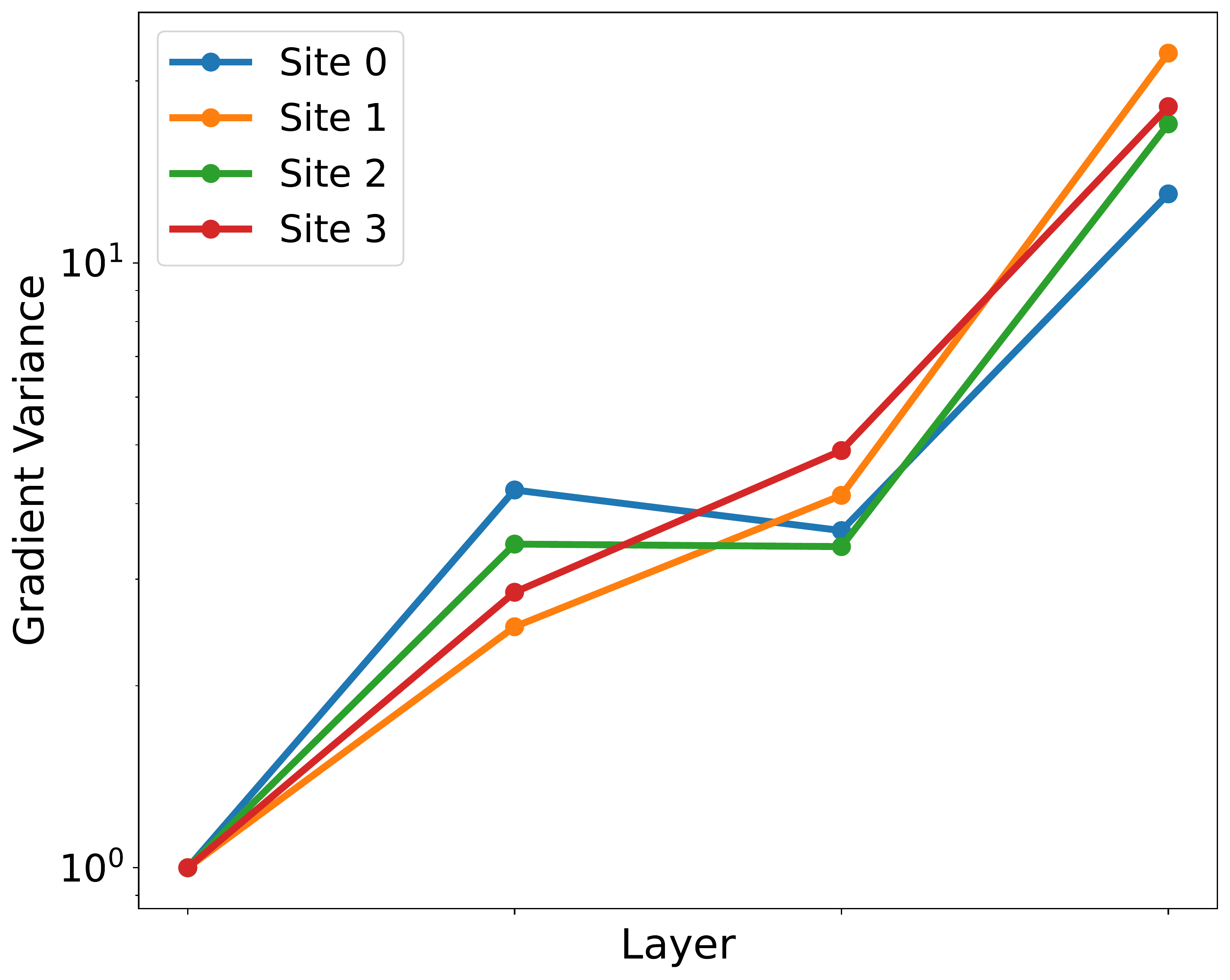}}
  \makebox[\linewidth][c]{\scriptsize\textit{(G) Fed-Heart-Disease}}  % Subfigure label
\end{minipage}
}

\caption{\textbf{Layer gradient variance for final models}. All models identically initialized and independently trained on non-IID data.}
\label{sfig:grad_var_best}
\end{center}

\end{figure}

\begin{figure}[ht]

\begin{center}

% First four subfigures
\makebox[\linewidth][c]{
\begin{minipage}{.24\linewidth}
  \centering
  \raisebox{-\height}{\includegraphics[width=\linewidth]{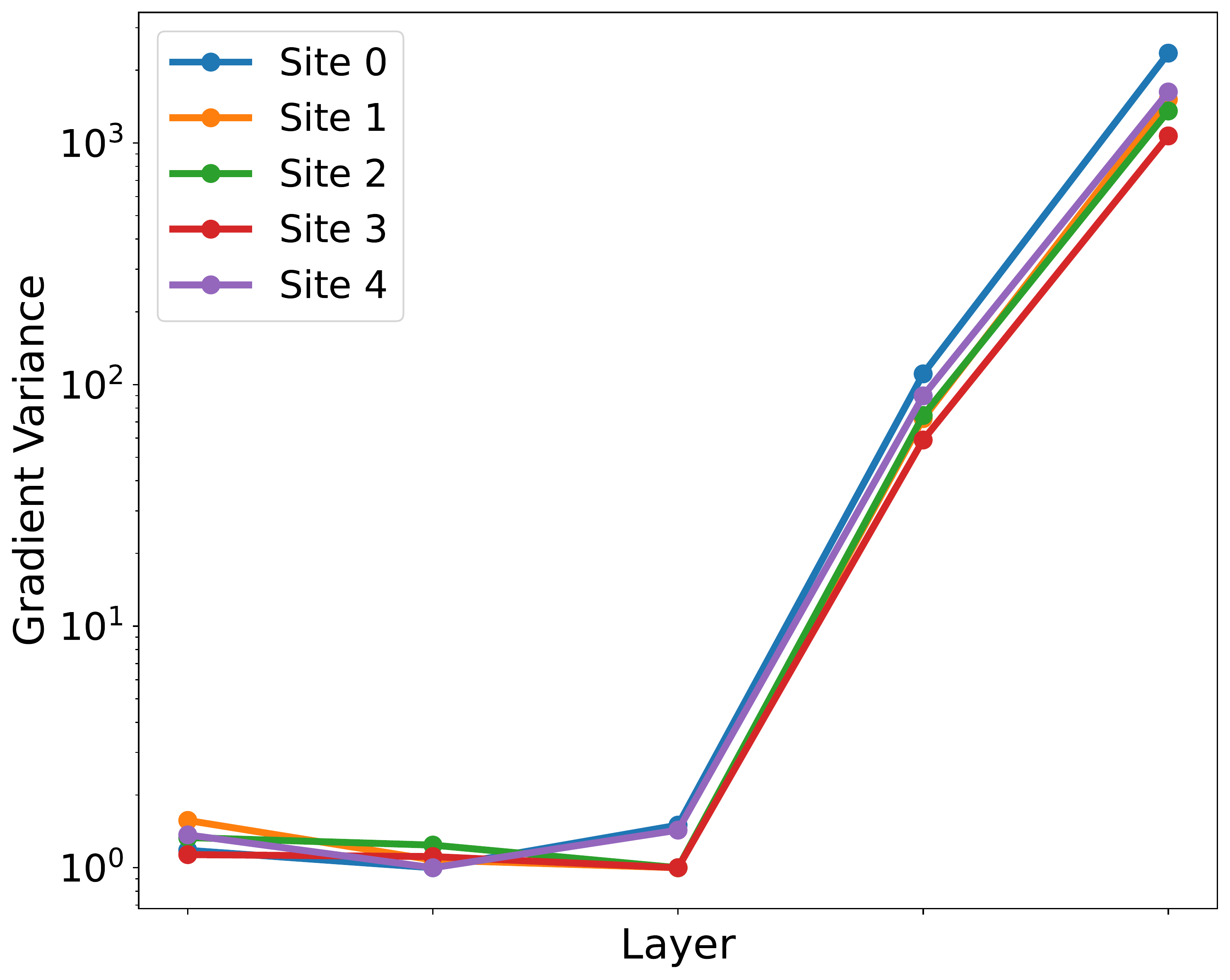}}
  \makebox[\linewidth][c]{\scriptsize\textit{(A) FashionMNIST}}  % Subfigure label
\end{minipage}%
\begin{minipage}{.24\linewidth}
  \centering
  \raisebox{-\height}{\includegraphics[width=\linewidth]{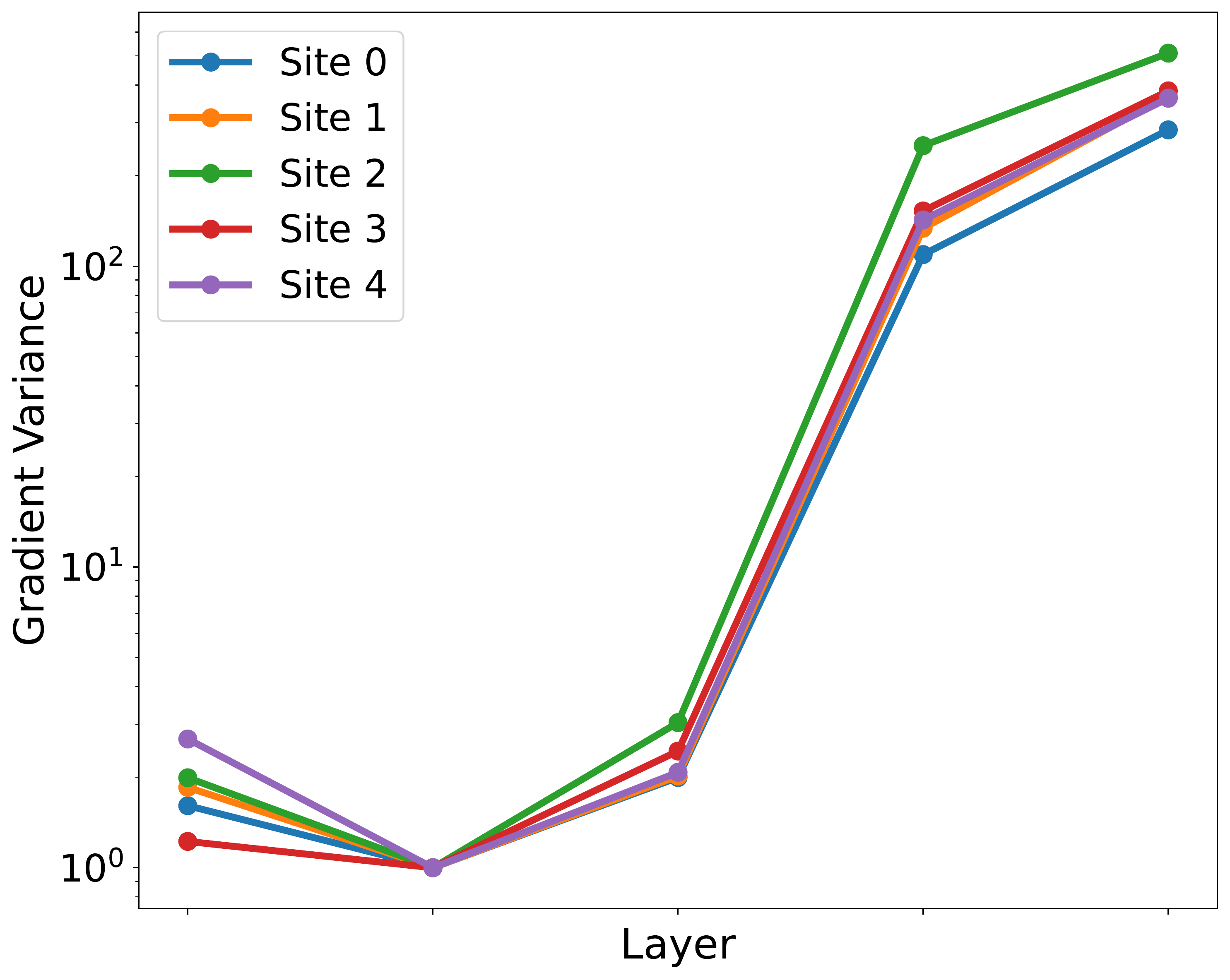}}
  \makebox[\linewidth][c]{\scriptsize\textit{(B) EMNIST}}  % Subfigure label
\end{minipage}%
\begin{minipage}{.24\linewidth}
  \centering
  \raisebox{-\height}{\includegraphics[width=\linewidth]{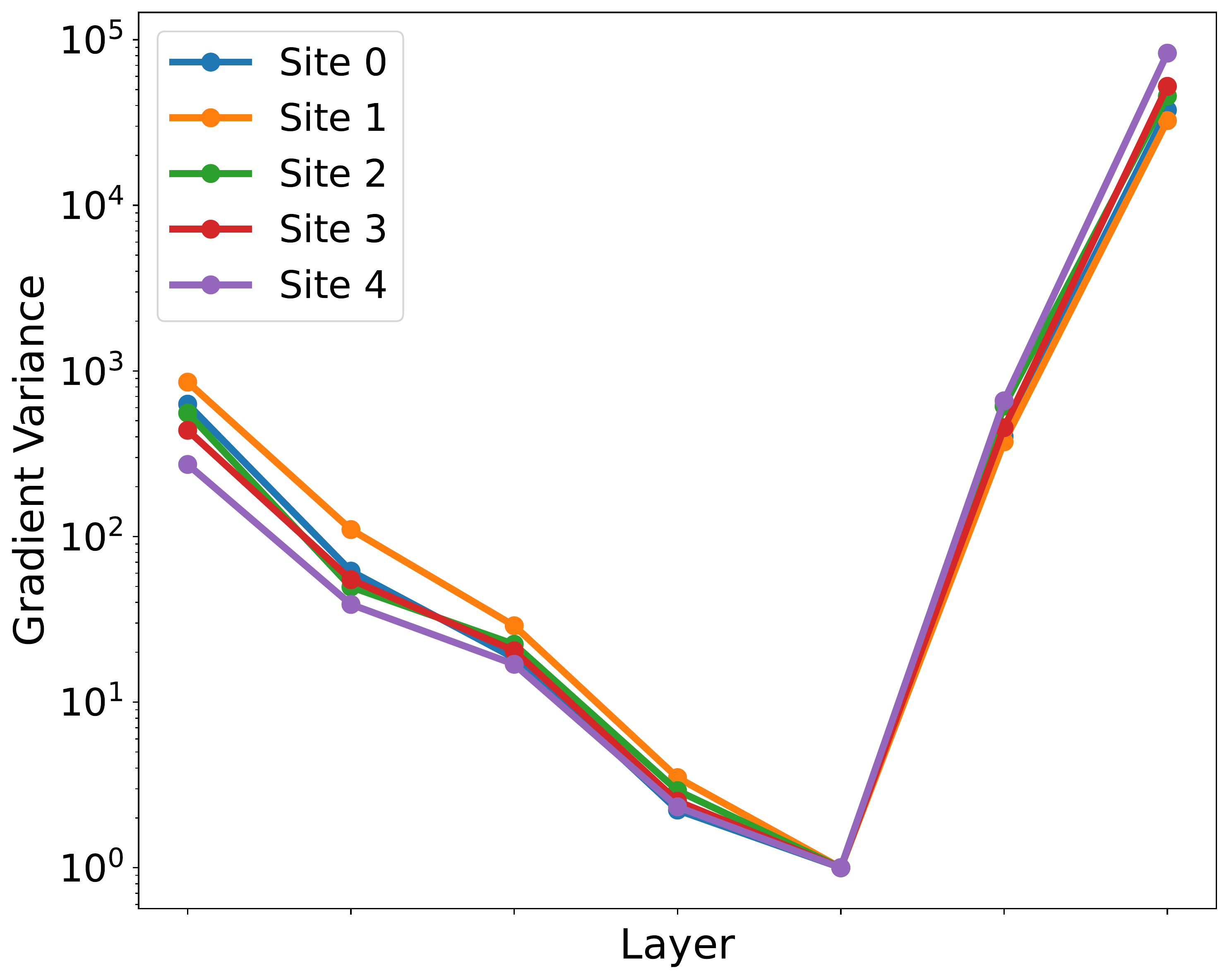}}
  \makebox[\linewidth][c]{\scriptsize\textit{(C) CIFAR-10}}  % Subfigure label
\end{minipage}%
\begin{minipage}{.24\linewidth}
  \centering
  \raisebox{-\height}{\includegraphics[width=\linewidth]{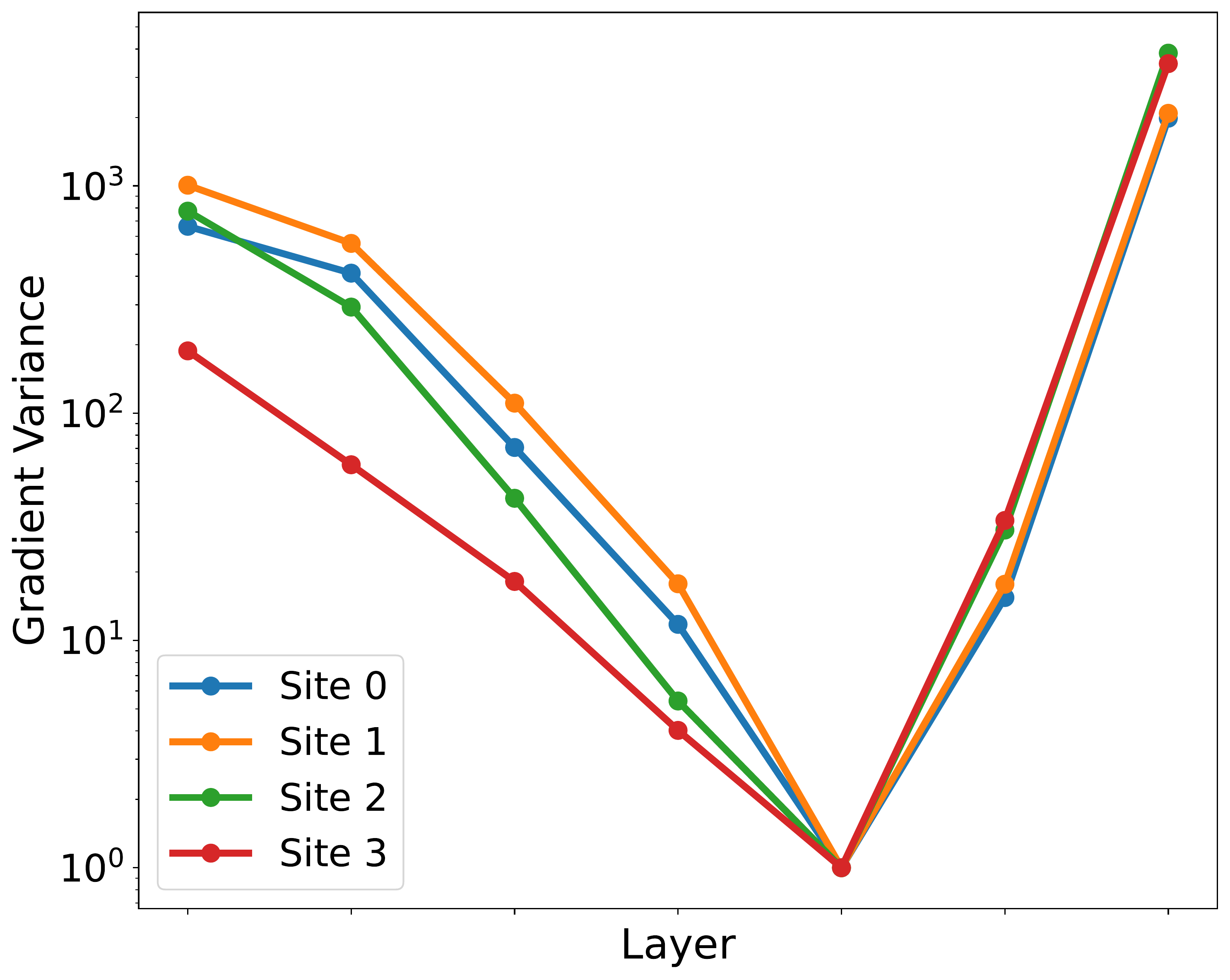}}
  \makebox[\linewidth][c]{\scriptsize\textit{(D) ISIC-2019}}  % Subfigure label
\end{minipage}
}
\makebox[\linewidth][c]{
\begin{minipage}{.24\linewidth}
  \centering
  \raisebox{-\height}{\includegraphics[width=\linewidth]{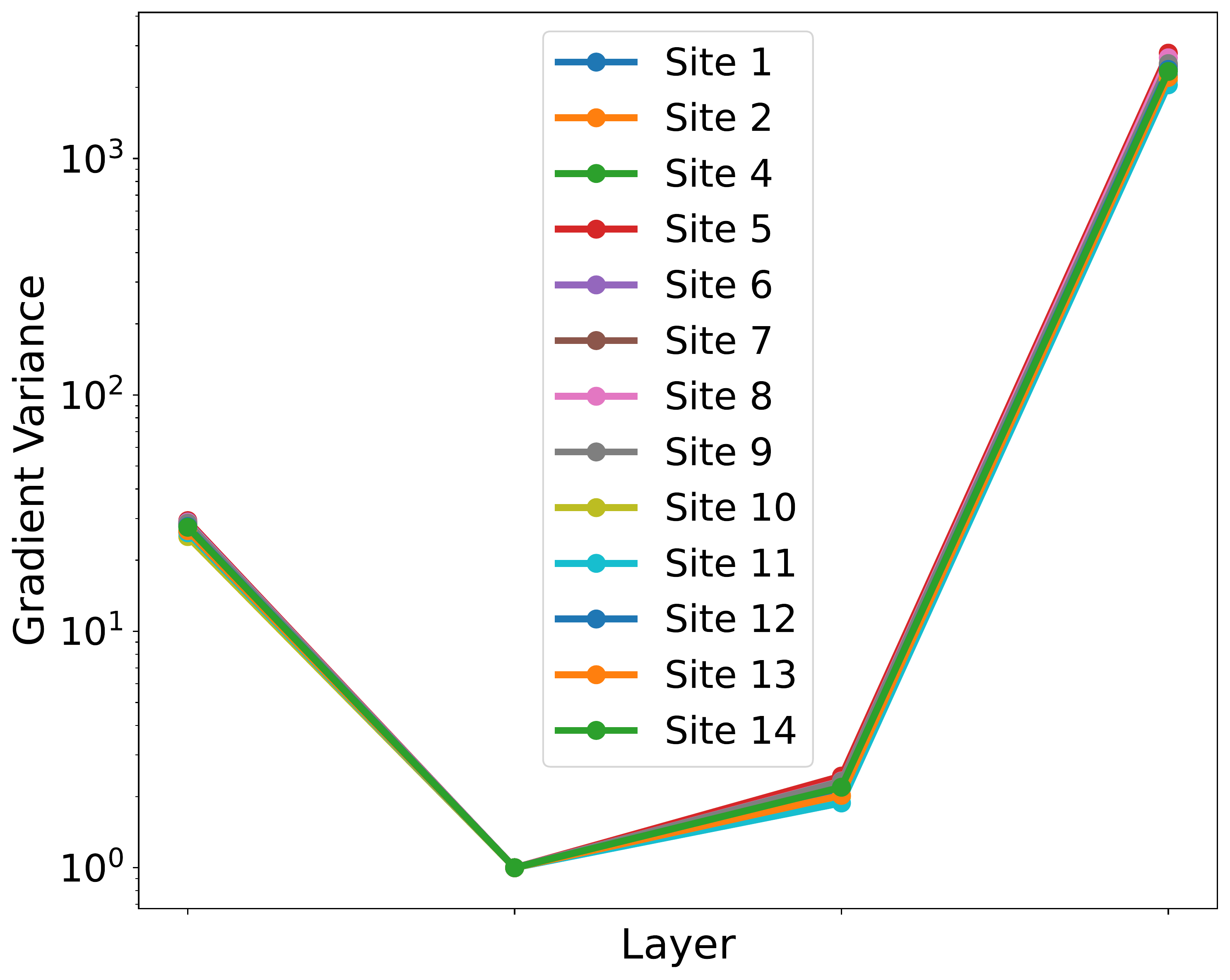}}
  \makebox[\linewidth][c]{\scriptsize\textit{(E) Sent-140}}  % Subfigure label
\end{minipage}%
\begin{minipage}{.24\linewidth}
  \centering
  \raisebox{-\height}{\includegraphics[width=\linewidth]{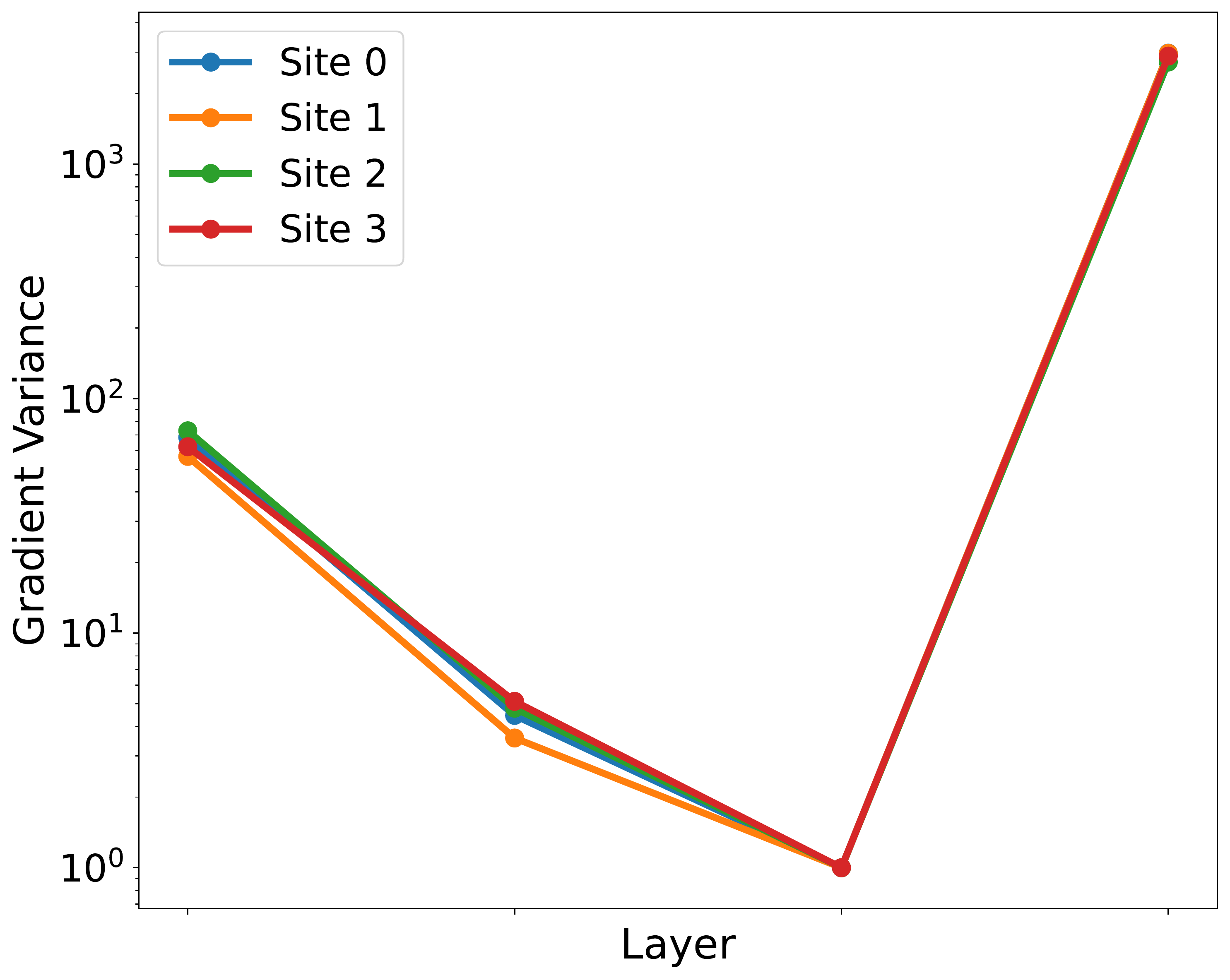}}
  \makebox[\linewidth][c]{\scriptsize\textit{(F) MIMIC-III}}  % Subfigure label
\end{minipage}%
\begin{minipage}{.24\linewidth}
  \centering
  \raisebox{-\height}{\includegraphics[width=\linewidth]{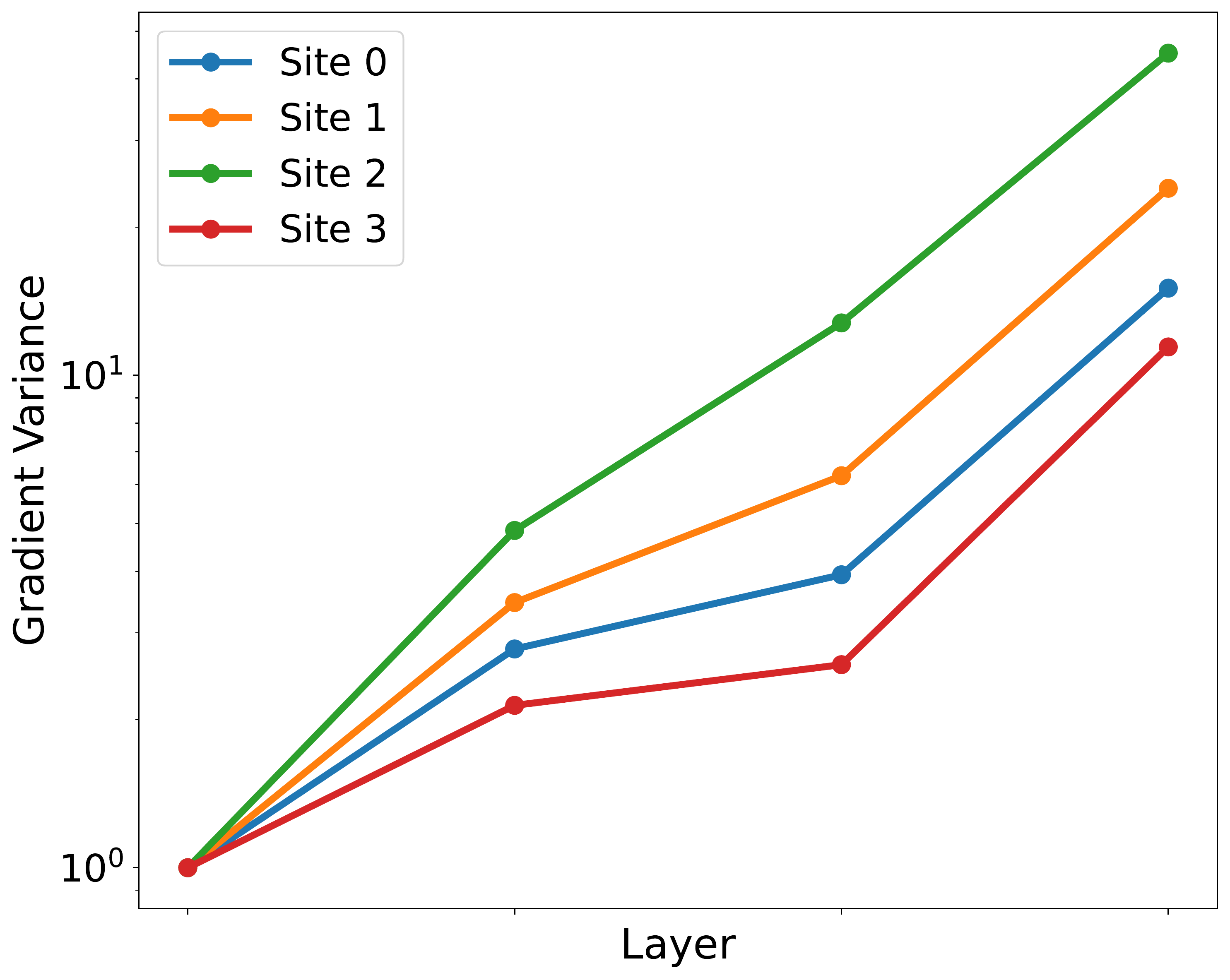}}
  \makebox[\linewidth][c]{\scriptsize\textit{(G) Fed-Heart-Disease}}  % Subfigure label
\end{minipage}
}

\caption{\textbf{Layer gradient variance for final models}. Models trained via FL on non-IID data.}
\label{sfig:grad_var_best_fl}
\end{center}

\end{figure}

Figures \ref{sfig:grad_var_first}, \ref{sfig:grad_var_best}, and \ref{sfig:grad_var_best_fl} display the gradient variance across all datasets, corresponding to the models trained for a single epoch, final model after independent training, and final model after FL training, respectively.

\subsection{Hessian eigenvalue sum}

\begin{figure}[ht]

\begin{center}

% First four subfigures
\makebox[\linewidth][c]{
\begin{minipage}{.24\linewidth}
  \centering
  \raisebox{-\height}{\includegraphics[width=\linewidth]{figures/FMNIST_Hessian_EV_sum_first-1.png}}
  \makebox[\linewidth][c]{\scriptsize\textit{(A) FashionMNIST}}  % Subfigure label
\end{minipage}%
\begin{minipage}{.24\linewidth}
  \centering
  \raisebox{-\height}{\includegraphics[width=\linewidth]{figures/EMNIST_Hessian_EV_sum_first-1.png}}
  \makebox[\linewidth][c]{\scriptsize\textit{(B) EMNIST}}  % Subfigure label
\end{minipage}%
\begin{minipage}{.24\linewidth}
  \centering
  \raisebox{-\height}{\includegraphics[width=\linewidth]{figures/CIFAR_Hessian_EV_sum_first-1.png}}
  \makebox[\linewidth][c]{\scriptsize\textit{(C) CIFAR-10}}  % Subfigure label
\end{minipage}%
\begin{minipage}{.24\linewidth}
  \centering
  \raisebox{-\height}{\includegraphics[width=\linewidth]{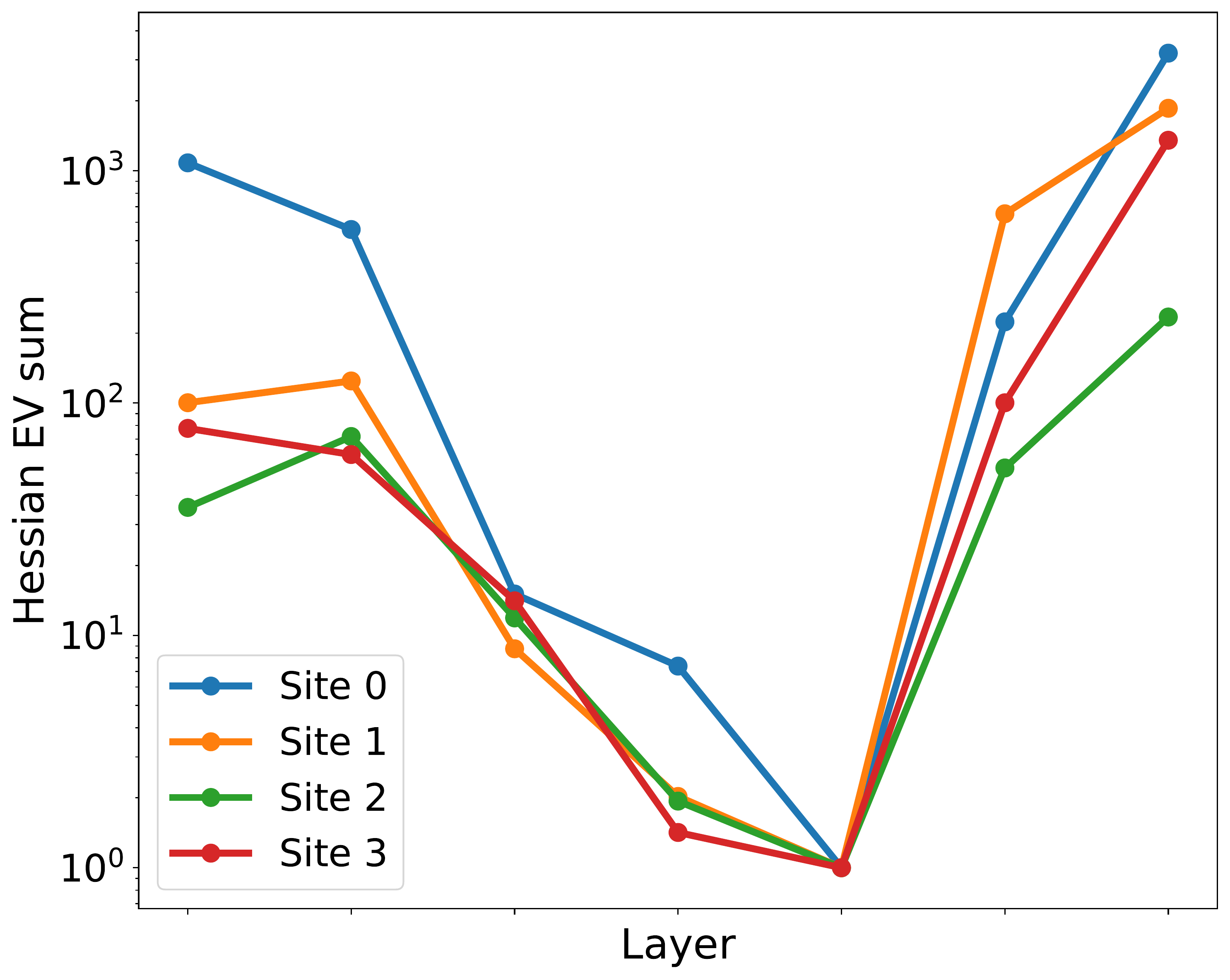}}
  \makebox[\linewidth][c]{\scriptsize\textit{(D) ISIC-2019}}  % Subfigure label
\end{minipage}
}
\makebox[\linewidth][c]{
\begin{minipage}{.24\linewidth}
  \centering
  \raisebox{-\height}{\includegraphics[width=\linewidth]{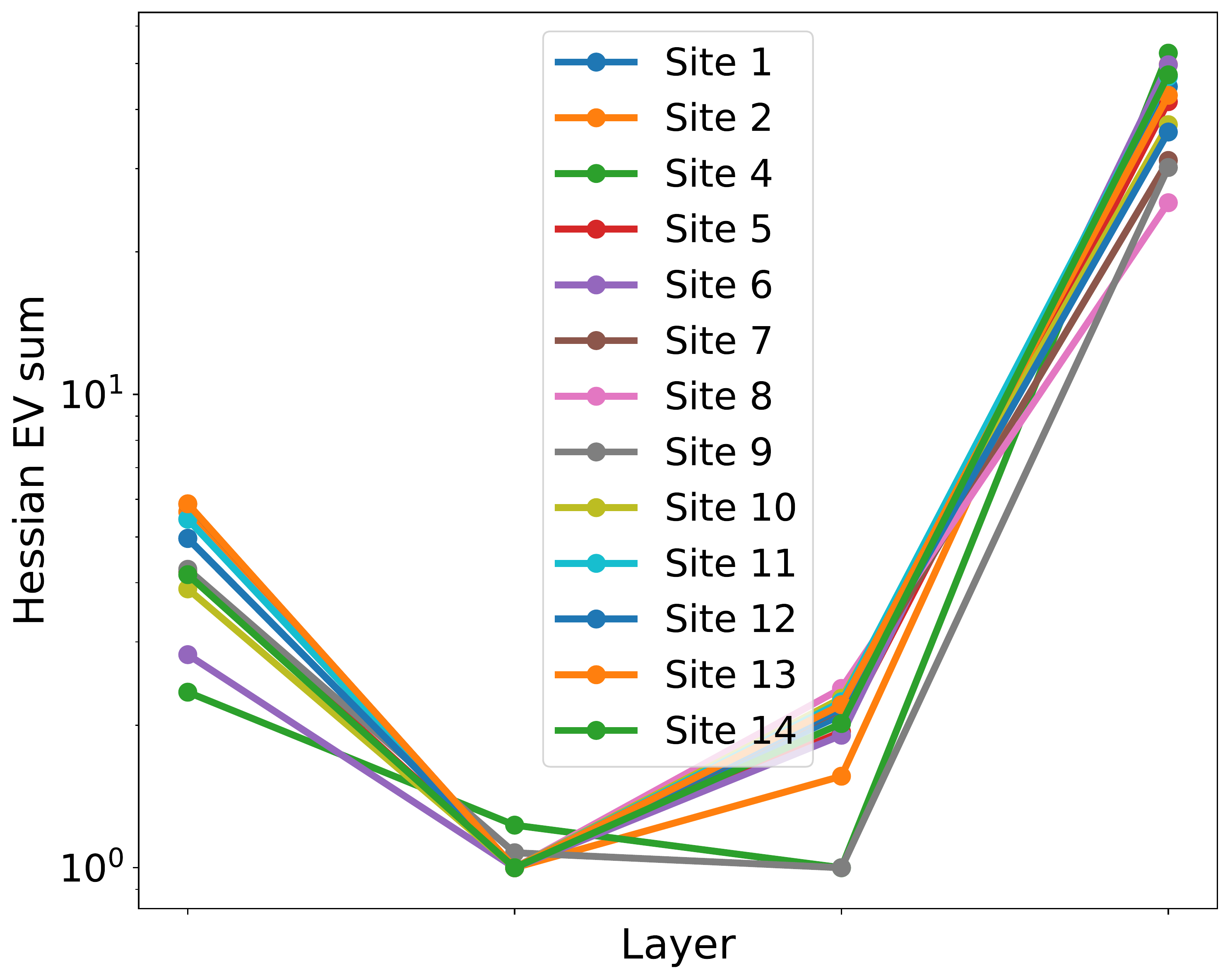}}
  \makebox[\linewidth][c]{\scriptsize\textit{(E) Sent-140}}  % Subfigure label
\end{minipage}%
\begin{minipage}{.24\linewidth}
  \centering
  \raisebox{-\height}{\includegraphics[width=\linewidth]{figures/mimic_Hessian_EV_sum_first-1.png}}
  \makebox[\linewidth][c]{\scriptsize\textit{(F) MIMIC-III}}  % Subfigure label
\end{minipage}%
\begin{minipage}{.24\linewidth}
  \centering
  \raisebox{-\height}{\includegraphics[width=\linewidth]{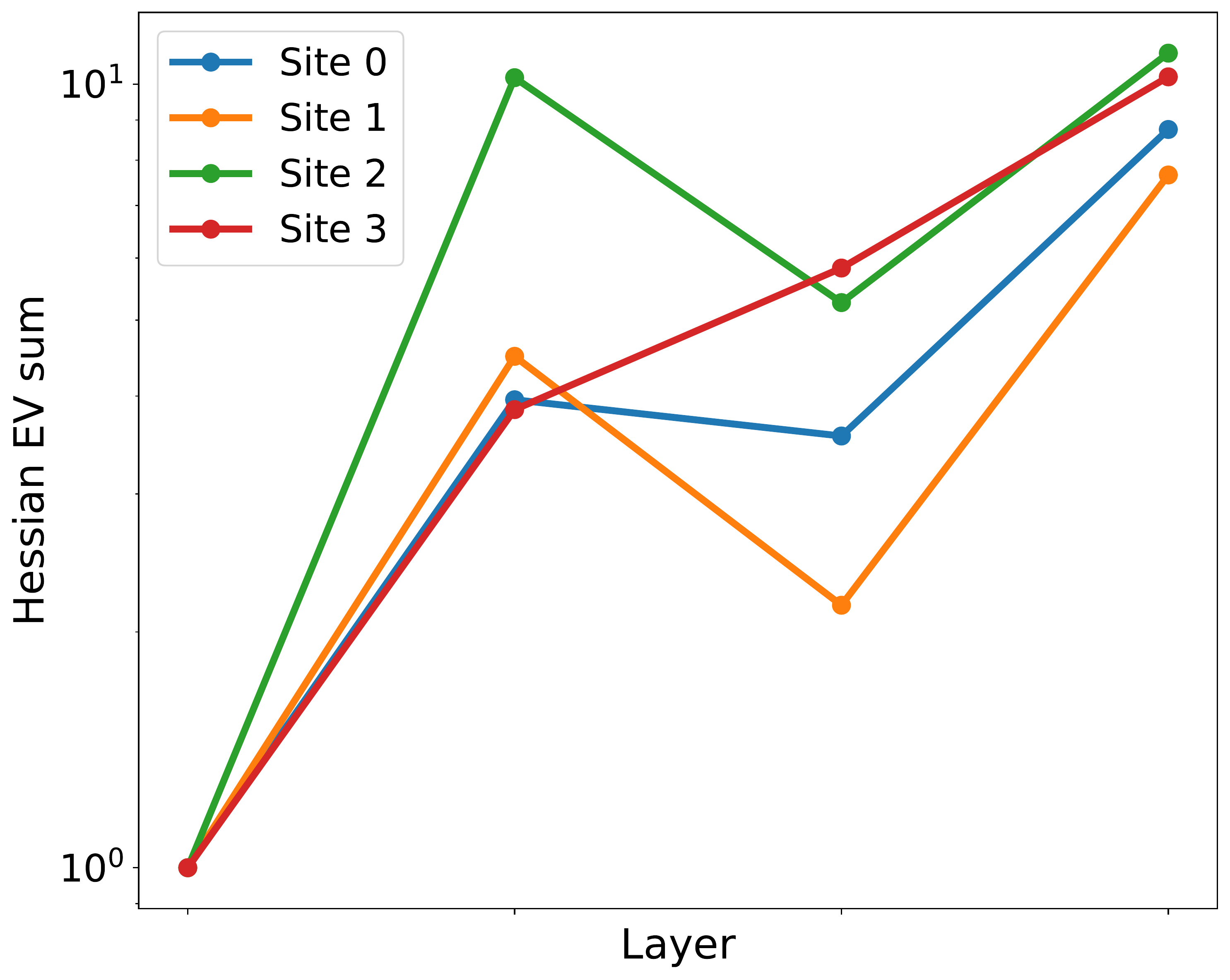}}
  \makebox[\linewidth][c]{\scriptsize\textit{(G) Fed-Heart-Disease}}  % Subfigure label
\end{minipage}
}

\caption{\textbf{Layer hessian eigenvalue sum after one epoch}. All models identically initialized and independently trained on non-IID data.}
\label{sfig:hess_eig_first}
\end{center}
\end{figure}

\begin{figure}[ht]

\begin{center}

% First four subfigures
\makebox[\linewidth][c]{
\begin{minipage}{.24\linewidth}
  \centering
  \raisebox{-\height}{\includegraphics[width=\linewidth]{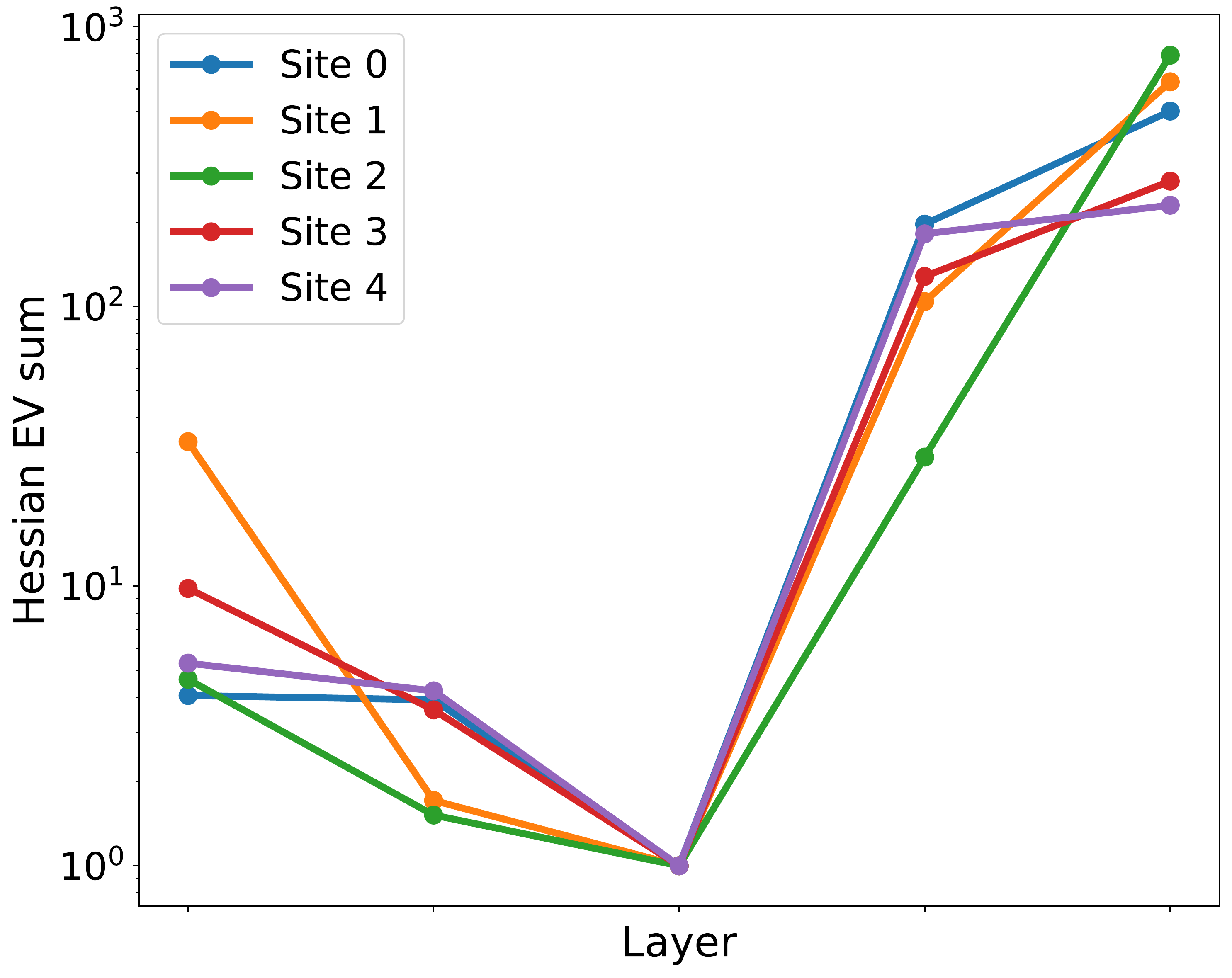}}
  \makebox[\linewidth][c]{\scriptsize\textit{(A) FashionMNIST}}  % Subfigure label
\end{minipage}%
\begin{minipage}{.24\linewidth}
  \centering
  \raisebox{-\height}{\includegraphics[width=\linewidth]{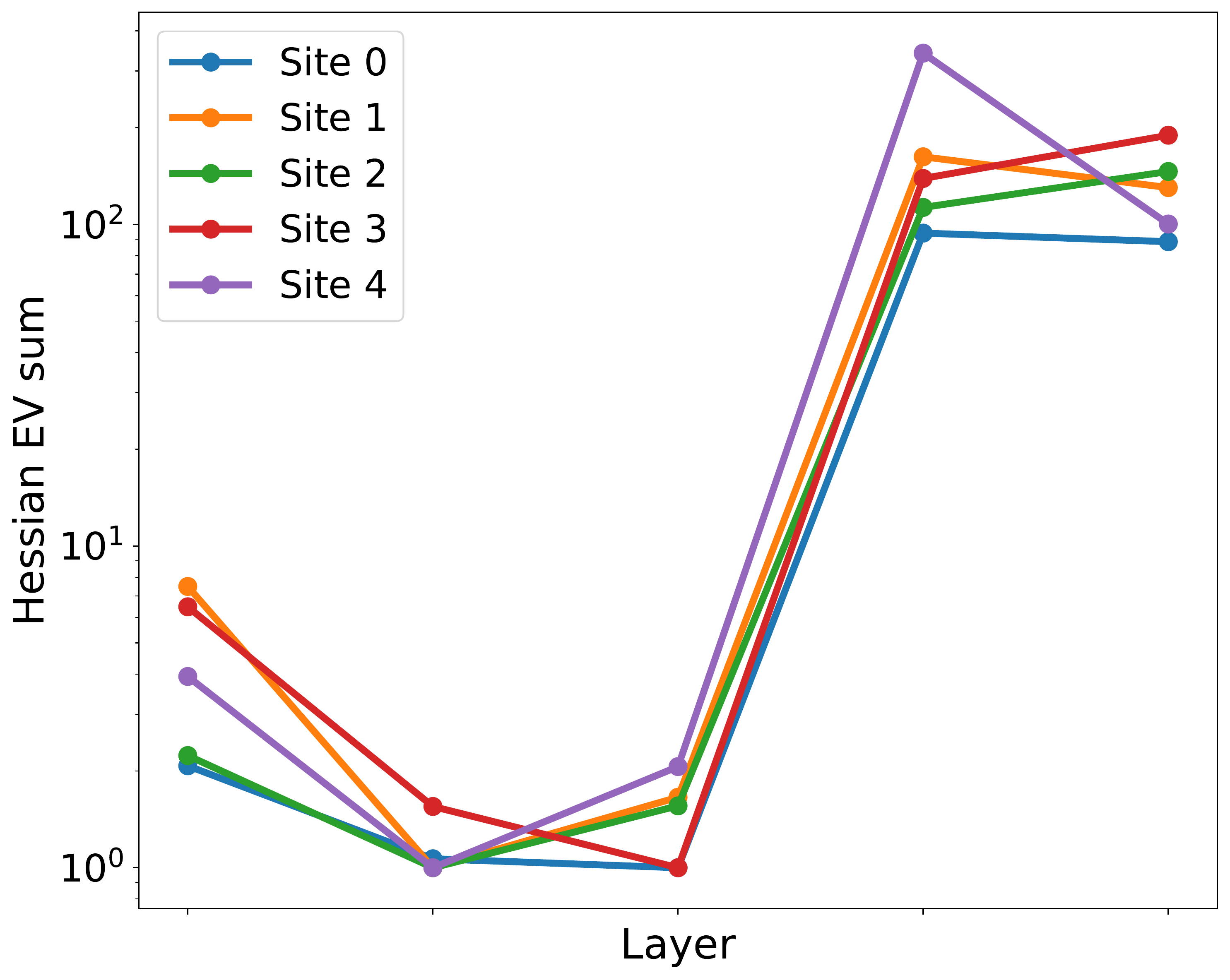}}
  \makebox[\linewidth][c]{\scriptsize\textit{(B) EMNIST}}  % Subfigure label
\end{minipage}%
\begin{minipage}{.24\linewidth}
  \centering
  \raisebox{-\height}{\includegraphics[width=\linewidth]{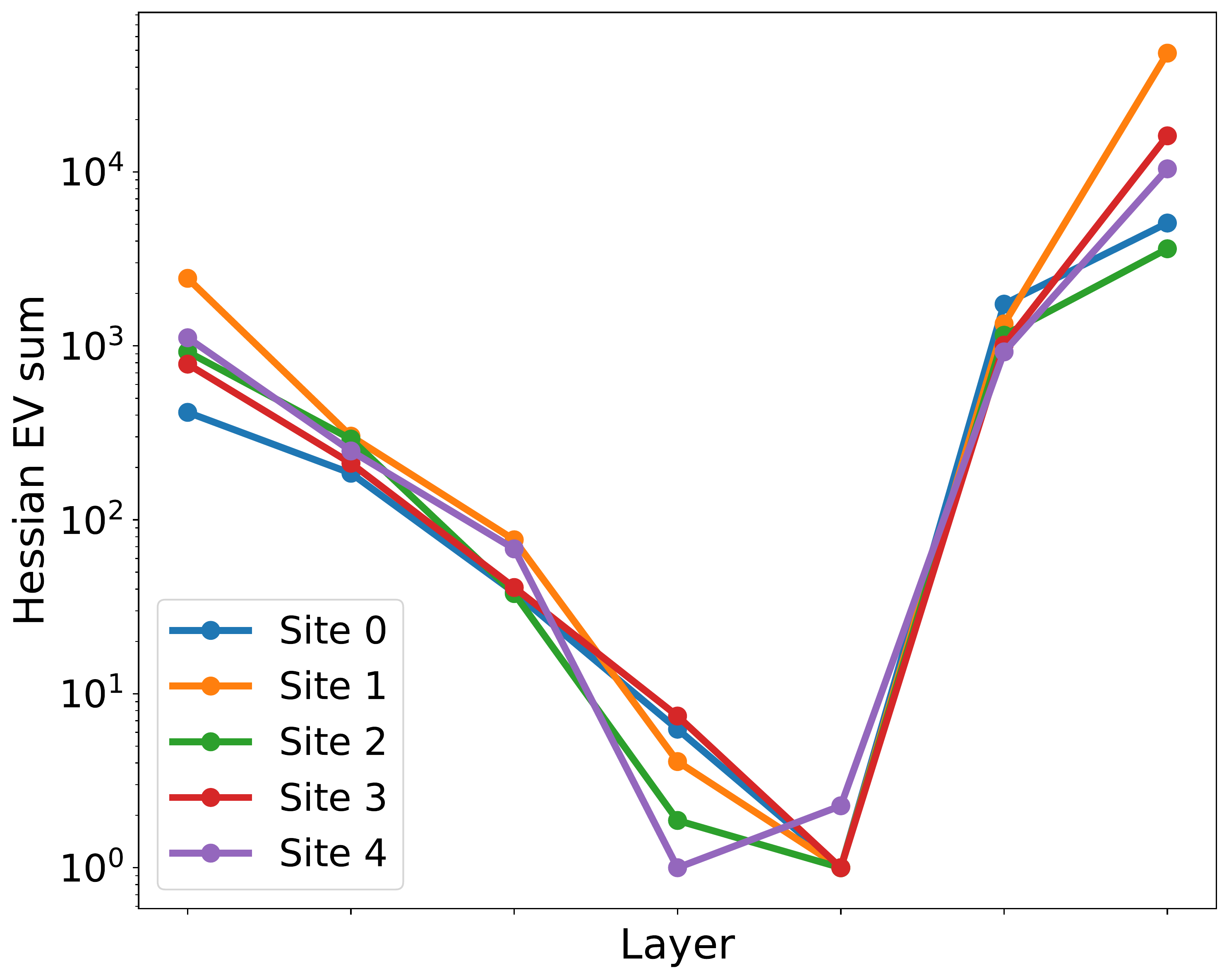}}
  \makebox[\linewidth][c]{\scriptsize\textit{(C) CIFAR-10}}  % Subfigure label
\end{minipage}%
\begin{minipage}{.24\linewidth}
  \centering
  \raisebox{-\height}{\includegraphics[width=\linewidth]{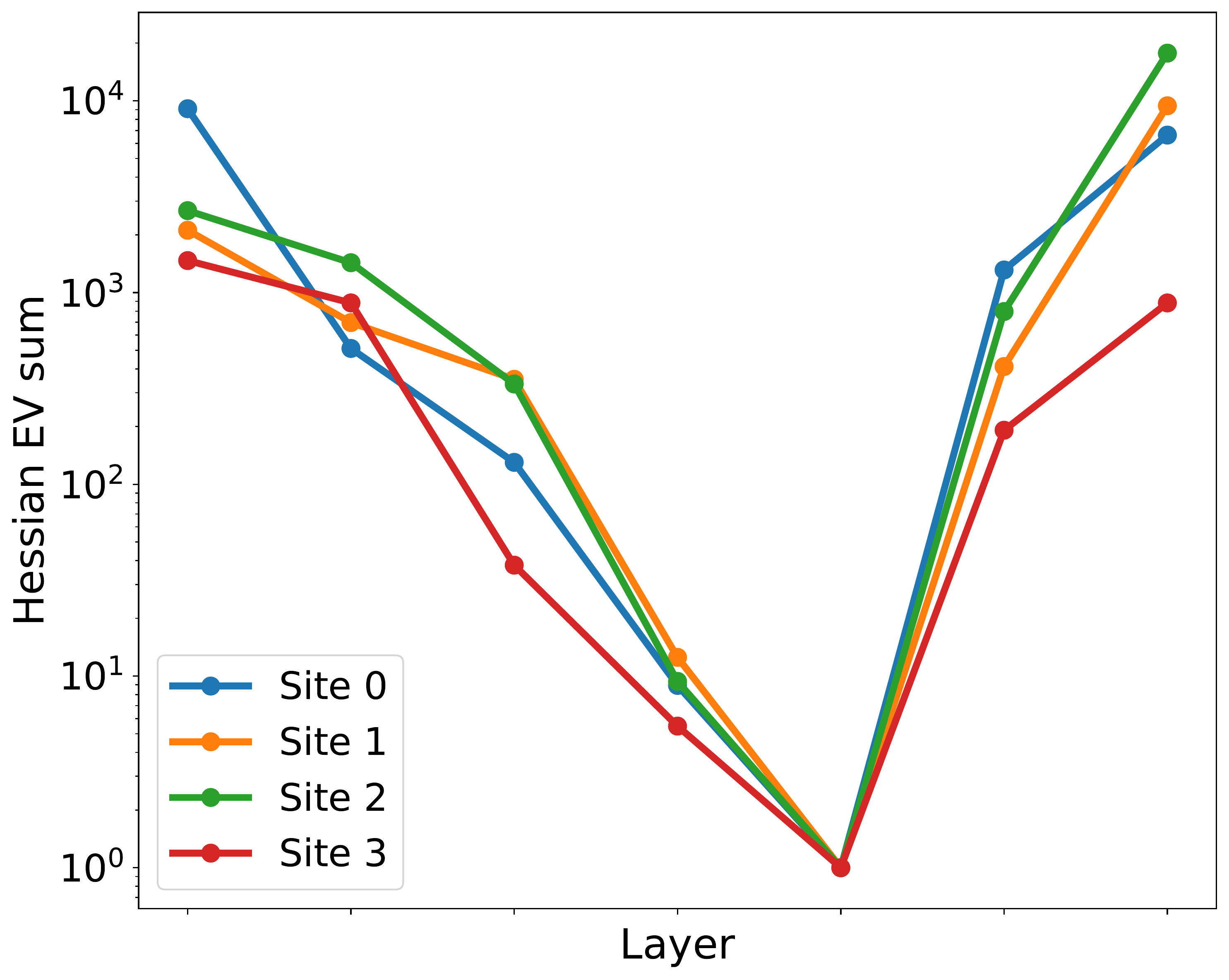}}
  \makebox[\linewidth][c]{\scriptsize\textit{(D) ISIC-2019}}  % Subfigure label
\end{minipage}
}
\makebox[\linewidth][c]{
\begin{minipage}{.24\linewidth}
  \centering
  \raisebox{-\height}{\includegraphics[width=\linewidth]{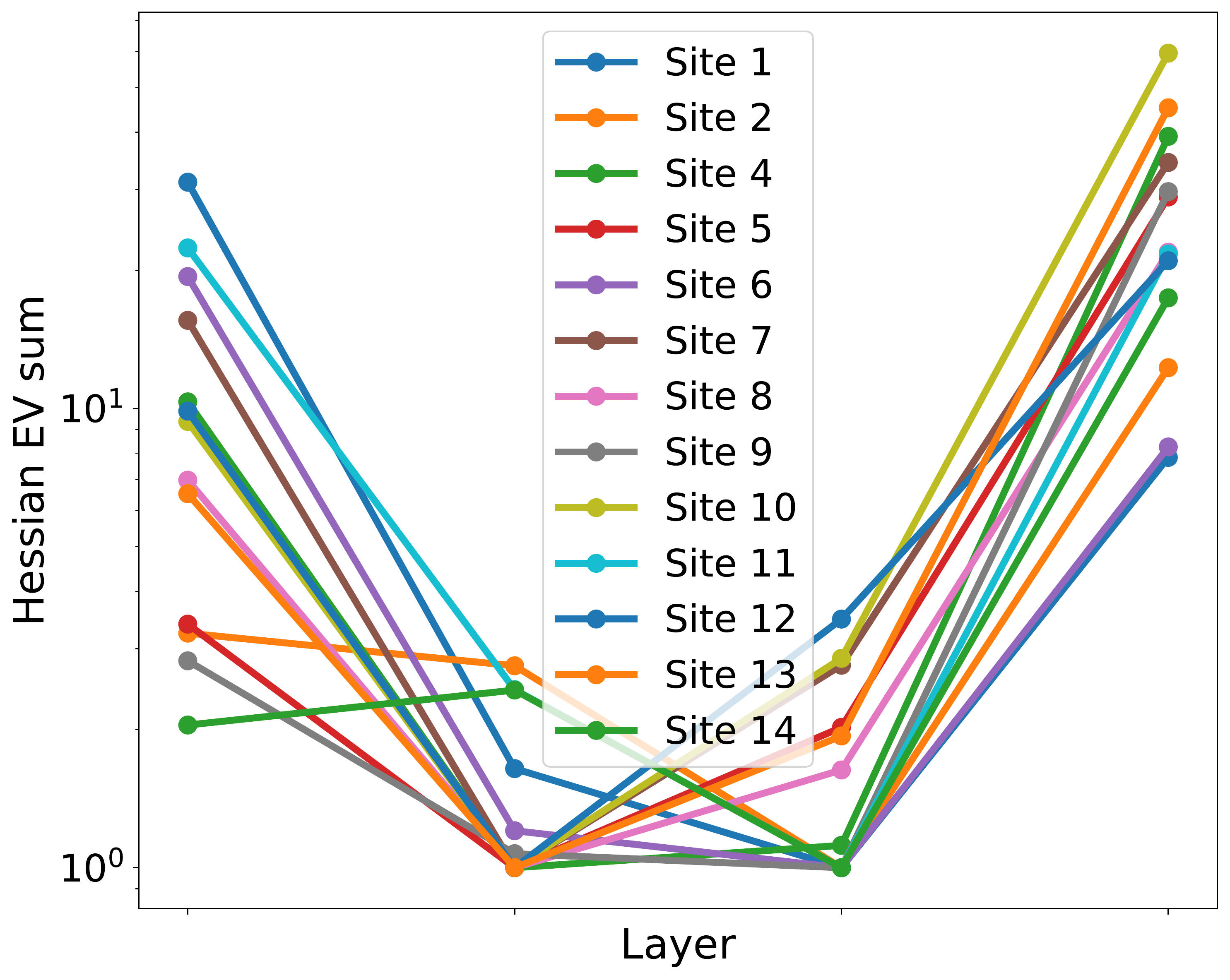}}
  \makebox[\linewidth][c]{\scriptsize\textit{(E) Sent-140}}  % Subfigure label
\end{minipage}%
\begin{minipage}{.24\linewidth}
  \centering
  \raisebox{-\height}{\includegraphics[width=\linewidth]{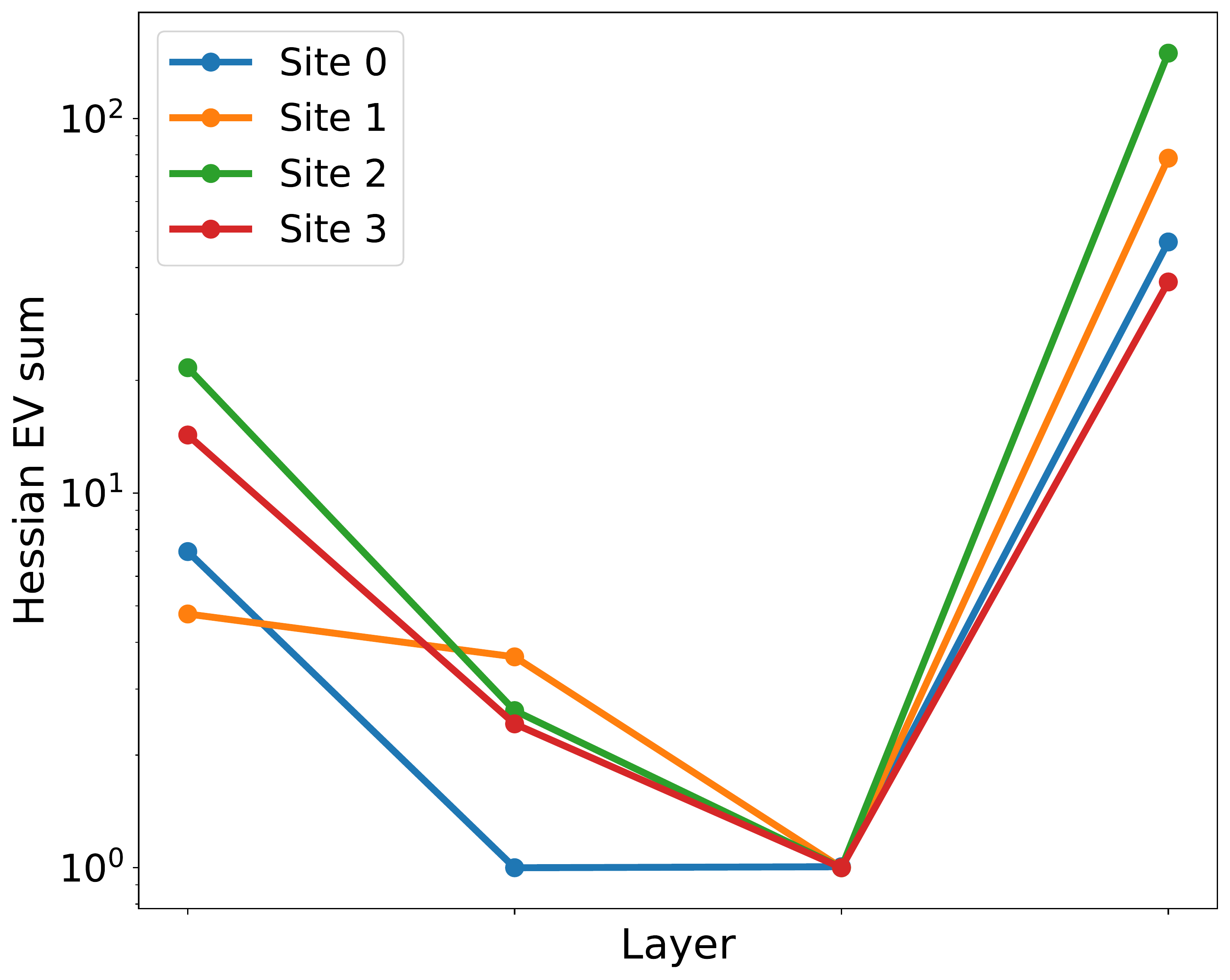}}
  \makebox[\linewidth][c]{\scriptsize\textit{(F) MIMIC-III}}  % Subfigure label
\end{minipage}%
\begin{minipage}{.24\linewidth}
  \centering
  \raisebox{-\height}{\includegraphics[width=\linewidth]{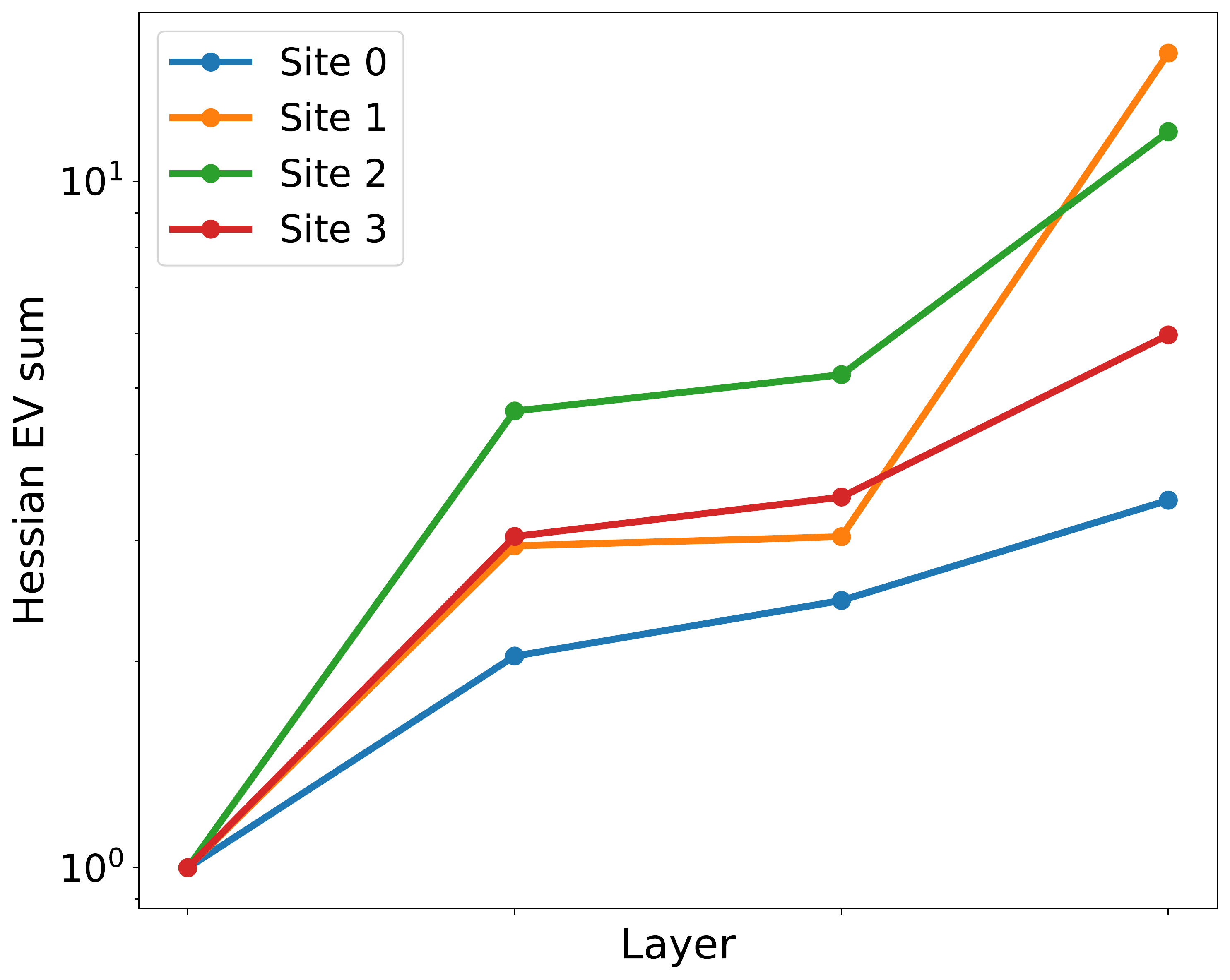}}
  \makebox[\linewidth][c]{\scriptsize\textit{(G) Fed-Heart-Disease}}  % Subfigure label
\end{minipage}
}

\caption{\textbf{Layer hessian eigenvalue sum for the final models}. All models identically initialized and independently trained on non-IID data.}
\label{sfig:hess_eig_best}
\end{center}

\end{figure}

\begin{figure}[ht]

\begin{center}

% First four subfigures
\makebox[\linewidth][c]{
\begin{minipage}{.24\linewidth}
  \centering
  \raisebox{-\height}{\includegraphics[width=\linewidth]{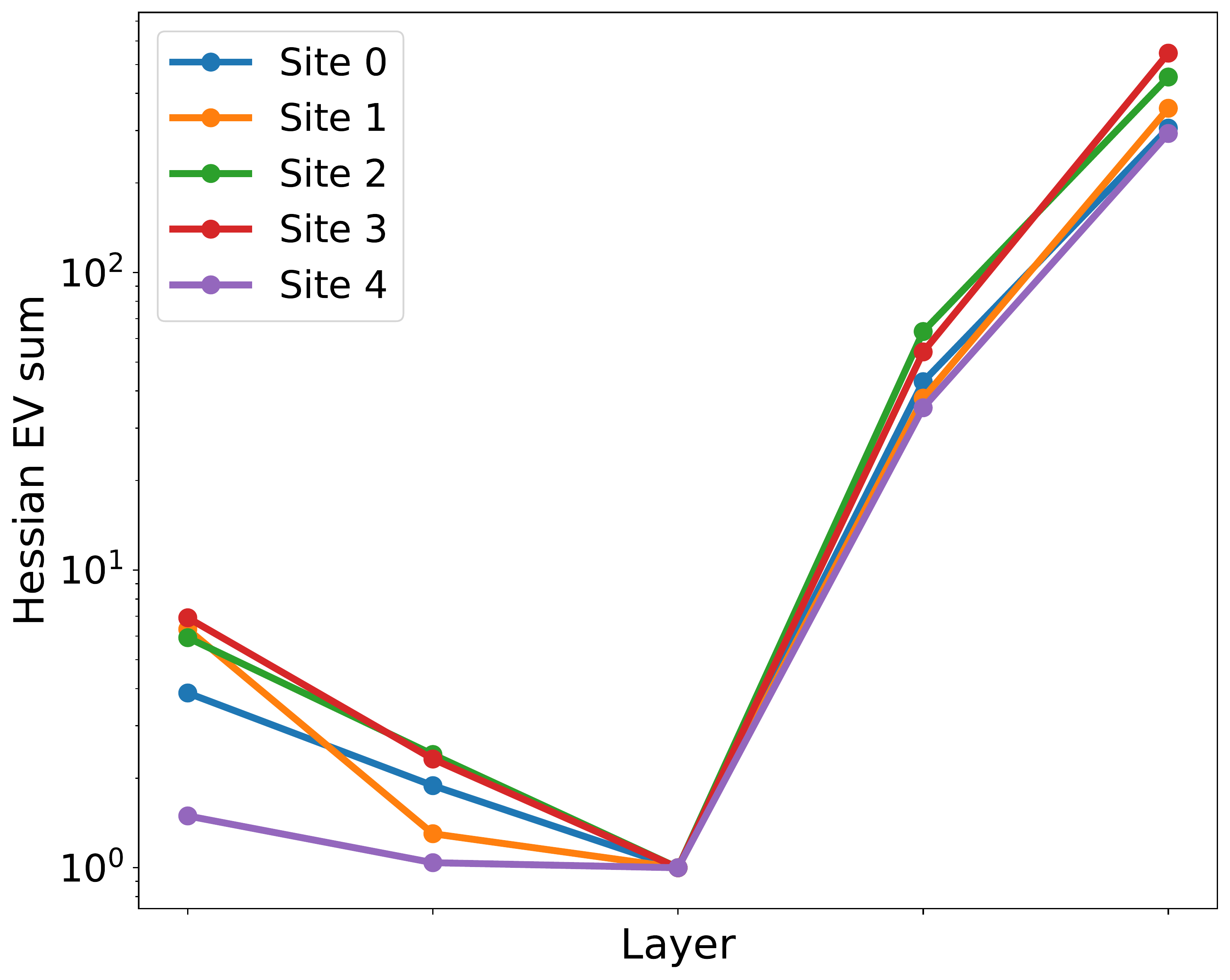}}
  \makebox[\linewidth][c]{\scriptsize\textit{(A) FashionMNIST}}  % Subfigure label
\end{minipage}%
\begin{minipage}{.24\linewidth}
  \centering
  \raisebox{-\height}{\includegraphics[width=\linewidth]{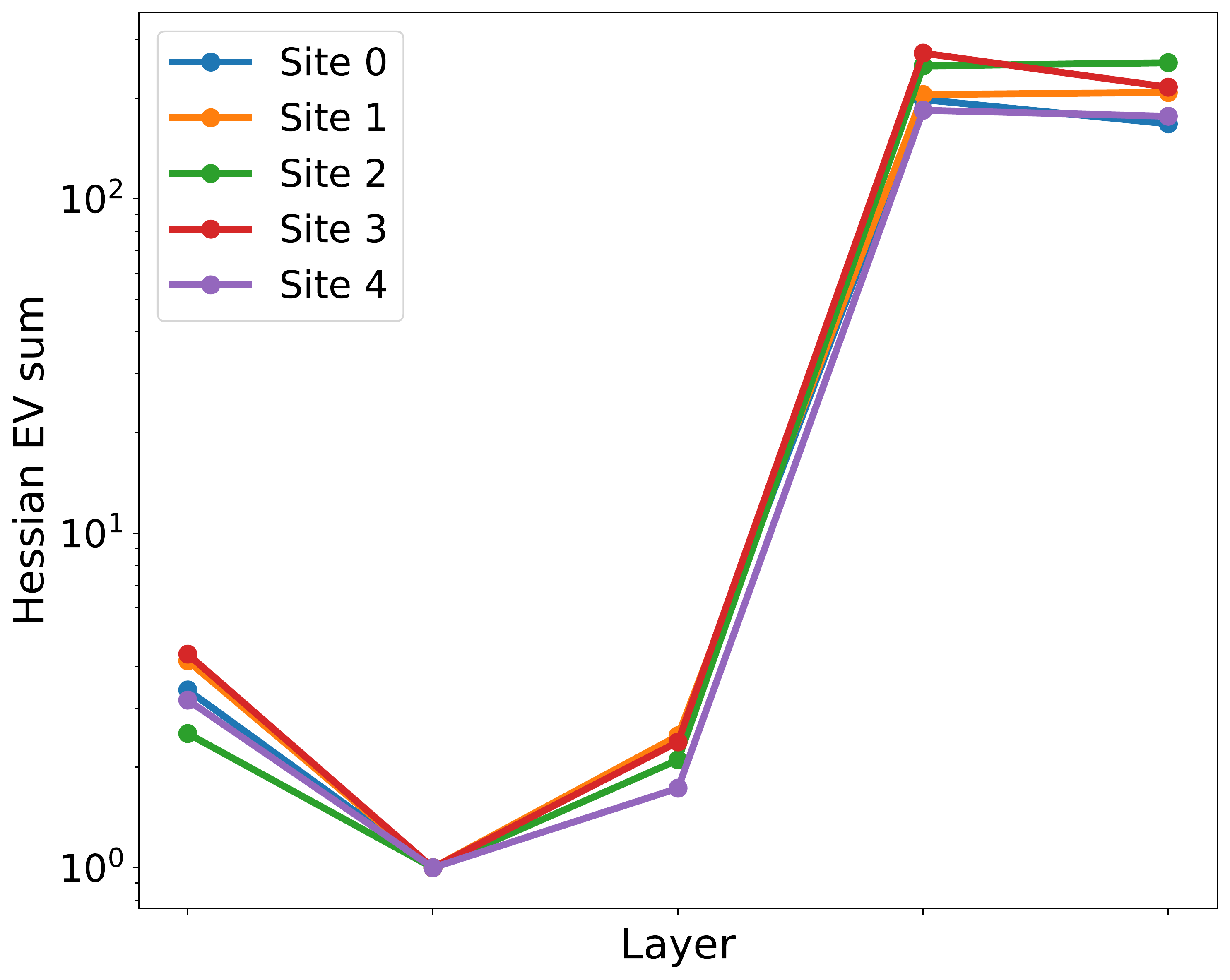}}
  \makebox[\linewidth][c]{\scriptsize\textit{(B) EMNIST}}  % Subfigure label
\end{minipage}%
\begin{minipage}{.24\linewidth}
  \centering
  \raisebox{-\height}{\includegraphics[width=\linewidth]{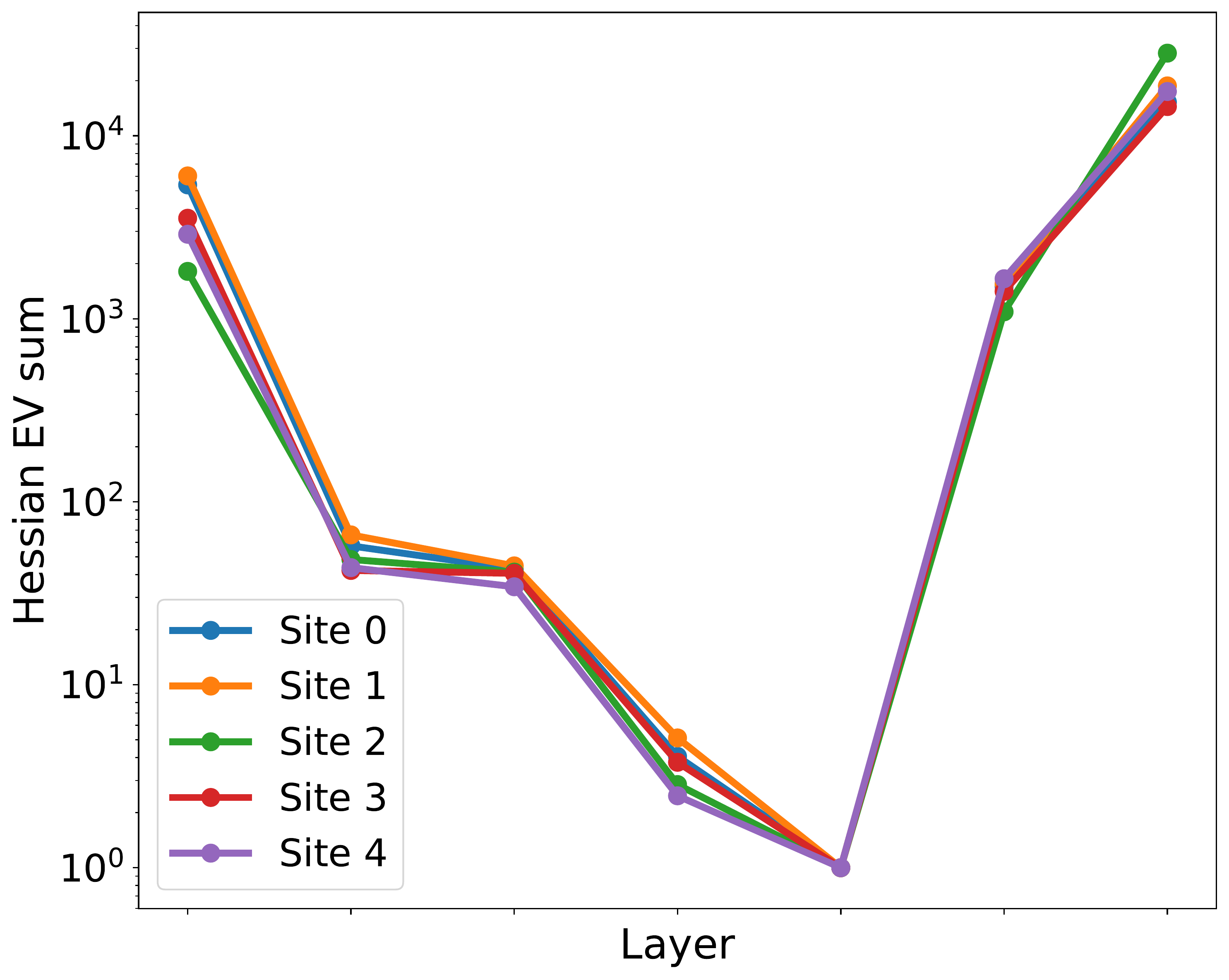}}
  \makebox[\linewidth][c]{\scriptsize\textit{(C) CIFAR-10}}  % Subfigure label
\end{minipage}%
\begin{minipage}{.24\linewidth}
  \centering
  \raisebox{-\height}{\includegraphics[width=\linewidth]{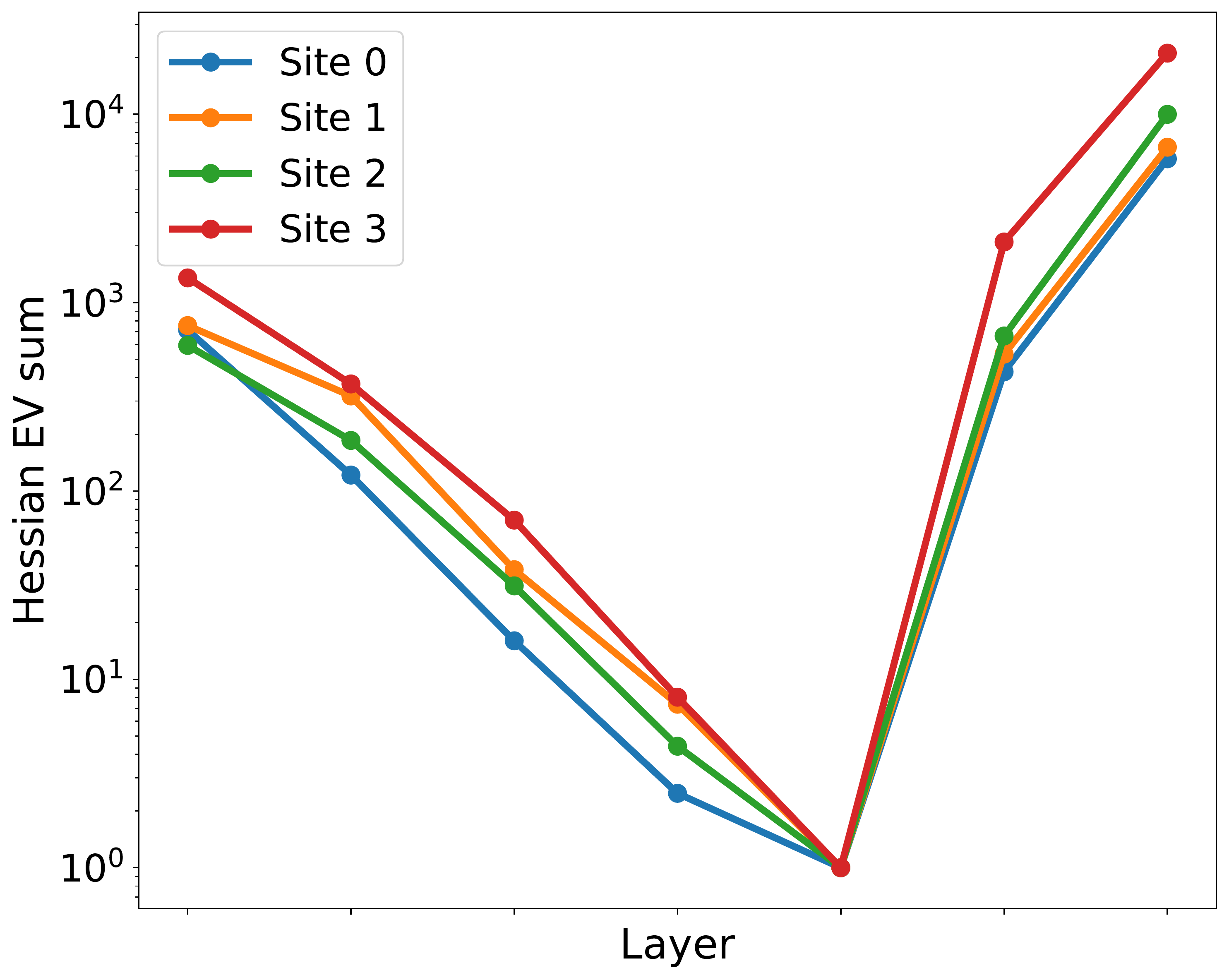}}
  \makebox[\linewidth][c]{\scriptsize\textit{(D) ISIC-2019}}  % Subfigure label
\end{minipage}
}
\makebox[\linewidth][c]{
\begin{minipage}{.24\linewidth}
  \centering
  \raisebox{-\height}{\includegraphics[width=\linewidth]{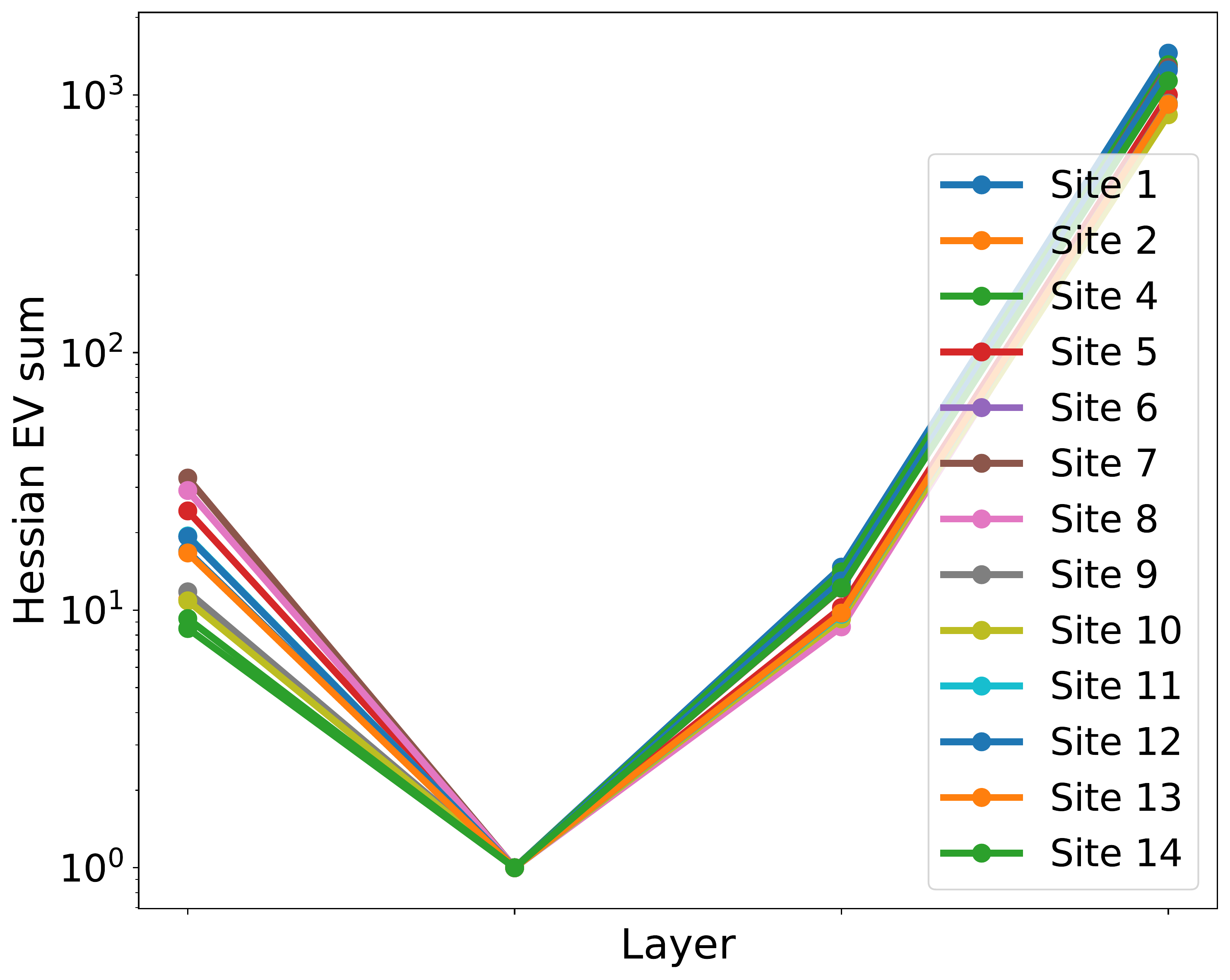}}
  \makebox[\linewidth][c]{\scriptsize\textit{(E) Sent-140}}  % Subfigure label
\end{minipage}%
\begin{minipage}{.24\linewidth}
  \centering
  \raisebox{-\height}{\includegraphics[width=\linewidth]{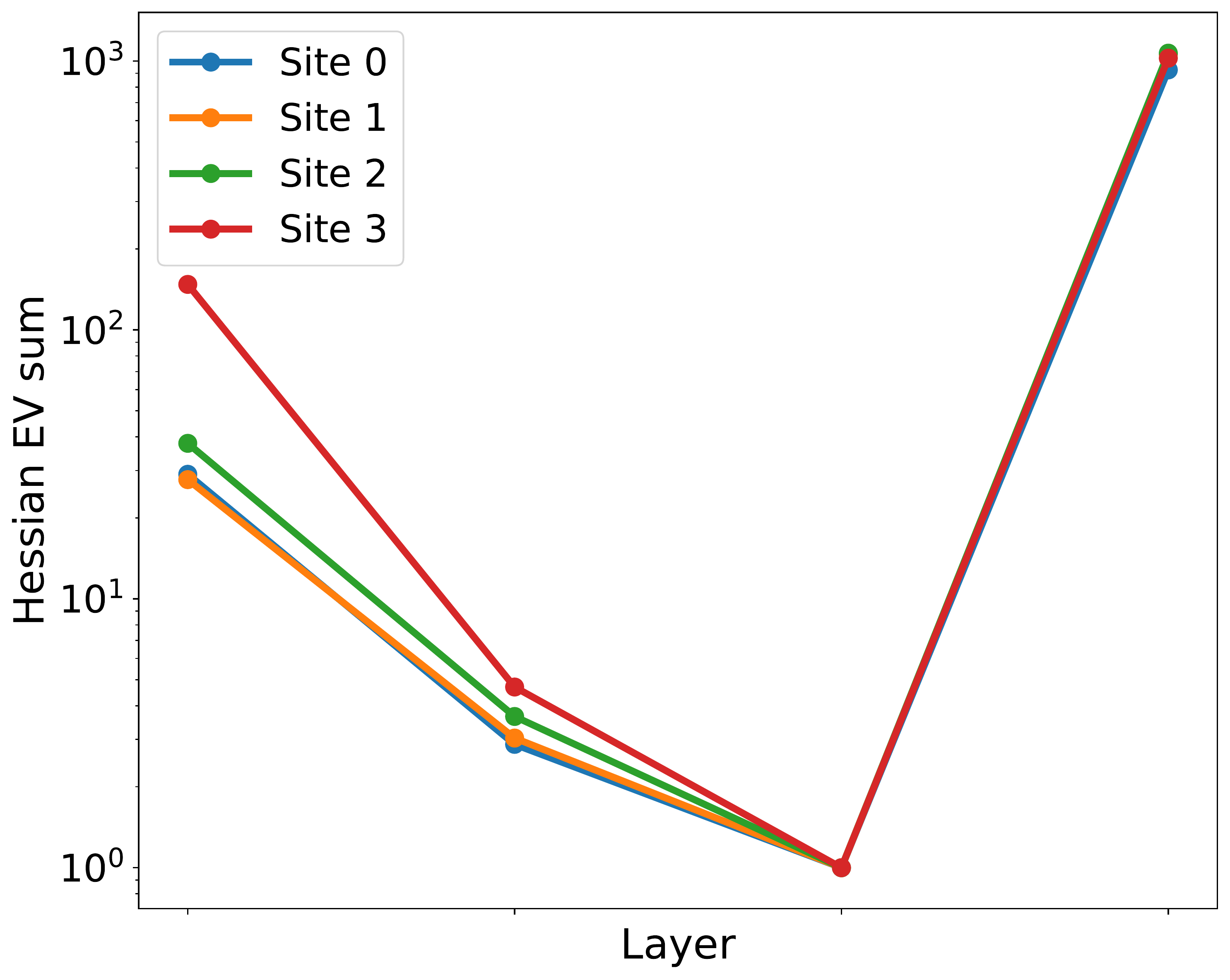}}
  \makebox[\linewidth][c]{\scriptsize\textit{(F) MIMIC-III}}  % Subfigure label
\end{minipage}%
\begin{minipage}{.24\linewidth}
  \centering
  \raisebox{-\height}{\includegraphics[width=\linewidth]{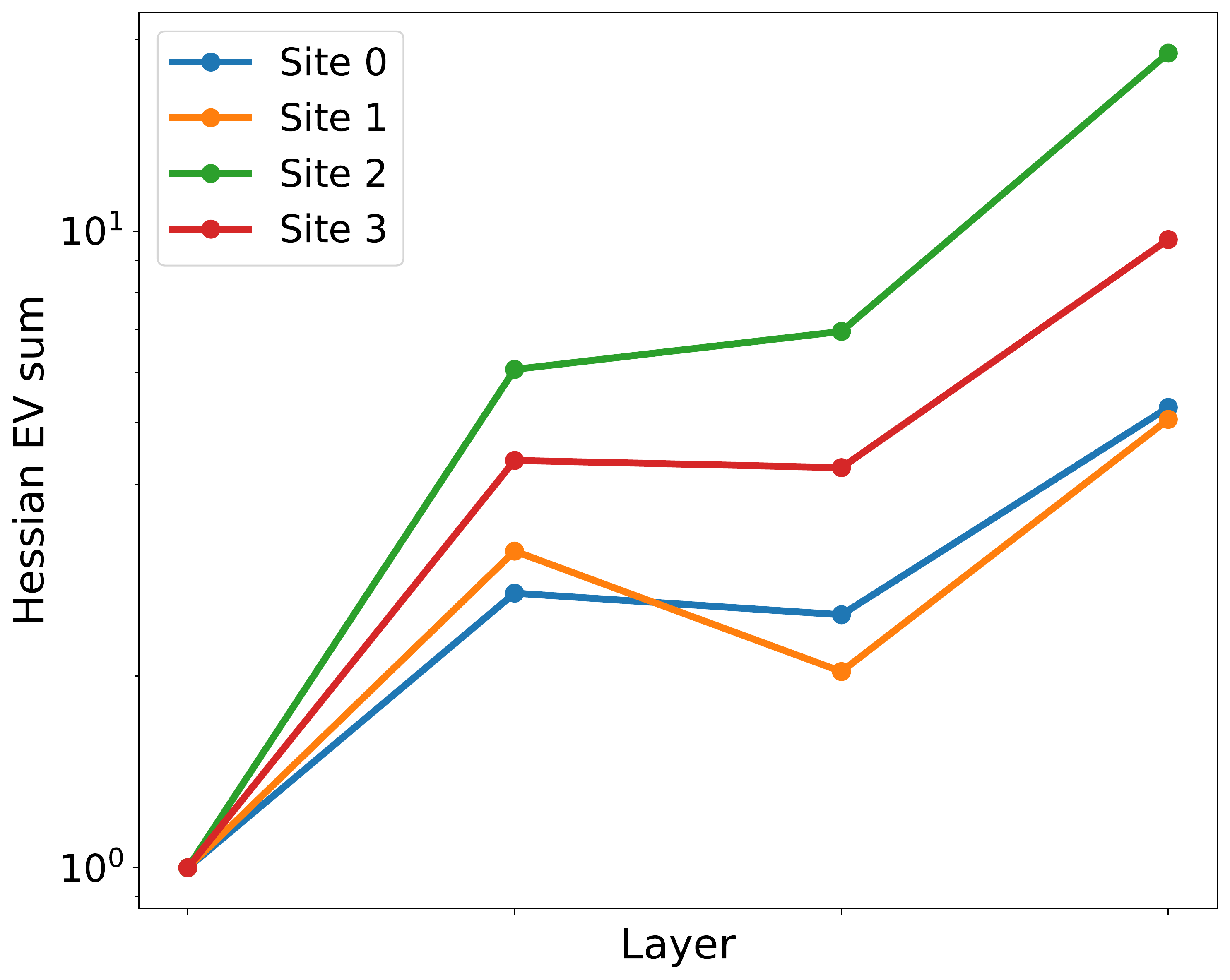}}
  \makebox[\linewidth][c]{\scriptsize\textit{(G) Fed-Heart-Disease}}  % Subfigure label
\end{minipage}
}

\caption{\textbf{Layer hessian eigenvalue sum for the final models}. Models trained via FL on non-IID data.}
\label{sfig:hess_eig_best_fl}
\end{center}

\end{figure}

Figures \ref{sfig:hess_eig_first}, \ref{sfig:hess_eig_best}, and \ref{sfig:hess_eig_best_fl} display the hessian eigenvalue sum across all datasets, corresponding to the models trained for a single epoch, final model after independent training, and final model after FL training, respectively.

\subsection{Sample representation}
\begin{figure}[ht]

\begin{center}

% First four subfigures
\makebox[\linewidth][c]{
\begin{minipage}{.24\linewidth}
  \centering
  \raisebox{-\height}{\includegraphics[width=\linewidth]{figures/FMNIST_similarity_first-1.png}}
  \makebox[\linewidth][c]{\scriptsize\textit{(A) FashionMNIST}}  % Subfigure label
\end{minipage}%
\begin{minipage}{.24\linewidth}
  \centering
  \raisebox{-\height}{\includegraphics[width=\linewidth]{figures/EMNIST_similarity_first-1.png}}
  \makebox[\linewidth][c]{\scriptsize\textit{(B) EMNIST}}  % Subfigure label
\end{minipage}%
\begin{minipage}{.24\linewidth}
  \centering
  \raisebox{-\height}{\includegraphics[width=\linewidth]{figures/CIFAR_similarity_first-1.png}}
  \makebox[\linewidth][c]{\scriptsize\textit{(C) CIFAR-10}}  % Subfigure label
\end{minipage}%
}
\makebox[\linewidth][c]{
\begin{minipage}{.24\linewidth}
  \centering
  \raisebox{-\height}{\includegraphics[width=\linewidth]{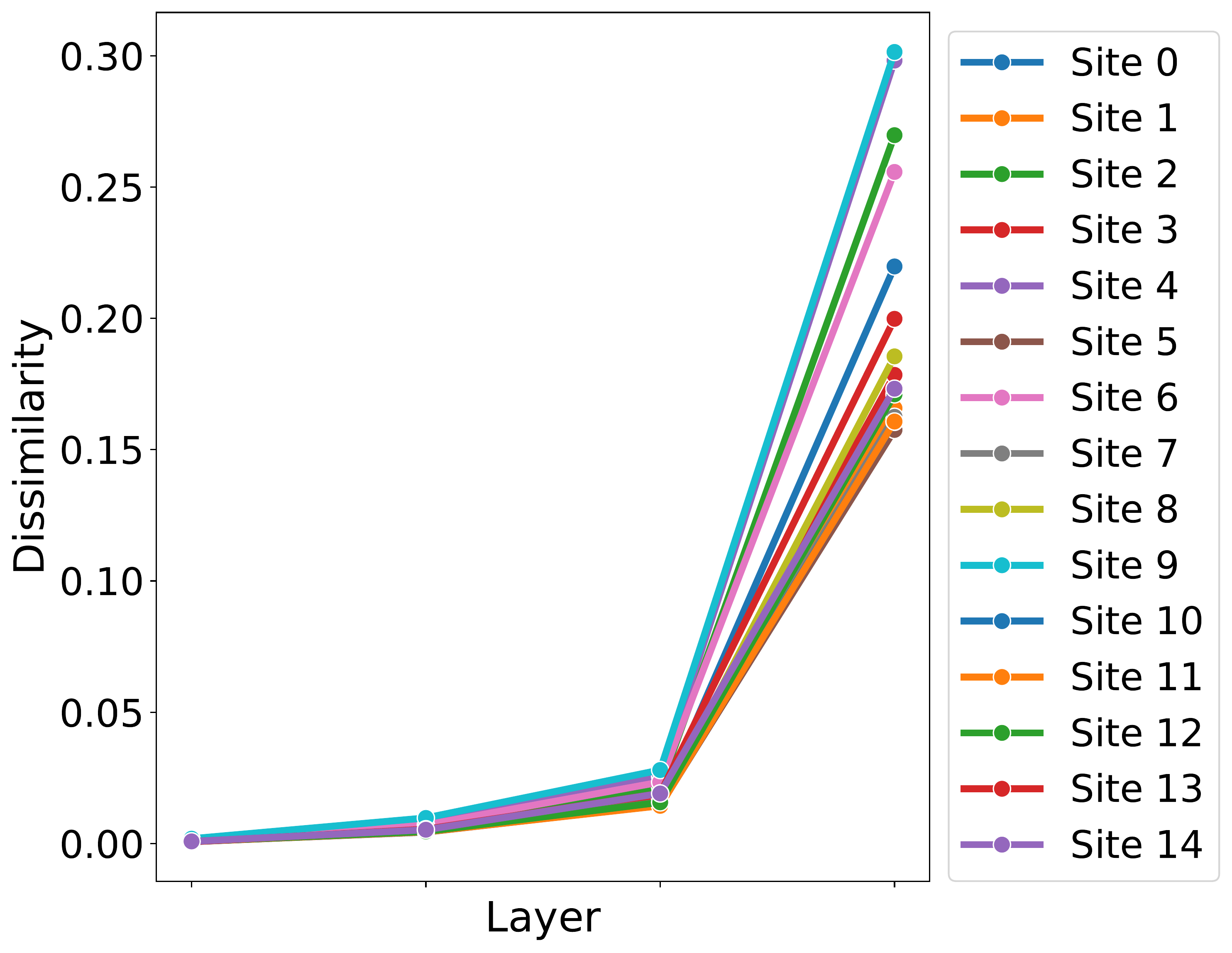}}
  \makebox[\linewidth][c]{\scriptsize\textit{(E) Sent-140}}  % Subfigure label
\end{minipage}%
\begin{minipage}{.24\linewidth}
  \centering
  \raisebox{-\height}{\includegraphics[width=\linewidth]{figures/mimic_similarity_first-1.png}}
  \makebox[\linewidth][c]{\scriptsize\textit{(F) MIMIC-III}}  % Subfigure label
\end{minipage}%
\begin{minipage}{.24\linewidth}
  \centering
  \raisebox{-\height}{\includegraphics[width=\linewidth]{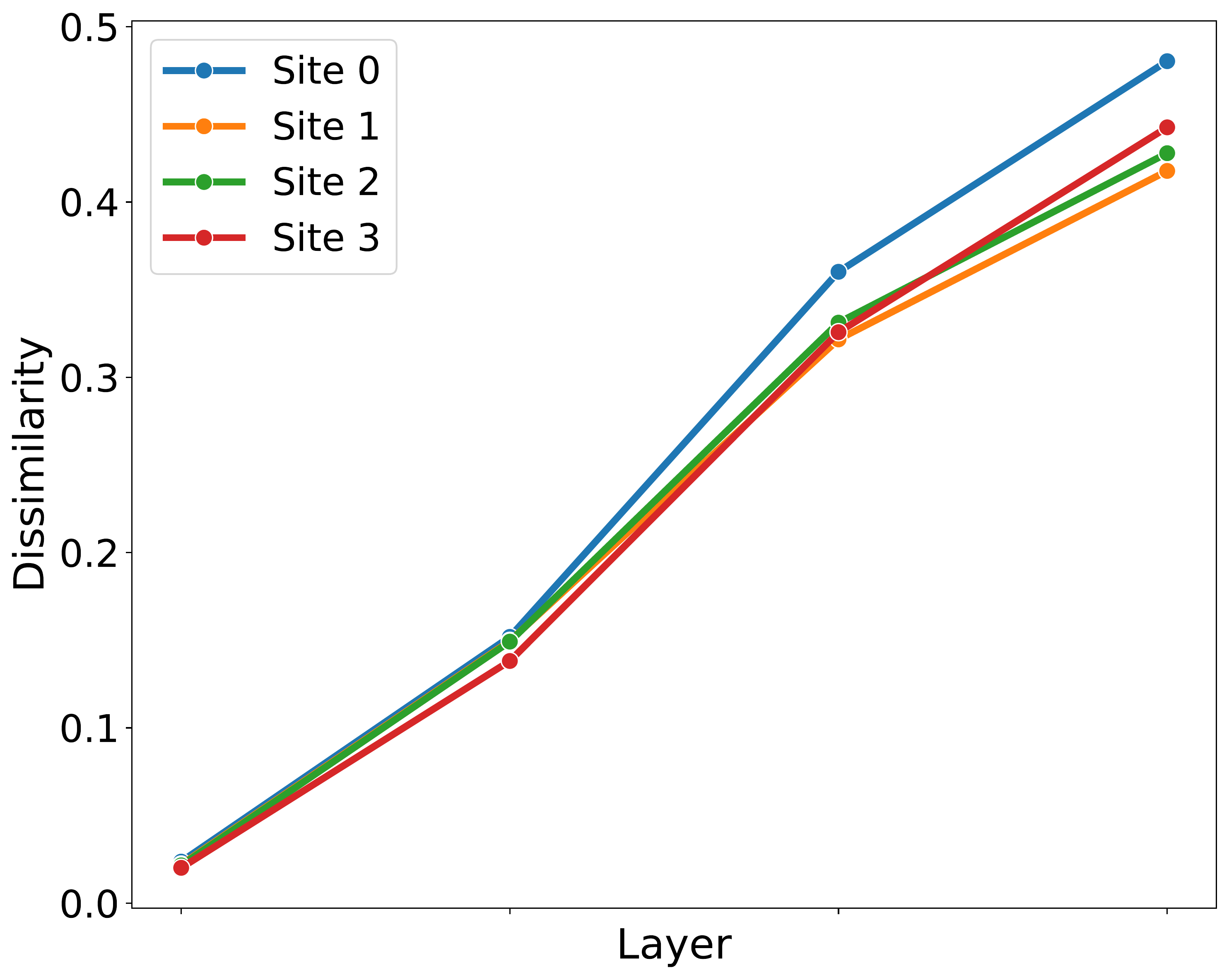}}
  \makebox[\linewidth][c]{\scriptsize\textit{(G) Fed-Heart-Disease}}  % Subfigure label
\end{minipage}
}

\caption{\textbf{Layer sample representation similarity after one epoch}. All models identically initialized and independently trained on non-IID data.}
\label{sfig:sample_rep_first}
\end{center}
   
\end{figure}

\begin{figure}[ht]

\begin{center}

% First four subfigures
\makebox[\linewidth][c]{
\begin{minipage}{.24\linewidth}
  \centering
  \raisebox{-\height}{\includegraphics[width=\linewidth]{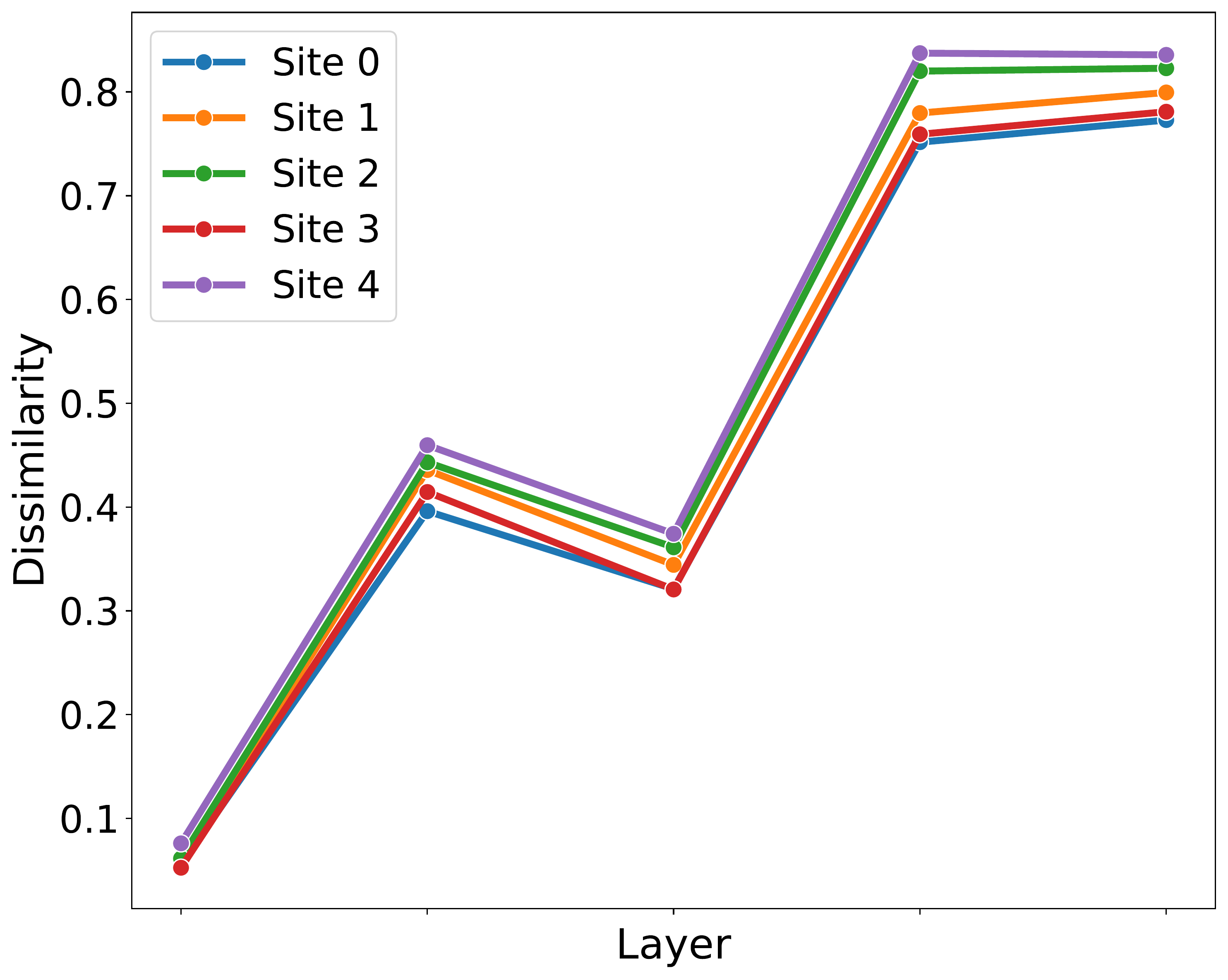}}
  \makebox[\linewidth][c]{\scriptsize\textit{(A) FashionMNIST}}  % Subfigure label
\end{minipage}%
\begin{minipage}{.24\linewidth}
  \centering
  \raisebox{-\height}{\includegraphics[width=\linewidth]{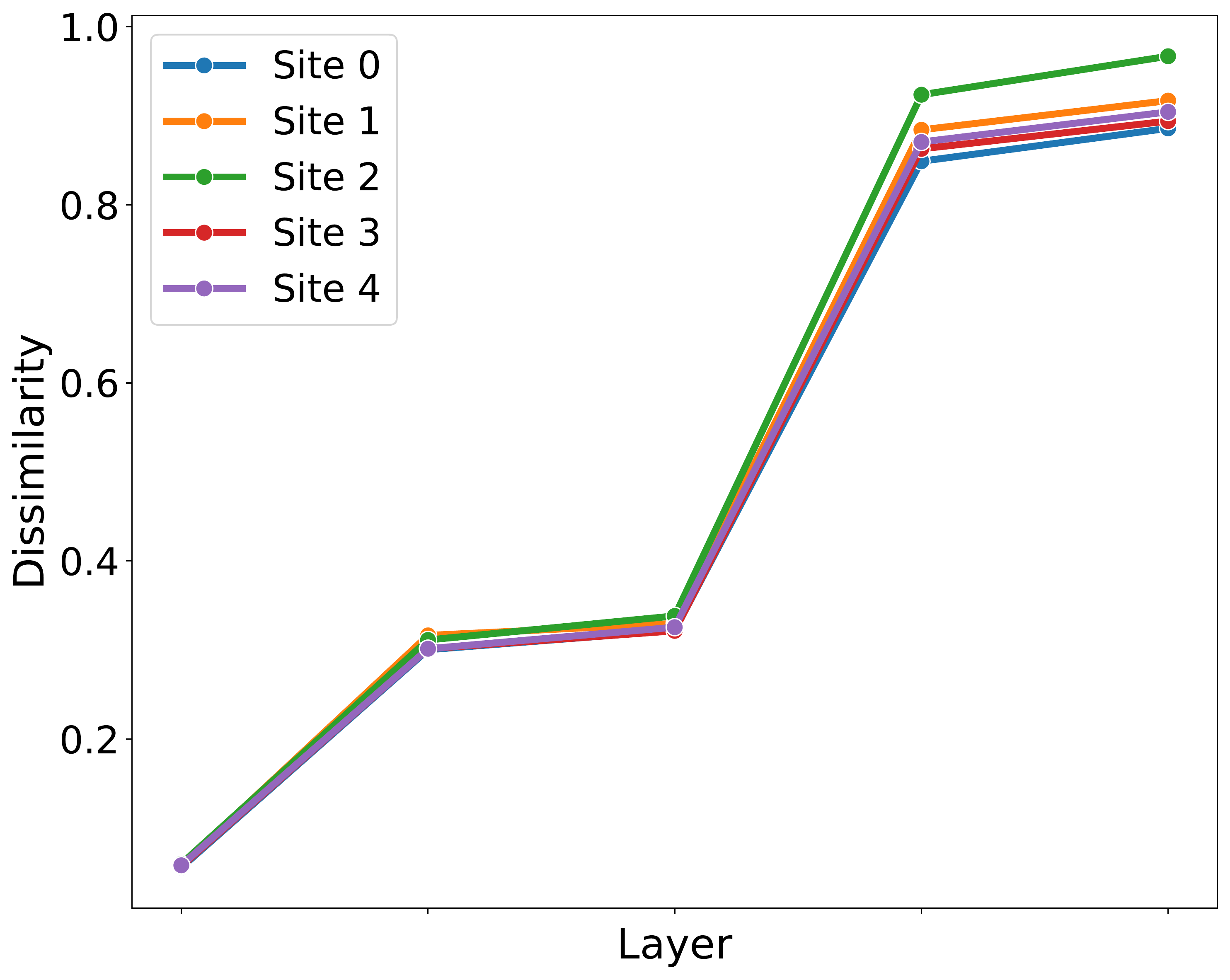}}
  \makebox[\linewidth][c]{\scriptsize\textit{(B) EMNIST}}  % Subfigure label
\end{minipage}%
\begin{minipage}{.24\linewidth}
  \centering
  \raisebox{-\height}{\includegraphics[width=\linewidth]{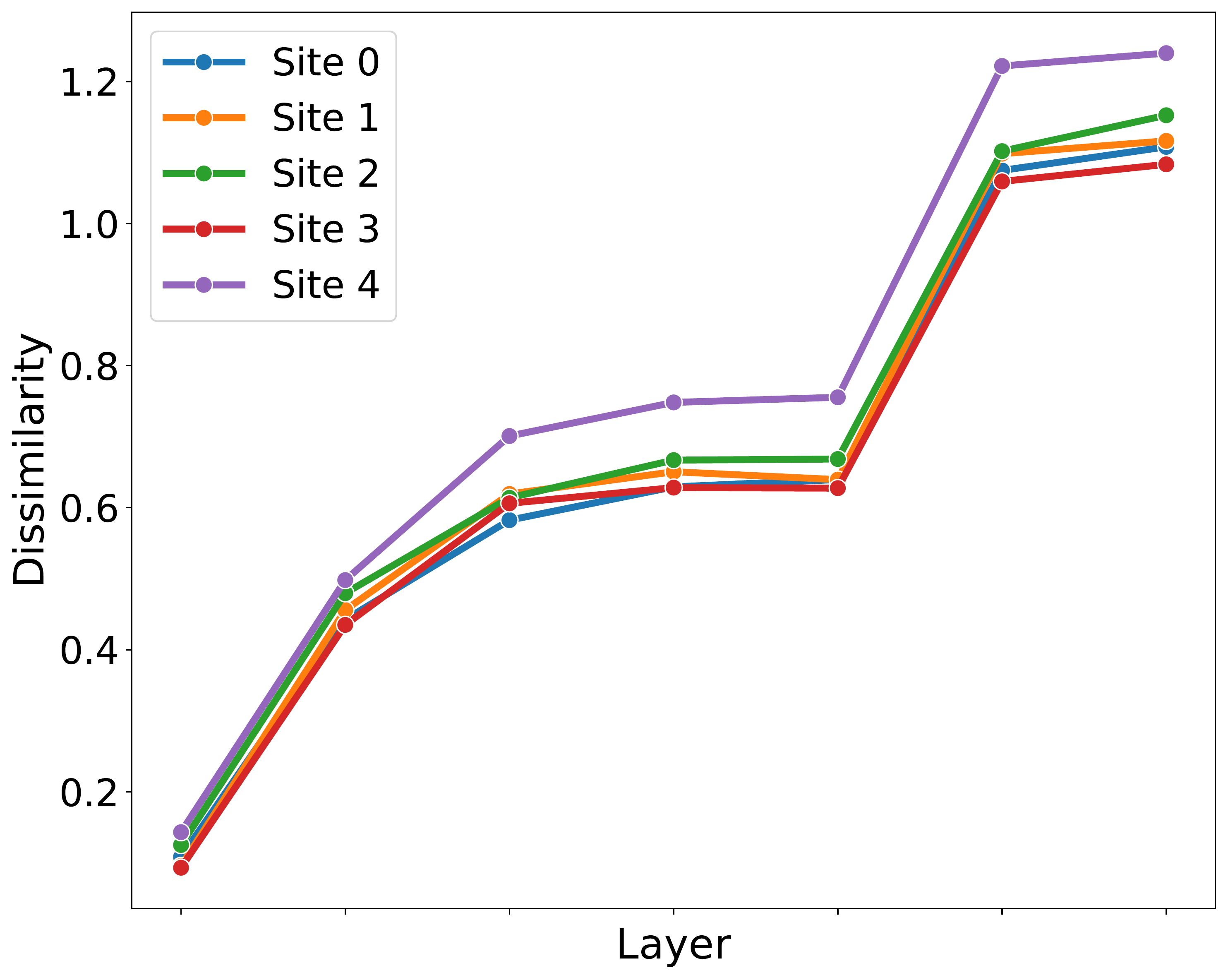}}
  \makebox[\linewidth][c]{\scriptsize\textit{(C) CIFAR-10}}  % Subfigure label
\end{minipage}%
}
\makebox[\linewidth][c]{
\begin{minipage}{.24\linewidth}
  \centering
  \raisebox{-\height}{\includegraphics[width=\linewidth]{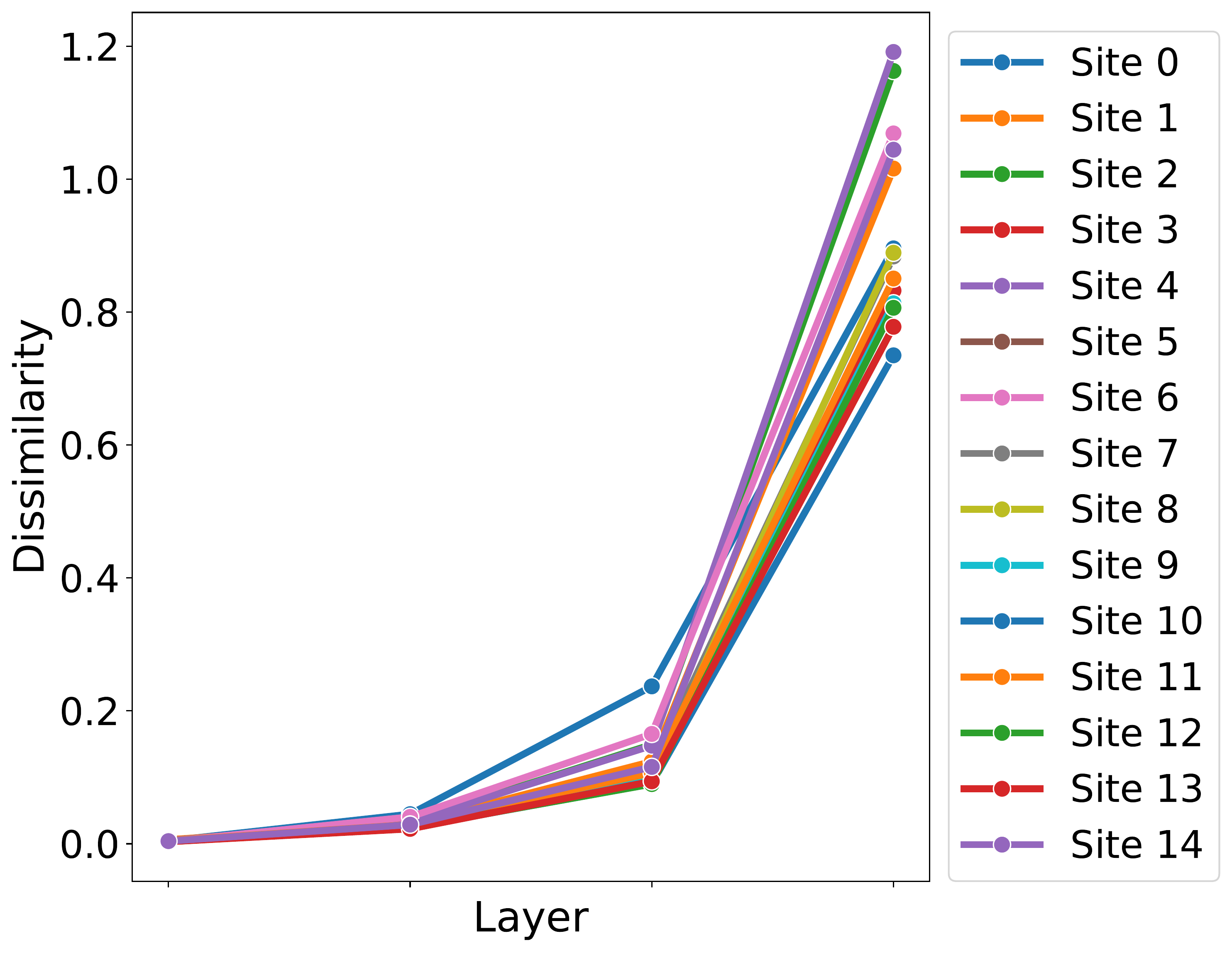}}
  \makebox[\linewidth][c]{\scriptsize\textit{(E) Sent-140}}  % Subfigure label
\end{minipage}%
\begin{minipage}{.24\linewidth}
  \centering
  \raisebox{-\height}{\includegraphics[width=\linewidth]{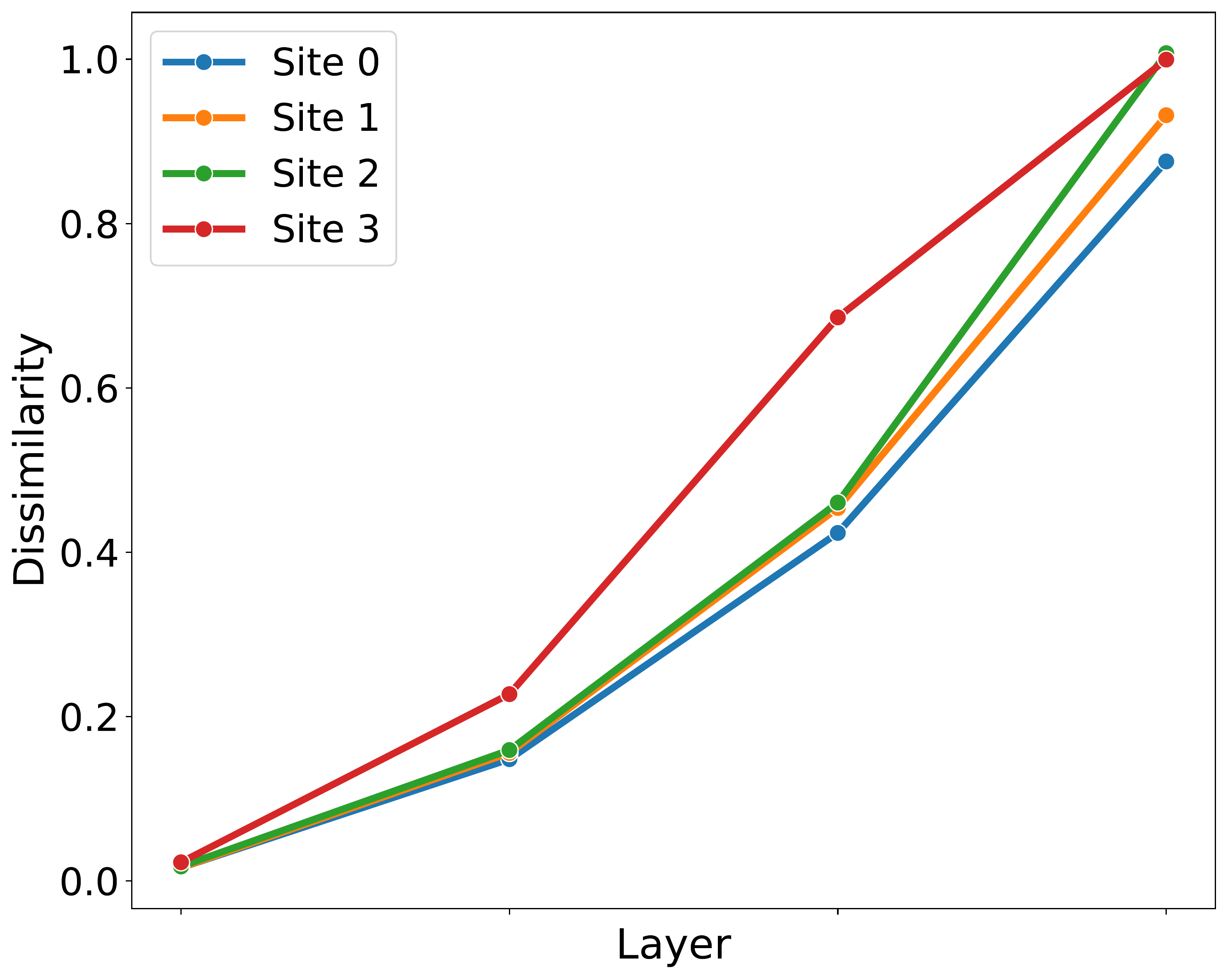}}
  \makebox[\linewidth][c]{\scriptsize\textit{(F) MIMIC-III}}  % Subfigure label
\end{minipage}%
\begin{minipage}{.24\linewidth}
  \centering
  \raisebox{-\height}{\includegraphics[width=\linewidth]{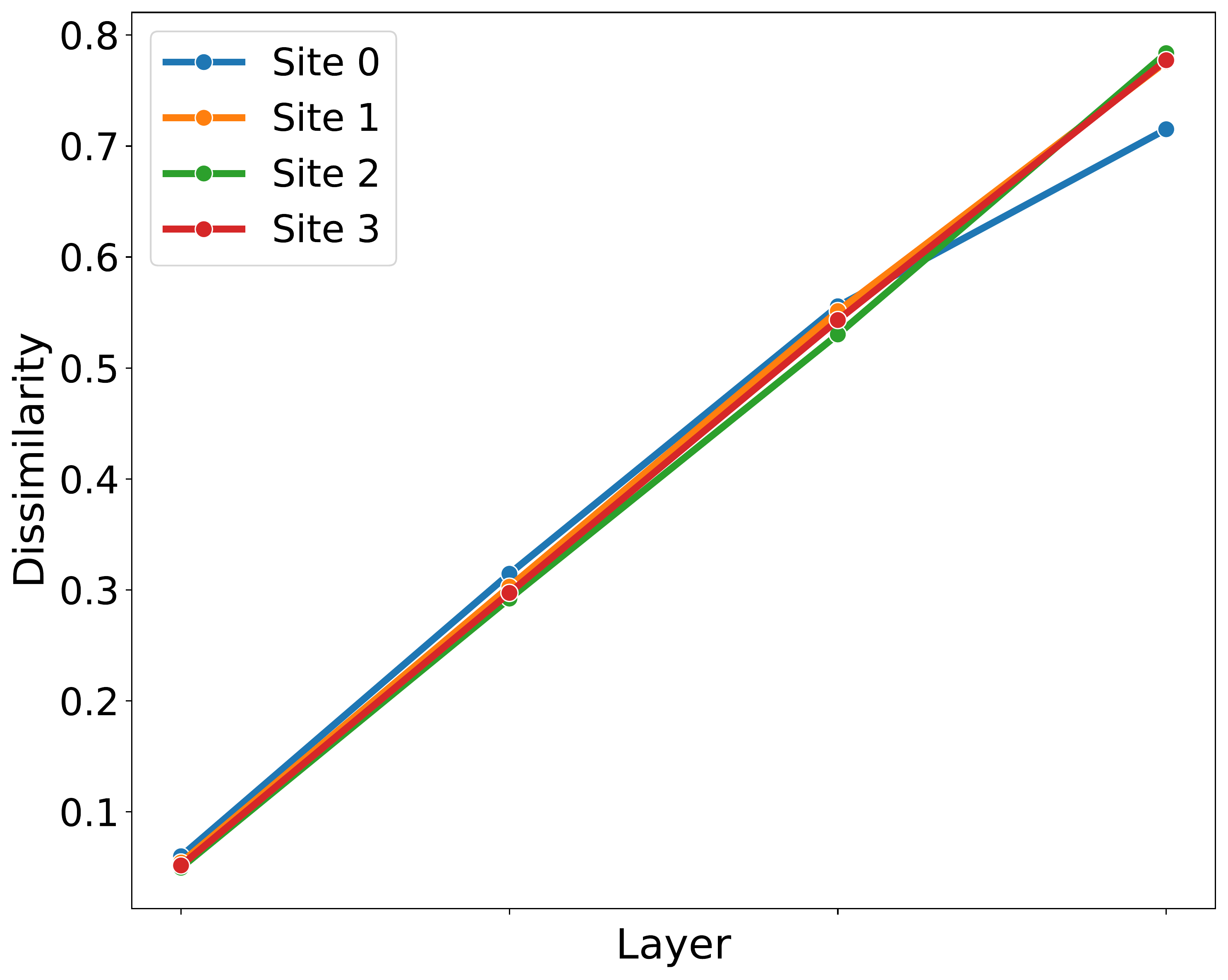}}
  \makebox[\linewidth][c]{\scriptsize\textit{(G) Fed-Heart-Disease}}  % Subfigure label
\end{minipage}
}

\caption{\textbf{Layer sample representation similarity for final models}. All models identically initialized and independently trained on non-IID data.}
\label{sfig:sample_rep_best}
\end{center}

\end{figure}

Figures \ref{sfig:sample_rep_first} and \ref{sfig:sample_rep_best} display the sample representation across all datasets, corresponding to the models trained for a single epoch and the final models, respectively.

\subsection{Federated sensitivity}
\begin{figure}[ht]

\begin{center}

% First four subfigures
\makebox[\linewidth][c]{
\begin{minipage}{.24\linewidth}
  \centering
  \raisebox{-\height}{\includegraphics[width=\linewidth]{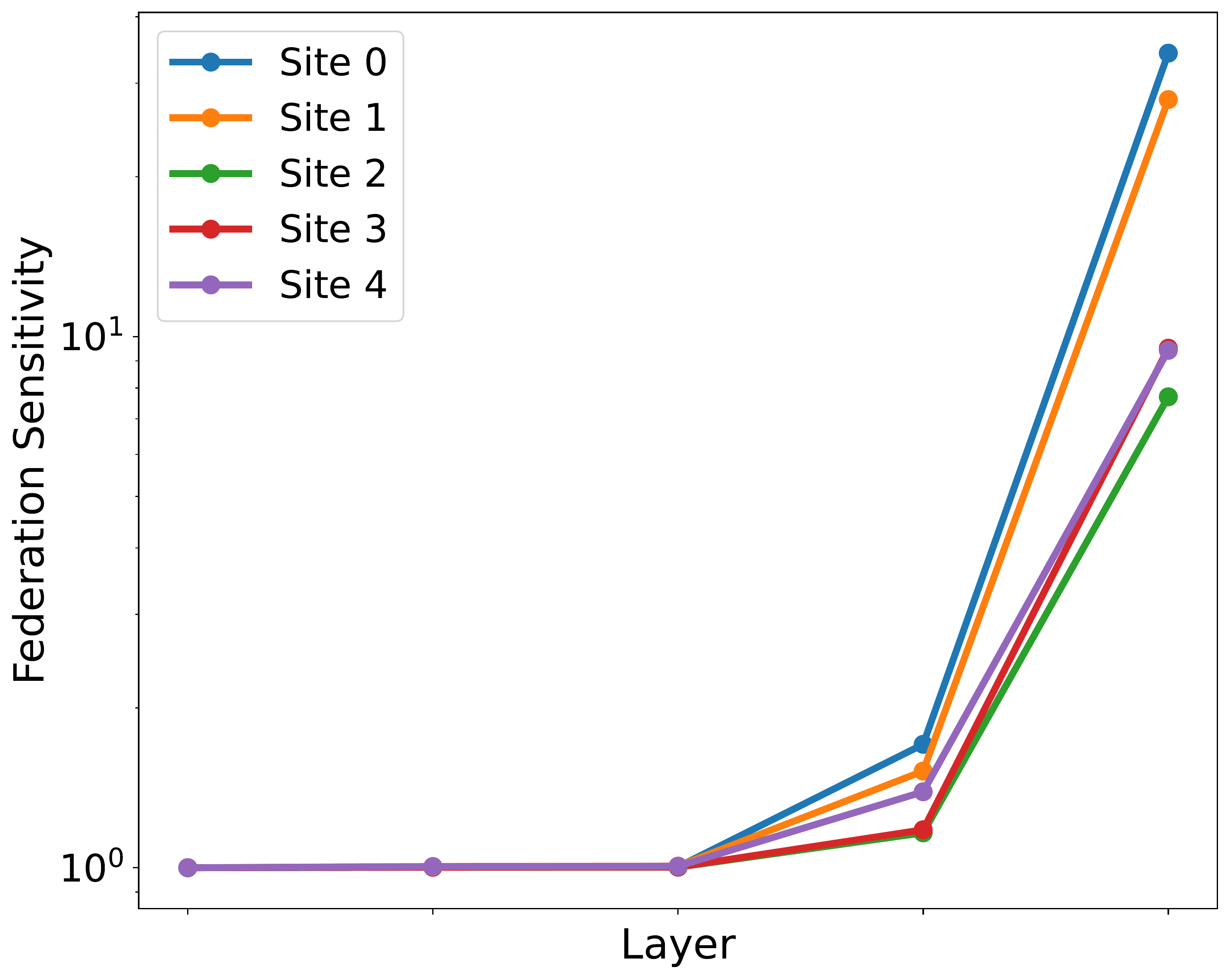}}
  \makebox[\linewidth][c]{\scriptsize\textit{(A) FashionMNIST}}  % Subfigure label
\end{minipage}%
\begin{minipage}{.24\linewidth}
  \centering
  \raisebox{-\height}{\includegraphics[width=\linewidth]{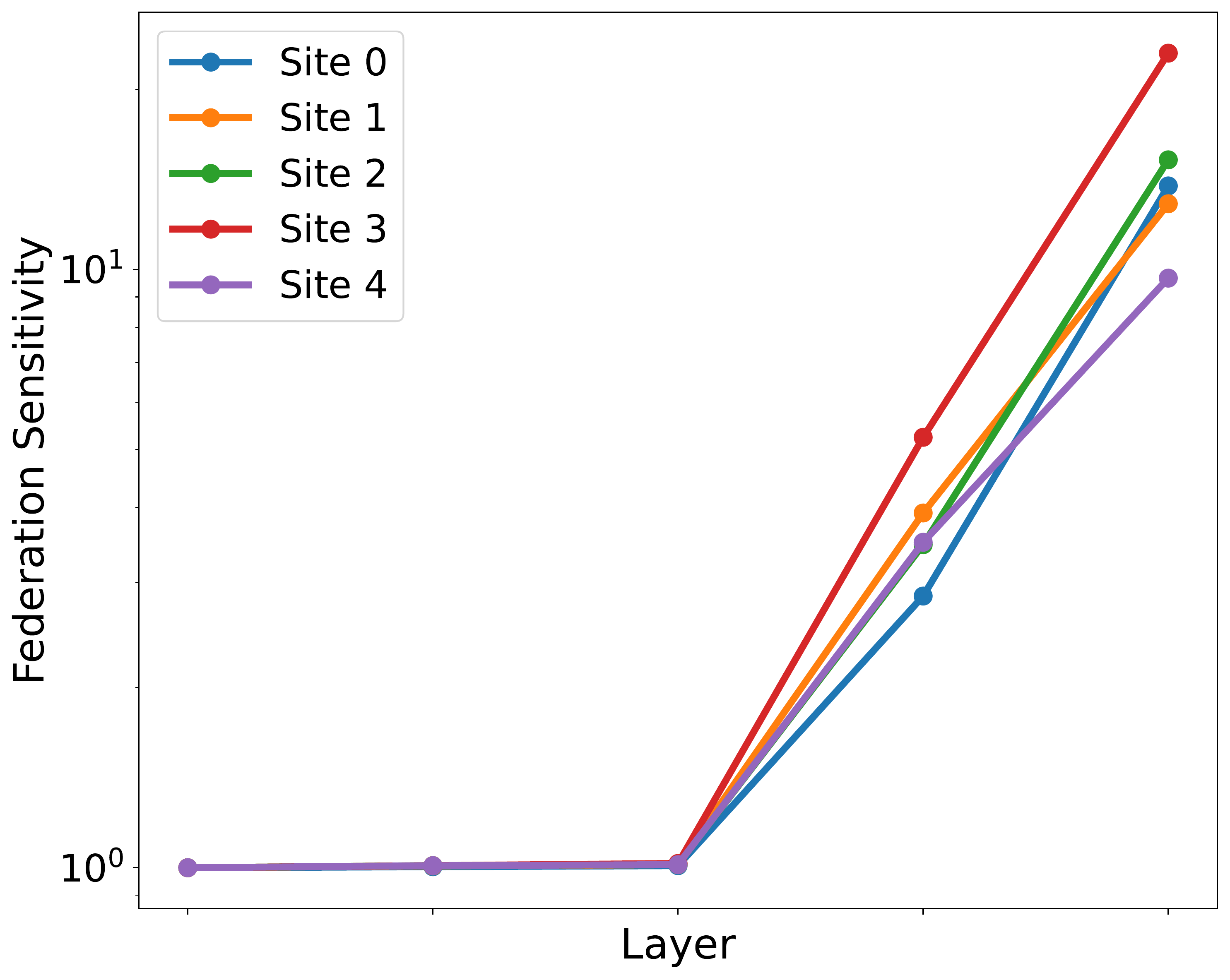}}
  \makebox[\linewidth][c]{\scriptsize\textit{(B) EMNIST}}  % Subfigure label
\end{minipage}%
\begin{minipage}{.24\linewidth}
  \centering
  \raisebox{-\height}{\includegraphics[width=\linewidth]{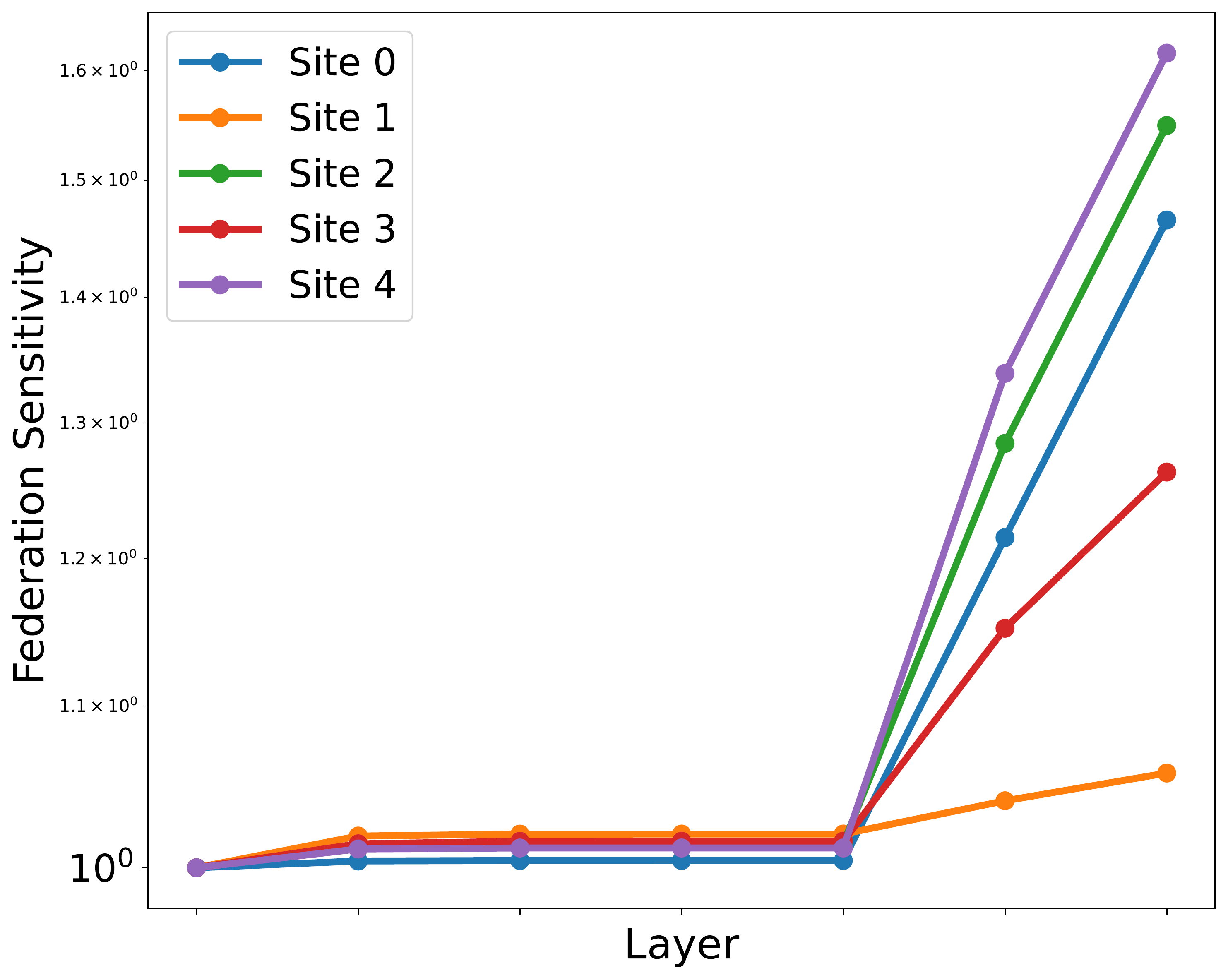}}
  \makebox[\linewidth][c]{\scriptsize\textit{(C) CIFAR-10}}  % Subfigure label
\end{minipage}%
\begin{minipage}{.24\linewidth}
  \centering
  \raisebox{-\height}{\includegraphics[width=\linewidth]{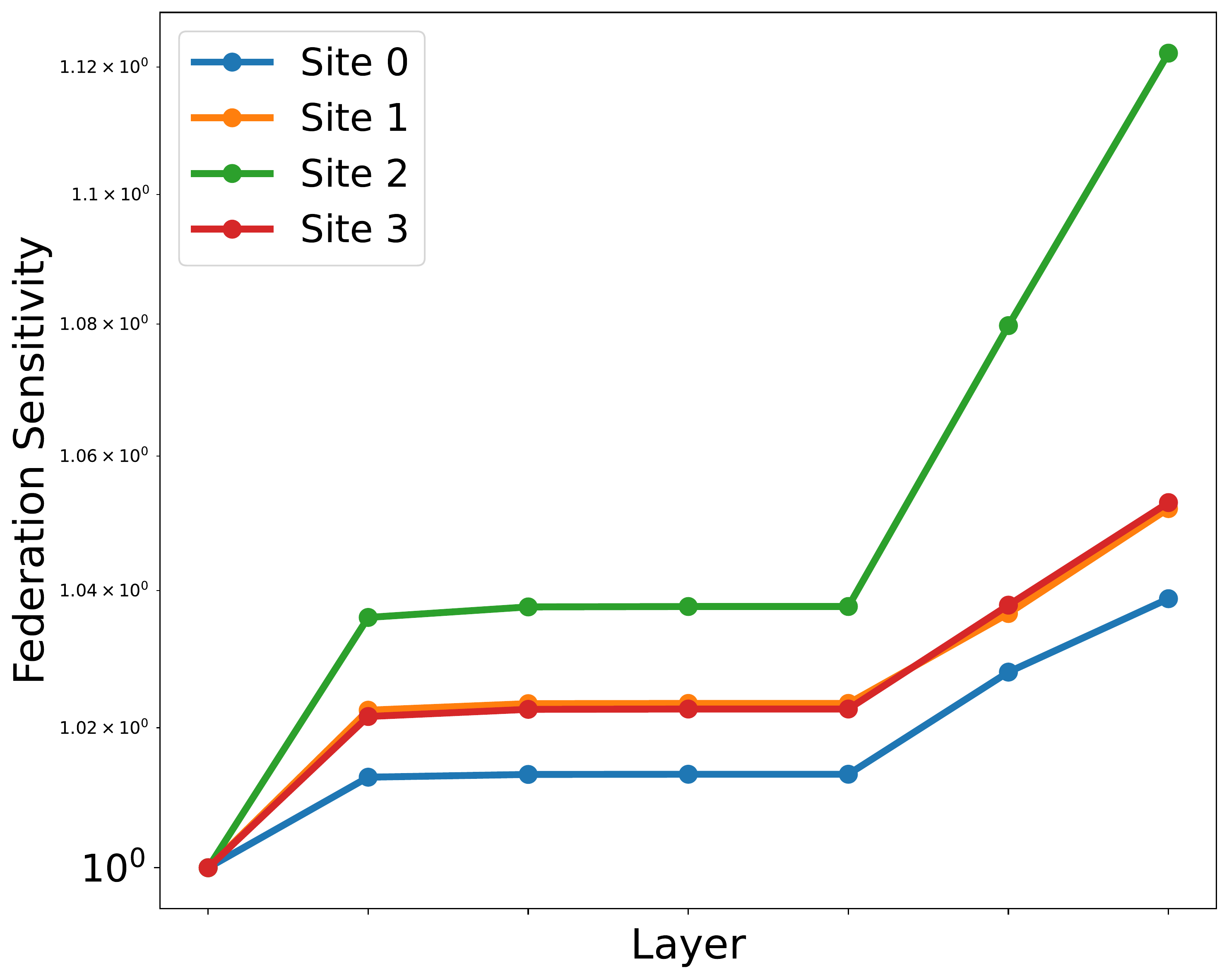}}
  \makebox[\linewidth][c]{\scriptsize\textit{(D) ISIC-2019}}  % Subfigure label
\end{minipage}
}
\makebox[\linewidth][c]{
\begin{minipage}{.24\linewidth}
  \centering
  \raisebox{-\height}{\includegraphics[width=\linewidth]{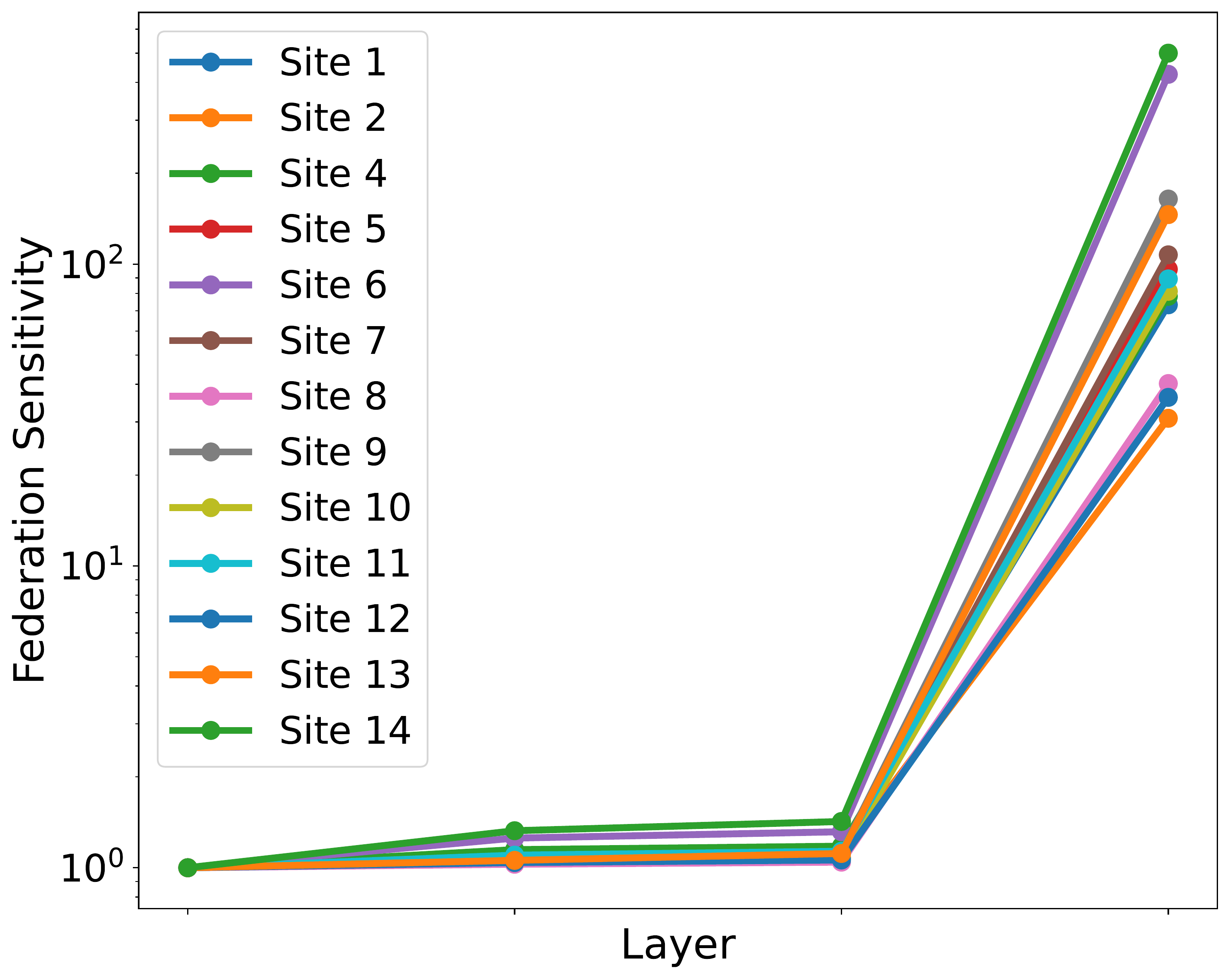}}
  \makebox[\linewidth][c]{\scriptsize\textit{(E) Sent-140}}  % Subfigure label
\end{minipage}%
\begin{minipage}{.24\linewidth}
  \centering
  \raisebox{-\height}{\includegraphics[width=\linewidth]{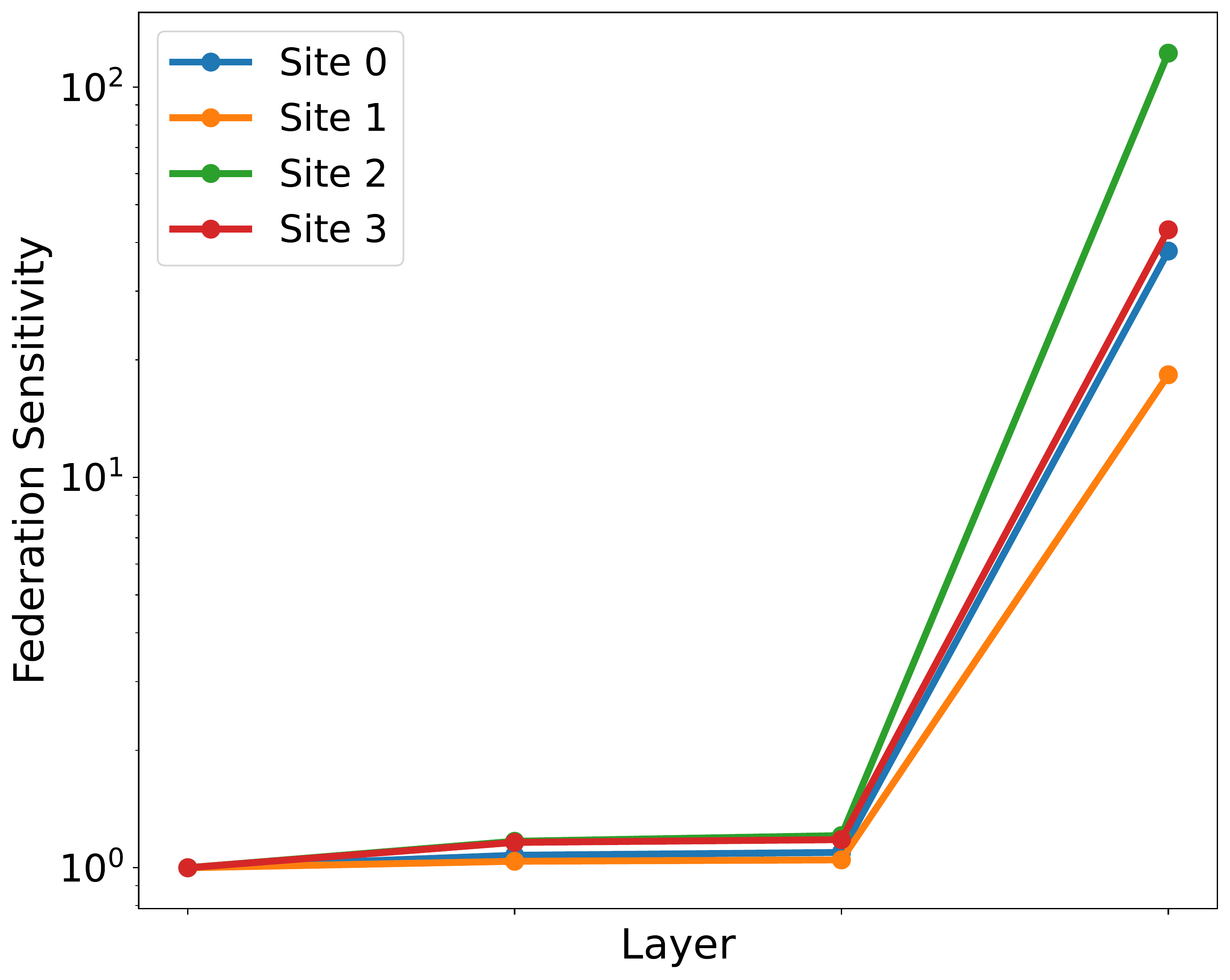}}
  \makebox[\linewidth][c]{\scriptsize\textit{(F) MIMIC-III}}  % Subfigure label
\end{minipage}%
\begin{minipage}{.24\linewidth}
  \centering
  \raisebox{-\height}{\includegraphics[width=\linewidth]{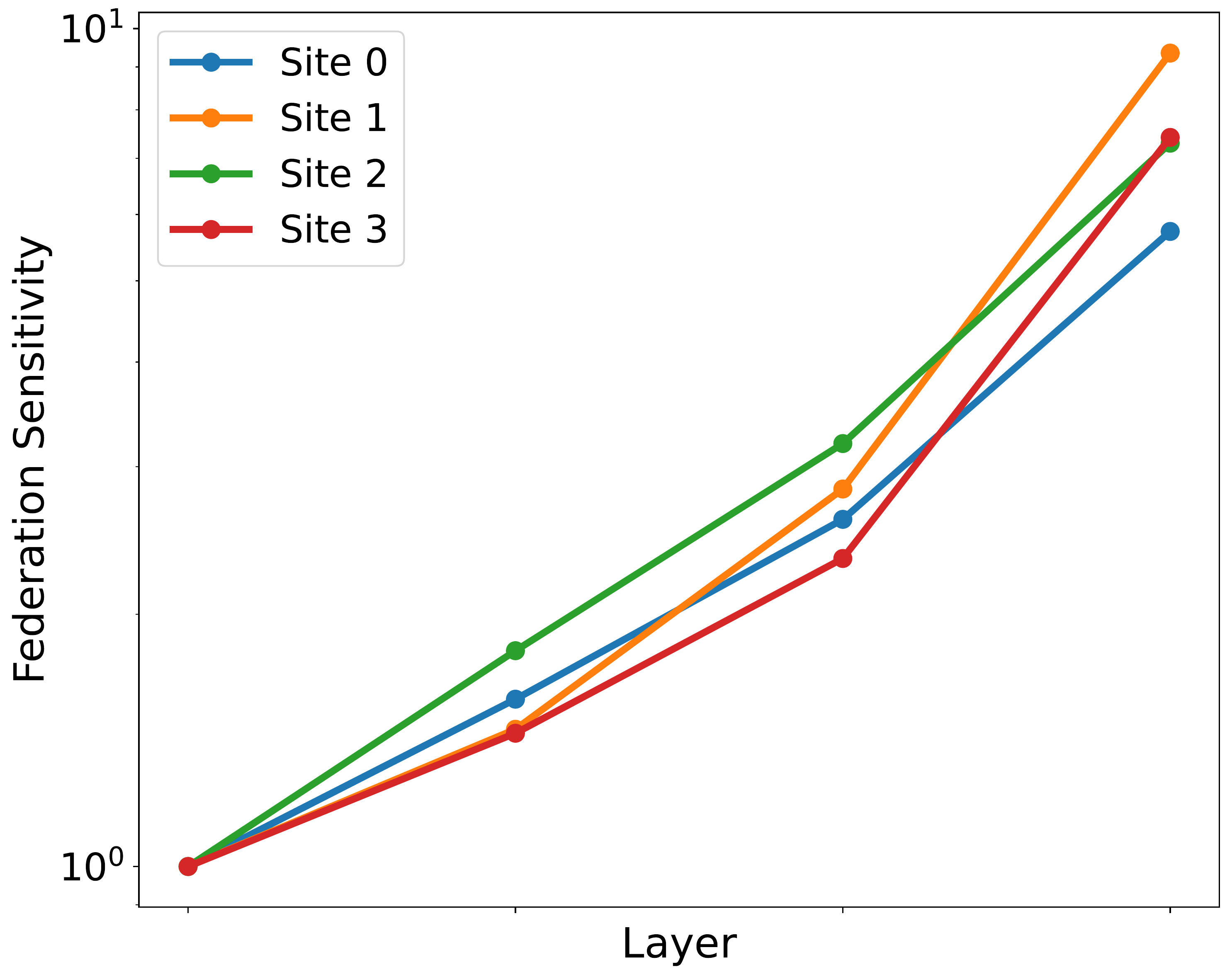}}
  \makebox[\linewidth][c]{\scriptsize\textit{(G) Fed-Heart-Disease}}  % Subfigure label
\end{minipage}
}

\caption{\textbf{Federation sensitivity for after one epoch}. All models identically initialized and independently trained on non-IID data.}
\label{sfig:layer_imp_best}
\end{center}

\end{figure}

\begin{figure}[ht]

\begin{center}

% First four subfigures
\makebox[\linewidth][c]{
\begin{minipage}{.24\linewidth}
  \centering
  \raisebox{-\height}{\includegraphics[width=\linewidth]{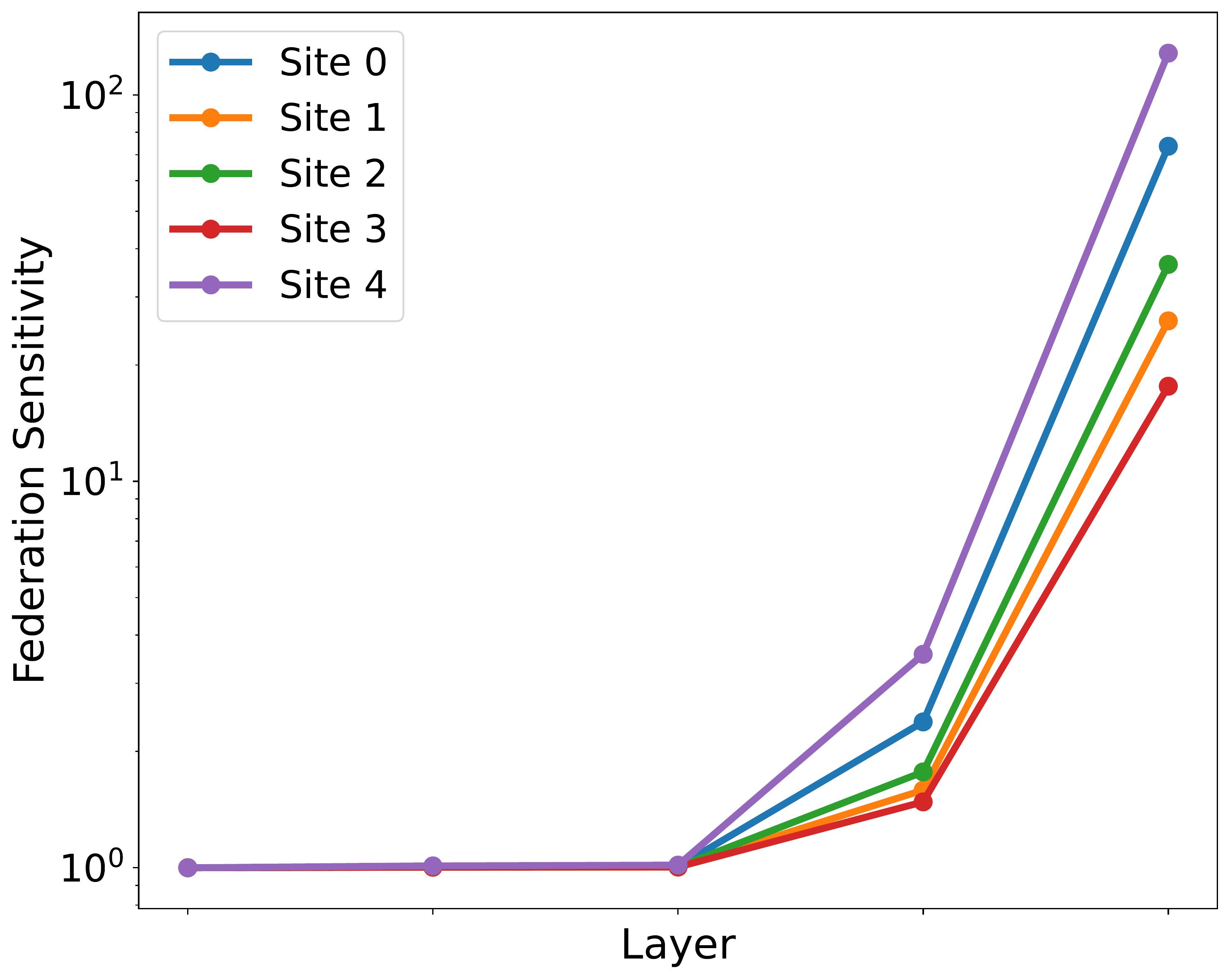}}
  \makebox[\linewidth][c]{\scriptsize\textit{(A) FashionMNIST}}  % Subfigure label
\end{minipage}%
\begin{minipage}{.24\linewidth}
  \centering
  \raisebox{-\height}{\includegraphics[width=\linewidth]{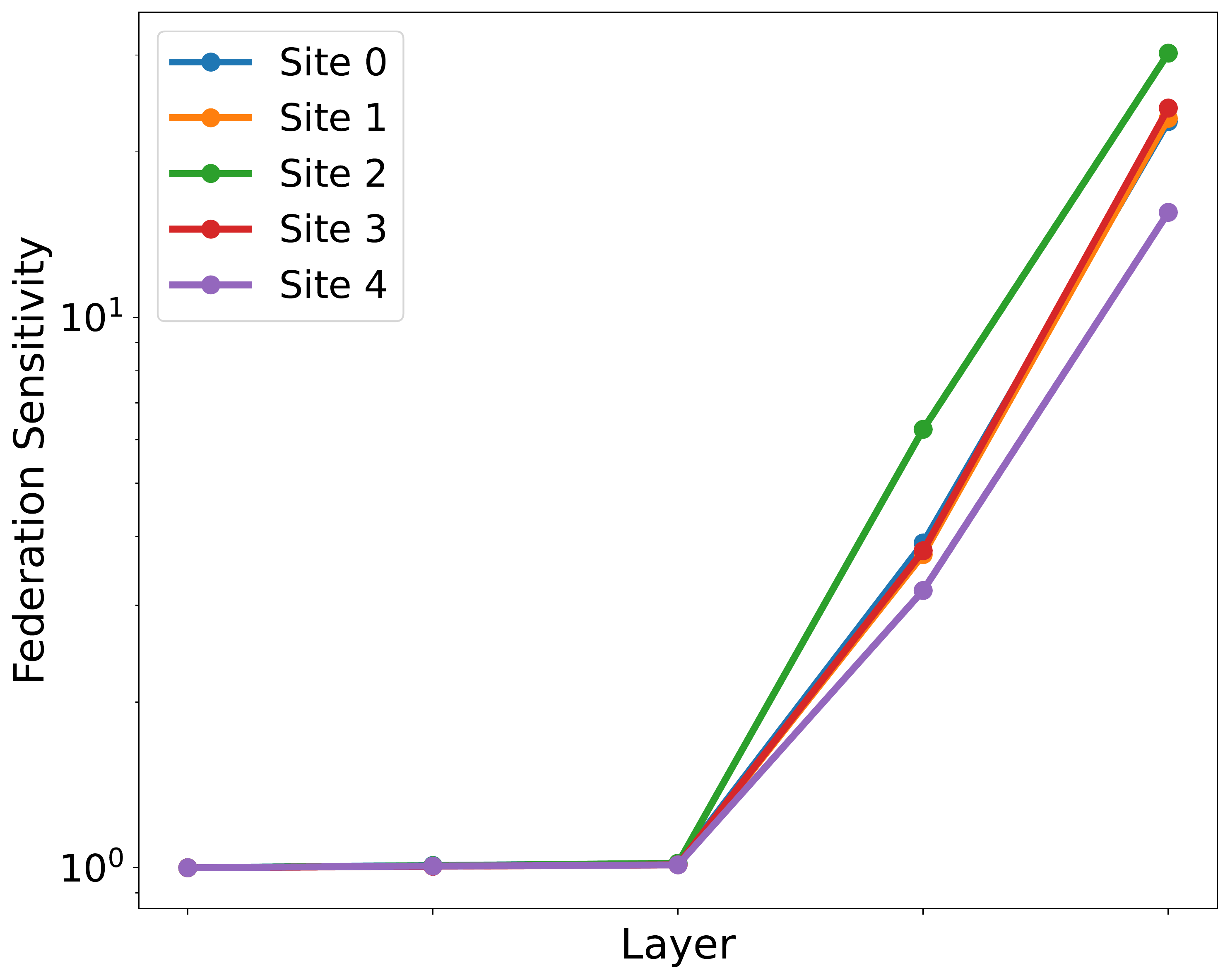}}
  \makebox[\linewidth][c]{\scriptsize\textit{(B) EMNIST}}  % Subfigure label
\end{minipage}%
\begin{minipage}{.24\linewidth}
  \centering
  \raisebox{-\height}{\includegraphics[width=\linewidth]{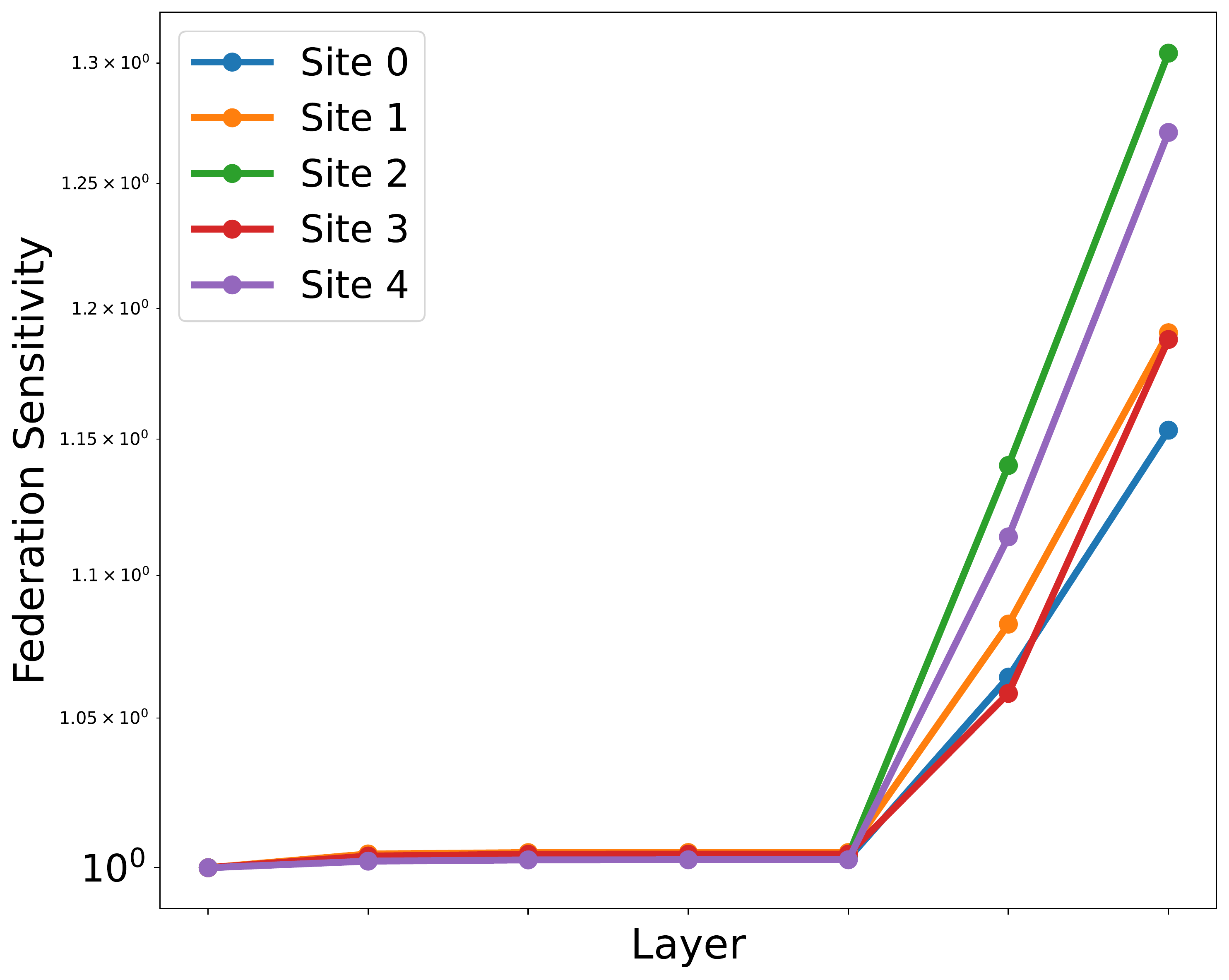}}
  \makebox[\linewidth][c]{\scriptsize\textit{(C) CIFAR-10}}  % Subfigure label
\end{minipage}%
\begin{minipage}{.24\linewidth}
  \centering
  \raisebox{-\height}{\includegraphics[width=\linewidth]{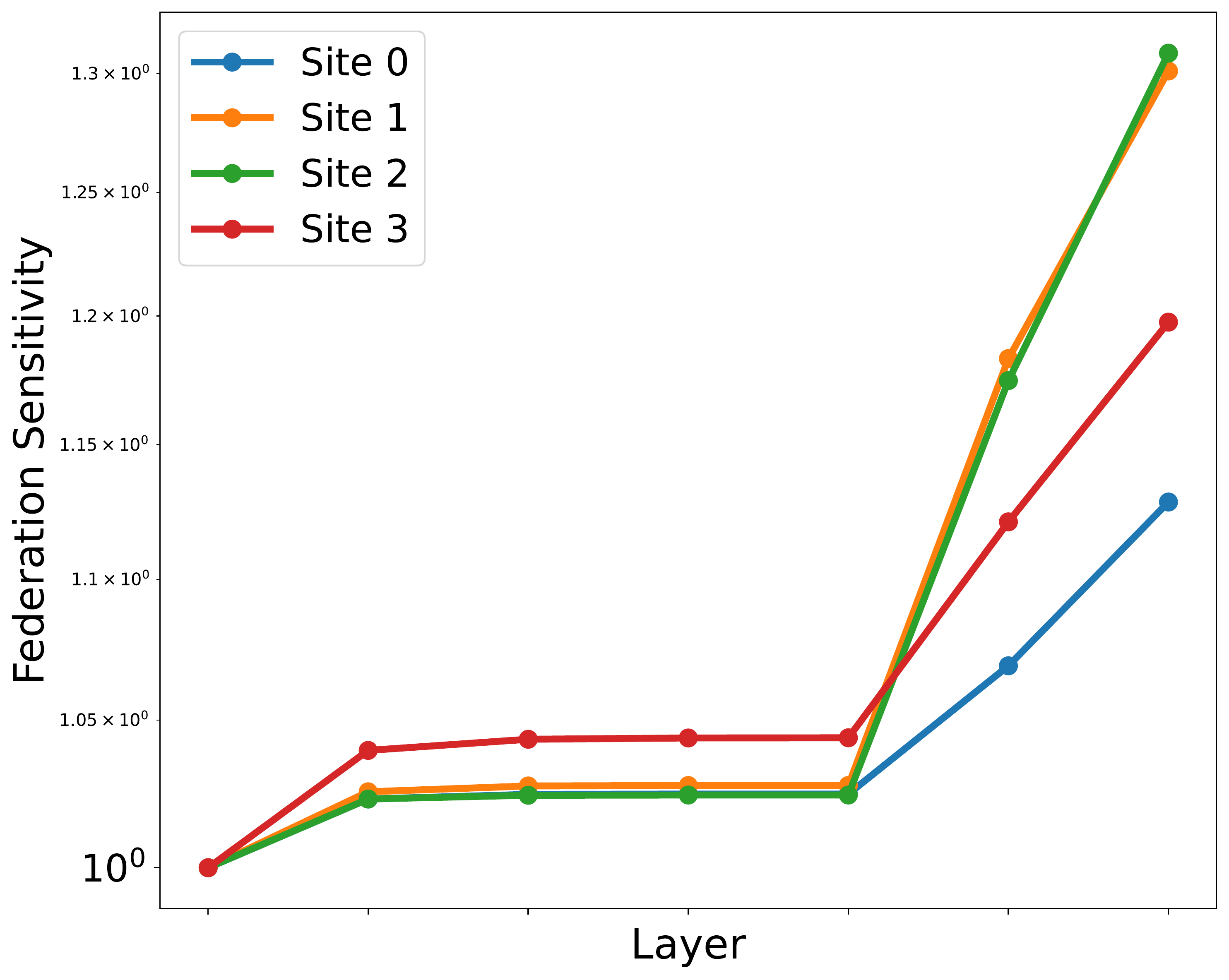}}
  \makebox[\linewidth][c]{\scriptsize\textit{(D) ISIC-2019}}  % Subfigure label
\end{minipage}
}
\makebox[\linewidth][c]{
\begin{minipage}{.24\linewidth}
  \centering
  \raisebox{-\height}{\includegraphics[width=\linewidth]{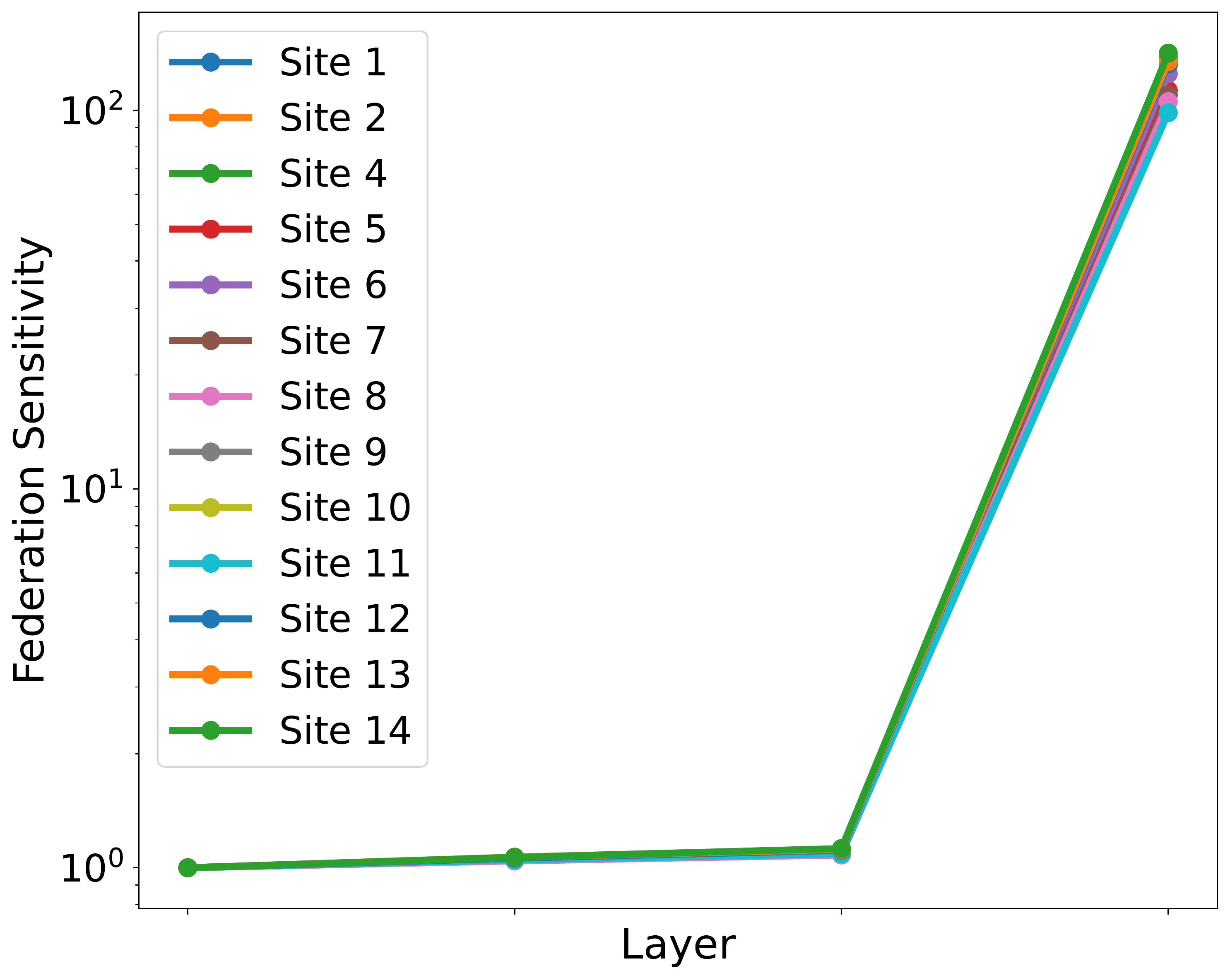}}
  \makebox[\linewidth][c]{\scriptsize\textit{(E) Sent-140}}  % Subfigure label
\end{minipage}%
\begin{minipage}{.24\linewidth}
  \centering
  \raisebox{-\height}{\includegraphics[width=\linewidth]{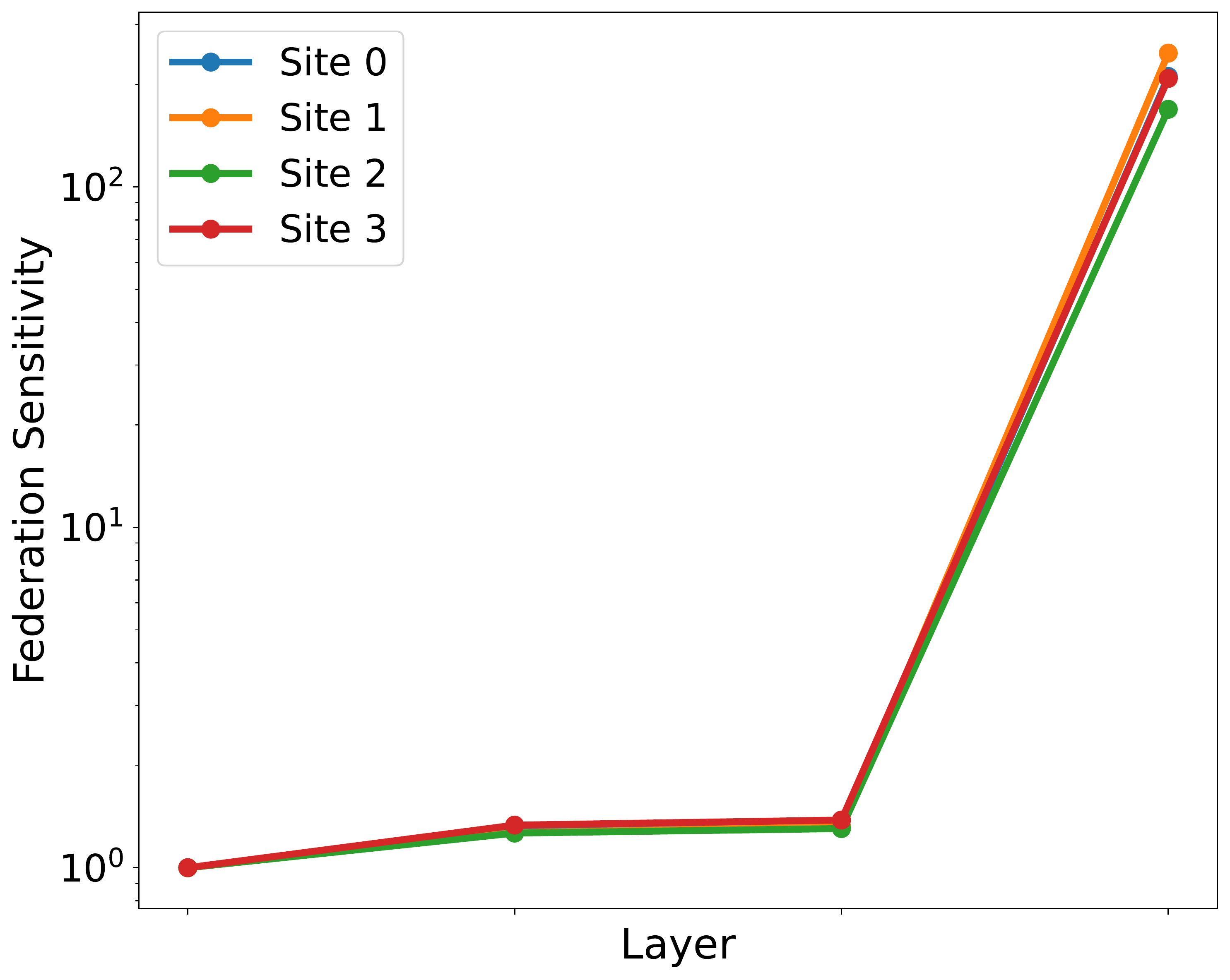}}
  \makebox[\linewidth][c]{\scriptsize\textit{(F) MIMIC-III}}  % Subfigure label
\end{minipage}%
\begin{minipage}{.24\linewidth}
  \centering
  \raisebox{-\height}{\includegraphics[width=\linewidth]{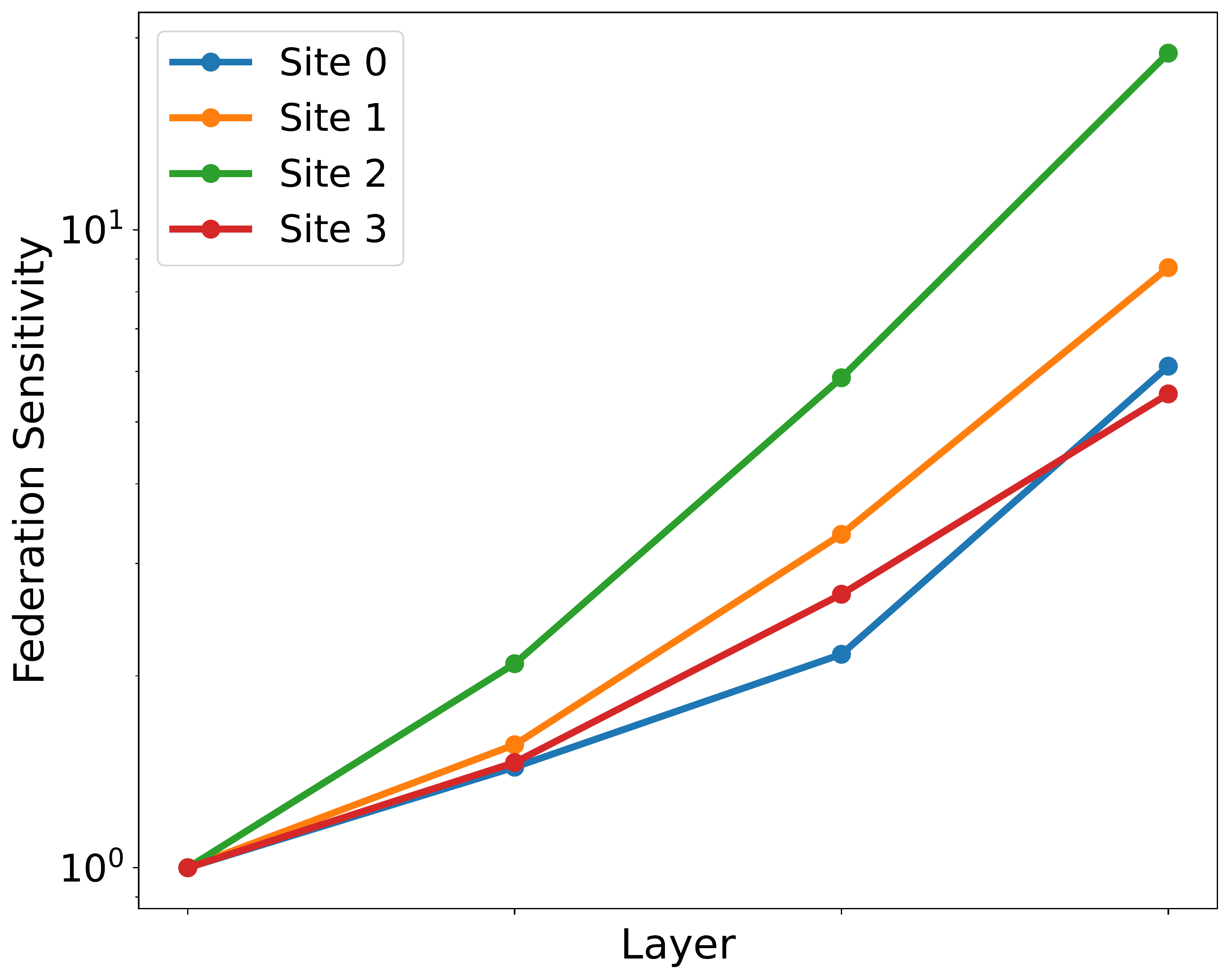}}
  \makebox[\linewidth][c]{\scriptsize\textit{(G) Fed-Heart-Disease}}  % Subfigure label
\end{minipage}
}

\caption{\textbf{Federation sensitivity for final models}. Models trained via FL on non-IID data.}
\label{sfig:layer_imp_best_fl}
\end{center}

\end{figure}

Figures \ref{sfig:layer_imp_best} and \ref{sfig:layer_imp_best_fl} display the federated sensitivity score across all datasets for the final model after independent training, and final model after FL training, respectively. 

\clearpage

\section{Results}
\subsection{F1 score: fairness and incentive}
Table \ref{supp:fairness-table} shows the variance in client results and Table \ref{supp:incentive-table} shows the \% of clients that beat local site or FedAvg training. 
\begin{table}[ht]
\caption{Variance in clients' F1 score (fairness). In \textbf{bold} is fairest model. Friedman rank test p-value $<5\times10^{-3}$}
\label{supp:fairness-table}
\begin{center}
\fontsize{9pt}{7pt}\selectfont
\begin{tabular}{lcccccccc}
\toprule
Algorithm & FMNIST & EMNIST & CIFAR & ISIC & Heart & Sent-140 & MIMIC-III & Rank \\
\midrule
Fedprox & $3.0\cdot 10^{-4}$ & $8.0\cdot 10^{-4}$ & $4.0\cdot 10^{-4}$ & $1.1\cdot 10^{-2}$ & $9.0\cdot 10^{-2}$ & $4.1\cdot 10^{-3}$ & $1.7\cdot 10^{-3}$ & 7.7 \\
pFedMe & $3.0\cdot 10^{-4}$ & $6.0\cdot 10^{-4}$ & $5.0\cdot 10^{-4}$ & $4.2\cdot 10^{-3}$ & $\mathbf{7.1\cdot 10^{-2}}$ & $3.7\cdot 10^{-2}$ & $\mathbf{1.4\cdot 10^{-3}}$ & 4.3 \\
Ditto & $5.0\cdot 10^{-4}$ & $6.0\cdot 10^{-4}$ & $2.0\cdot 10^{-4}$ & $3.6\cdot 10^{-3}$ & $8.2\cdot 10^{-2}$ & $3.6\cdot 10^{-2}$ & $1.6\cdot 10^{-3}$ & 5.0 \\
LocalAdaptation & $\mathbf{1.0\cdot 10^{-4}}$ & $7.0\cdot 10^{-4}$ & $4.0\cdot 10^{-4}$ & $9.0\cdot 10^{-3}$ & $8.3\cdot 10^{-2}$ & $4.1\cdot 10^{-2}$ & $2.2\cdot 10^{-3}$ & 6.4 \\
FedBABU & $2.0\cdot 10^{-4}$ & $4.0\cdot 10^{-4}$ & $4.0\cdot 10^{-4}$ & $2.1\cdot 10^{-3}$ & $8.4\cdot 10^{-2}$ & $3.5\cdot 10^{-2}$ & $2.0\cdot 10^{-3}$ & 4.4 \\
FedLP & $3.0\cdot 10^{-4}$ & $4.0\cdot 10^{-4}$ & $1.0\cdot 10^{-3}$ & $1.1\cdot 10^{-2}$ & $8.5\cdot 10^{-2}$ & $3.7\cdot 10^{-2}$ & $1.6\cdot 10^{-3}$ & 6.1 \\
FedLAMA & $2.0\cdot 10^{-4}$ & $6.0\cdot 10^{-4}$ & $5.0\cdot 10^{-4}$ & $4.7\cdot 10^{-3}$ & $8.7\cdot 10^{-2}$ & $4.1\cdot 10^{-2}$ & $2.0\cdot 10^{-3}$ & 5.6 \\
pFedLA & $4.0\cdot 10^{-4}$ & $1.0\cdot 10^{-4}$ & $\mathbf{1.0\cdot 10^{-4}}$ & $3.5\cdot 10^{-3}$ & $7.4\cdot 10^{-2}$ & $4.1\cdot 10^{-2}$ & $2.1\cdot 10^{-3}$ & 5.1 \\
\midrule
PLayer-FL & $4.0\cdot 10^{-4}$ & $5.0\cdot 10^{-4}$ & $6.0\cdot 10^{-4}$ & $1.3\cdot 10^{-3}$ & $7.3\cdot 10^{-2}$ & $\mathbf{3.3\cdot 10^{-2}}$ & $1.5\cdot 10^{-3}$ & \textbf{3.8} \\
PLayer-FL-Random & $8.0\cdot 10^{-4}$ & $\mathbf{3.0\cdot 10^{-4}}$ & $1.0\cdot 10^{-3}$ & $\mathbf{7.0\cdot 10^{-4}}$ & $7.3\cdot 10^{-2}$ & $3.4\cdot 10^{-2}$ & $1.9\cdot 10^{-3}$ & 6.5 \\
\bottomrule
\end{tabular}
\end{center}
\end{table}
\begin{table}[ht]
\centering
\caption{Incentivized participation rate (\%) using F1 score. In \textbf{bold} is model with highest IPR. Friedman rank test p-value $=0.043$}

\label{supp:incentive-table}
\begin{tabular}{lcccccccc}
\toprule
Algorithm & FMNIST & EMNIST & CIFAR & ISIC & Heart & Sentiment & Mimic-III & Rank \\
\midrule
FedProx & 0.0 & 20.0 & 40.0 & 0.0 & 0.0 & 6.7 & 50.0 & 5.4 \\
pFedMe & 80.0 & 20.0 & 0.0 & 0.0 & \textbf{50.0} & 6.7 & 0.0 & 5.1 \\
Ditto & 60.0 & 0.0 & 0.0 & 0.0 & 0.0 & \textbf{26.7} & 25.0 & 5.6 \\
LocalAdaptation & 0.0 & 40.0 & 40.0 & 0.0 & 0.0 & 0.0 & \textbf{75.0} & 5.2 \\
FedBABU & 0.0 & 40.0 & 80.0 & 0.0 & 0.0 & 20.0 & 50.0 & 4.4 \\
FedLP & 0.0 & \textbf{60.0} & 0.0 & 0.0 & 0.0 & 6.7 & 50.0 & 6.2 \\
FedLAMA & 0.0 & 0.0 & 0.0 & 0.0 & 0.0 & 0.0 & 25.0 & 7.3 \\
pFedLA & 0.0 & 0.0 & 0.0 & 0.0 & 0.0 & 0.0 & 0.0 & 7.7 \\
\midrule
PLayer-FL & \textbf{100.0} & 0.0 & \textbf{100.0} & 0.0 & 25.0 & 20.0 & 50.0 & \textbf{3.4} \\
PLayer-FL-Random & 60.0 & 0.0 & 80.0 & 0.0 & 25.0 & 6.7 & 25.0 & 4.8 \\
\bottomrule
\end{tabular}
\end{table}

\subsection{Accuracy}
\label{supp:accuracy_results}
Table \ref{supp:acc-table} shows clients' median accuracy scores (95\% confidence interval), Table \ref{supp:fairness_accuracy} shows the variance in client accuracy scores and Table \ref{supp:incentive_accuracy} shows the \% of clients that beat single or FedAvg training in accuracy. 

\begin{table*}[ht]
\caption{Accuracy and average rank. In \textbf{bold} is the top-performing model. Friedman rank test p-value $<5\times10^{-3}$}
\label{supp:acc-table}
\begin{center}
\setlength{\tabcolsep}{5.4pt}
\begin{tabular}{lcccccccc}
\toprule
Algorithm & FMNIST & EMNIST & CIFAR & ISIC & Heart & Sentiment & mimic & Avg Rank \\
\midrule
Local & \specialcell{88.7}{0.9} & \specialcell{84.2}{0.9} & \specialcell{79.4}{1.3} & \textbf{\specialcell{68.9}{1.8}} & \specialcell{53.7}{1.0} & \specialcell{79.8}{0.3} & \specialcell{74.7}{2.1} & 6.7 \\
FedAvg & \specialcell{89.1}{0.5} & \textbf{\specialcell{86.5}{0.6}} & \specialcell{80.0}{0.6} & \specialcell{66.4}{1.2} & \specialcell{54.1}{0.7} & \specialcell{78.8}{0.2} & \specialcell{74.7}{0.6} & 6.4 \\
FedProx & \specialcell{89.2}{0.4} & \specialcell{86.2}{0.9} & \specialcell{80.2}{0.6} & \specialcell{66.5}{1.6} & \specialcell{52.9}{1.6} & \specialcell{78.8}{0.2} & \textbf{\specialcell{75.7}{1.7}} & 6.4 \\
pFedMe & \specialcell{88.2}{0.8} & \specialcell{85.7}{0.9} & \specialcell{67.4}{1.1} & \specialcell{66.8}{0.6} & \specialcell{55.0}{1.0} & \specialcell{80.0}{0.2} & \specialcell{73.3}{0.9} & 6.9 \\
Ditto & \specialcell{89.0}{0.6} & \specialcell{85.6}{1.1} & \specialcell{67.8}{0.8} & \specialcell{66.0}{1.0} & \textbf{\specialcell{55.5}{1.0}} & \specialcell{80.3}{0.6} & \specialcell{73.5}{0.9} & 7.0 \\
LocalAdaptation & \specialcell{89.3}{0.8} & \specialcell{86.3}{0.6} & \specialcell{80.0}{1.1} & \specialcell{66.9}{0.2} & \specialcell{53.8}{0.8} & \specialcell{78.9}{0.3} & \specialcell{74.0}{2.2} & 6.3 \\
FedBABU & \specialcell{89.2}{0.6} & \specialcell{86.4}{0.7} & \specialcell{80.7}{1.4} & \specialcell{67.3}{1.6} & \specialcell{54.2}{0.7} & \textbf{\specialcell{81.0}{0.2}} & \specialcell{74.9}{0.6} & 3.4 \\
FedLP & \specialcell{89.1}{0.6} & \specialcell{86.3}{0.9} & \specialcell{78.6}{1.6} & \specialcell{66.6}{0.7} & \specialcell{54.1}{0.7} & \specialcell{79.4}{0.3} & \specialcell{74.6}{1.3} & 7.0 \\
FedLama & \specialcell{86.3}{0.8} & \specialcell{83.3}{0.6} & \specialcell{63.4}{2.0} & \specialcell{62.4}{1.5} & \specialcell{54.7}{0.8} & \specialcell{78.0}{0.1} & \specialcell{75.2}{1.2} & 8.9 \\
pFedLA & \specialcell{70.8}{0.2} & \specialcell{41.2}{2.7} & \specialcell{37.1}{2.3} & \specialcell{56.7}{1.6} & \specialcell{52.6}{1.4} & \specialcell{78.0}{0.0} & \specialcell{74.7}{2.6} & 11.1 \\
\midrule
PLayer-FL & \textbf{\specialcell{89.8}{0.7}} & \specialcell{86.1}{0.8} & \textbf{\specialcell{81.5}{1.1}} & \specialcell{68.6}{0.8} & \specialcell{55.4}{1.2} & \specialcell{80.7}{0.7} & \textbf{\specialcell{75.7}{0.6}} & \textbf{2.2} \\
PLayer-FL-Random & \specialcell{89.2}{0.7} & \specialcell{84.1}{0.8} & \specialcell{81.2}{1.2} & \specialcell{68.6}{0.8} & \specialcell{54.4}{1.2} & \specialcell{79.0}{0.4} & \specialcell{74.5}{2.0} & 5.8 \\
\bottomrule
\bottomrule
\end{tabular}
\end{center}
\end{table*}

\begin{table}[ht]
\caption{Variance in clients' accuracy (fairness). In \textbf{bold} is the fairest model. Friedman rank test p-value $<5\times10^{-3}$}
\label{supp:fairness_accuracy}

\begin{center}
\begin{small}
\begin{tabular}{lcccccccc}
\toprule
Algorithm & FMNIST & EMNIST & CIFAR & ISIC & Heart & Sentiment & mimic & Avg Rank \\
\midrule
FedProx & $1.0\cdot 10^{-4}$ & $3.0\cdot 10^{-4}$ & $1.0\cdot 10^{-4}$ & $3.5\cdot 10^{-2}$ & $6.7\cdot 10^{-2}$ & $2.4\cdot 10^{-2}$ & $1.1\cdot 10^{-2}$ & 5.7 \\
pFedMe & $\mathbf{1.0\cdot 10^{-5}}$ & $3.0\cdot 10^{-4}$ & $4.0\cdot 10^{-4}$ & $3.3\cdot 10^{-2}$ & $4.6\cdot 10^{-2}$ & $2.1\cdot 10^{-2}$ & $1.4\cdot 10^{-2}$ & 5.0 \\
ditto & $1.0\cdot 10^{-4}$ & $2.0\cdot 10^{-4}$ & $3.0\cdot 10^{-4}$ & $3.5\cdot 10^{-2}$ & $5.7\cdot 10^{-2}$ & $2.3\cdot 10^{-2}$ & $1.3\cdot 10^{-2}$ & 5.3 \\
LocalAdaptation & $1.0\cdot 10^{-4}$ & $2.0\cdot 10^{-4}$ & $2.0\cdot 10^{-4}$ & $3.2\cdot 10^{-2}$ & $5.7\cdot 10^{-2}$ & $2.4\cdot 10^{-2}$ & $1.4\cdot 10^{-2}$ & 5.5 \\
FedBABU & $1.0\cdot 10^{-4}$ & $2.0\cdot 10^{-4}$ & $1.0\cdot 10^{-4}$ & $3.3\cdot 10^{-2}$ & $6.0\cdot 10^{-2}$ & $\mathbf{1.8\cdot 10^{-2}}$ & $1.2\cdot 10^{-2}$ & 4.9 \\
FedLP & $1.0\cdot 10^{-4}$ & $2.0\cdot 10^{-4}$ & $1.0\cdot 10^{-4}$ & $3.6\cdot 10^{-2}$ & $5.7\cdot 10^{-2}$ & $2.2\cdot 10^{-2}$ & $1.5\cdot 10^{-2}$ & 5.3 \\
FedLama & $1.0\cdot 10^{-4}$ & $2.0\cdot 10^{-4}$ & $3.0\cdot 10^{-4}$ & $4.2\cdot 10^{-3}$ & $6.3\cdot 10^{-2}$ & $2.6\cdot 10^{-2}$ & $1.3\cdot 10^{-2}$ & 7.1 \\
pFedLA & $9.0\cdot 10^{-4}$ & $6.0\cdot 10^{-4}$ & $2.3\cdot 10^{-3}$ & $6.3\cdot 10^{-2}$ & $5.1\cdot 10^{-2}$ & $2.6\cdot 10^{-2}$ & $2.5\cdot 10^{-2}$ & 8.9 \\
\midrule
PLayer-FL & $1.0\cdot 10^{-4}$ & $\mathbf{1.0\cdot 10^{-4}}$ & $1.0\cdot 10^{-4}$ & $\mathbf{2.9\cdot 10^{-2}}$ & $4.7\cdot 10^{-2}$ & $1.9\cdot 10^{-2}$ & $1.3\cdot 10^{-2}$ & \textbf{2.5} \\
PLayer-FL-Random & $1.0\cdot 10^{-4}$ & $3.0\cdot 10^{-4}$ & $\mathbf{1.0\cdot 10^{-4}}$ & $\mathbf{2.9\cdot 10^{-2}}$ & $5.7\cdot 10^{-2}$ & $2.5\cdot 10^{-2}$ & $\mathbf{9.7\cdot 10^{-3}}$ & 4.9 \\
\bottomrule
\end{tabular}
\end{small}
\end{center}

\end{table}

\begin{table}[ht]
\centering
\caption{Incentivized Participation Rate using accuracy (\%). In \textbf{bold} is model with highest IPR. Friedman rank test p-value $<5\times10^{-3}$}
\label{supp:incentive_accuracy}
\begin{tabular}{lcccccccc}
\toprule
Algorithm & FMNIST & EMNIST & CIFAR & ISIC & Heart & Sent-140 & Mimic-III & Avg Rank \\ 
\midrule
FedProx & 20.0 & \textbf{40.0} & 40.0 & 0.0 & 0.0 & 0.0 & 50.0 & 5.6 \\
pFedMe & 40.0 & \textbf{40.0} & 0.0 & 0.0 & \textbf{25.0} & 6.7 & 0.0 & 5.6 \\
Ditto & 60.0 & 0.0 & 0.0 & 0.0 & 0.0 & 6.7 & 0.0 & 6.8 \\
LocalAdaptation & 60.0 & 40.0 & 60.0 & 0.0 & 0.0 & 0.0 & 25.0 & 5.3 \\
FedBABU & 80.0 & 20.0 & 60.0 & 0.0 & 0.0 & \textbf{20.0} & \textbf{25.0} & 4.3 \\
FedLP & 60.0 & 20.0 & 20.0 & 0.0 & 0.0 & 6.7 & \textbf{25.0} & 5.4 \\
FedLama & 0.0 & 0.0 & 0.0 & 0.0 & 0.0 & 0.0 & 0.0 & 8.1 \\
pFedLA & 0.0 & 0.0 & 0.0 & 0.0 & 25.0 & 0.0 & 50.0 & 6.5 \\
\midrule
PLayer-FL & \textbf{80.0} & 0.0 & \textbf{100.0} & \textbf{25.0} & \textbf{25.0} & \textbf{20.0} & \textbf{75.0} & \textbf{2.4} \\
PLayer-FL-Random & 40.0 & 0.0 & \textbf{100.0} & \textbf{25.0} & 0.0 & 6.7 & 25.0 & 4.9 \\
\bottomrule
\end{tabular}
\end{table}

\subsection{Loss}
\label{supp:loss_results}
Table \ref{supp:loss-table} shows clients' median test loss (95\% confidence interval), Table \ref{supp:fairness_loss} shows the variance in client test loss and Table \ref{supp:incentive_loss} shows the \% of clients that beat single or FedAvg training in test loss. 

\begin{table*}[ht]
\caption{Test loss ($\times 1\cdot10^{-4}$) and average rank. In \textbf{bold} is the top-performing model. Friedman rank test p-value $<5\times10^{-3}$}
\label{supp:loss-table}
\begin{center}
\setlength{\tabcolsep}{4.5pt}
\begin{tabular}{lcccccccc}
\toprule
Algorithm & FMNIST & EMNIST & CIFAR & ISIC & Heart & Sentiment & mimic & Avg Rank \\
\midrule
Local client & \specialcell{34.7}{2.0} & \specialcell{58.6}{3.7} & \specialcell{60.7}{4.3} & \textbf{\specialcell{111.2}{0.3}} & \specialcell{29.6}{0.8} & \specialcell{43.7}{0.3} & \specialcell{71.4}{0.8} & 8.3 \\
FedAvg & \specialcell{30.7}{1.1} & \specialcell{42.2}{2.8} & \specialcell{57.4}{3.1} & \specialcell{121.8}{1.6} & \specialcell{27.5}{0.5} & \specialcell{44.1}{0.2} & \specialcell{65.8}{0.9} & 6.1 \\
FedProx & \specialcell{31.5}{1.9} & \specialcell{41.6}{2.9} & \specialcell{56.3}{1.3} & \specialcell{121.6}{1.1} & \specialcell{28.4}{0.9} & \specialcell{43.9}{0.2} & \specialcell{65.8}{1.1} & 6.3 \\
pFedMe & \specialcell{30.4}{2.3} & \specialcell{42.9}{2.7} & \specialcell{90.9}{2.8} & \specialcell{118.0}{1.0} & \specialcell{26.4}{0.9} & \specialcell{42.2}{0.6} & \specialcell{67.4}{0.8} & 5.1 \\
Ditto & \specialcell{30.8}{0.2} & \specialcell{45.0}{3.0} & \specialcell{89.9}{3.9} & \specialcell{120.6}{2.5} & \textbf{\specialcell{26.2}{0.4}} & \specialcell{42.5}{0.6} & \specialcell{67.3}{0.1} & 5.9 \\
LocalAdaptation & \specialcell{30.5}{1.9} & \textbf{\specialcell{41.1}{3.4}} & \specialcell{57.0}{3.0} & \specialcell{121.4}{1.3} & \specialcell{28.1}{0.6} & \specialcell{43.6}{0.1} & \specialcell{65.7}{0.6} & 4.9 \\
FedBABU & \specialcell{30.0}{1.9} & \specialcell{42.2}{3.5} & \specialcell{55.2}{2.8} & \specialcell{120.6}{1.5} & \specialcell{29.1}{1.0} & \specialcell{41.8}{0.2} & \specialcell{65.5}{0.7} & 3.8 \\
FedLP & \specialcell{30.1}{1.0} & \specialcell{43.1}{3.2} & \specialcell{61.4}{3.6} & \specialcell{128.7}{1.5} & \specialcell{27.3}{0.5} & \specialcell{42.9}{0.2} & \specialcell{65.8}{0.6} & 6.7 \\
FedLAMA & \specialcell{37.6}{2.0} & \specialcell{53.2}{2.3} & \specialcell{102.4}{3.0} & \specialcell{133.0}{1.8} & \specialcell{27.2}{0.5} & \specialcell{45.0}{0.3} & \textbf{\specialcell{65.2}{0.5}} & 8.3 \\
pFedLA & \specialcell{88.6}{6.9} & \specialcell{253.4}{11.2} & \specialcell{176.2}{10.6} & \specialcell{151.2}{1.8} & \specialcell{36.1}{4.9} & \specialcell{47.2}{0.6} & \specialcell{75.4}{0.6} & 12.0 \\
\midrule
PLayer-FL & \textbf{\specialcell{27.7}{2.0}} & \specialcell{45.4}{3.9} & \textbf{\specialcell{52.6}{2.6}} & \specialcell{113.4}{0.6} & \specialcell{26.7}{0.6} & \textbf{\specialcell{41.4}{0.4}} & \specialcell{66.2}{0.7} & \textbf{3.4} \\
PLayer-FL-Random & \specialcell{33.3}{2.3} & \specialcell{58.2}{1.8} & \specialcell{55.6}{3.7} & \specialcell{113.1}{1.6} & \specialcell{27.9}{1.0} & \specialcell{44.6}{0.7} & \specialcell{69.9}{1.2} & 7.4 \\
\bottomrule
\end{tabular}
\end{center}
\end{table*}

\begin{table}[ht]
\caption{Variance in clients' test loss (fairness). In \textbf{bold} is the fairest model. Friedman rank test p-value $<5\times10^{-3}$}
\label{supp:fairness_loss}
\begin{center}
\begin{small}
\begin{tabular}{lcccccccc}
\toprule
Algorithm & FMNIST & EMNIST & CIFAR & ISIC & Heart & Sent-140 & MIMIC-III & Avg Rank \\ 
\midrule
FedProx & $1.1\cdot 10^{-3}$ & $1.7\cdot 10^{-3}$ & $1.1\cdot 10^{-3}$ & $3.8\cdot 10^{-1}$ & $1.8\cdot 10^{-2}$ & $5.5\cdot 10^{-2}$ & $4.7\cdot 10^{-2}$ & 6.4 \\
pFedMe & $5.0\cdot 10^{-4}$ & $2.3\cdot 10^{-3}$ & $1.8\cdot 10^{-3}$ & $3.6\cdot 10^{-1}$ & $1.3\cdot 10^{-2}$ & $5.2\cdot 10^{-2}$ & $3.5\cdot 10^{-2}$ & 4.1 \\
Ditto & $6.0\cdot 10^{-4}$ & $2.2\cdot 10^{-3}$ & $2.3\cdot 10^{-4}$ & $3.8\cdot 10^{-1}$ & $1.7\cdot 10^{-2}$ & $5.2\cdot 10^{-2}$ & $3.8\cdot 10^{-2}$ & 4.6 \\
LocalAdaptation & $1.0\cdot 10^{-3}$ & $2.1\cdot 10^{-3}$ & $1.2\cdot 10^{-3}$ & $4.0\cdot 10^{-1}$ & $1.9\cdot 10^{-2}$ & $5.5\cdot 10^{-2}$ & $4.8\cdot 10^{-2}$ & 7.1 \\
FedBABU & $4.0\cdot 10^{-4}$ & $\mathbf{1.2\cdot 10^{-3}}$ & $7.0\cdot 10^{-4}$ & $4.1\cdot 10^{-1}$ & $2.1\cdot 10^{-2}$ & $\mathbf{4.6\cdot 10^{-2}}$ & $4.5\cdot 10^{-2}$ & 4.4 \\
FedLP & $7.0\cdot 10^{-4}$ & $1.9\cdot 10^{-3}$ & $8.0\cdot 10^{-4}$ & $4.3\cdot 10^{-1}$ & $1.7\cdot 10^{-2}$ & $5.2\cdot 10^{-2}$ & $4.3\cdot 10^{-2}$ & 5.1 \\
FedLAMA & $\mathbf{2.0\cdot 10^{-4}}$ & $2.9\cdot 10^{-3}$ & $2.4\cdot 10^{-3}$ & $4.6\cdot 10^{-1}$ & $1.8\cdot 10^{-2}$ & $5.3\cdot 10^{-2}$ & $4.3\cdot 10^{-2}$ & 6.7 \\
pFedLA & $1.0\cdot 10^{-3}$ & $2.3\cdot 10^{-2}$ & $5.1\cdot 10^{-2}$ & $5.7\cdot 10^{-1}$ & $2.6\cdot 10^{-1}$ & $5.1\cdot 10^{-2}$ & $\mathbf{3.2\cdot 10^{-2}}$ & 7.6 \\
\midrule
PLayer-FL & $7.0\cdot 10^{-4}$ & $2.5\cdot 10^{-3}$ & $4.0\cdot 10^{-4}$ & $\mathbf{3.2\cdot 10^{-1}}$ & $\mathbf{1.4\cdot 10^{-2}}$ & $5.1\cdot 10^{-2}$ & $4.2\cdot 10^{-2}$ & \textbf{3.6} \\
PLayer-FL-Random & $8.0\cdot 10^{-4}$ & $4.1\cdot 10^{-3}$ & $\mathbf{1.0\cdot 10^{-4}}$ & $\mathbf{3.2\cdot 10^{-1}}$ & $1.6\cdot 10^{-2}$ & $6.1\cdot 10^{-2}$ & $4.0\cdot 10^{-2}$ & 5.2 \\
\bottomrule
\end{tabular}
\end{small}
\end{center}
\end{table}

\begin{table}[ht]
\centering
\caption{Incentivized Participation Rate using test loss (\%) and Average Algorithm Rank. Friedman rank test p-value $=0.084$}
\label{supp:incentive_loss}
\begin{tabular}{lcccccccc}
\toprule
Algorithm & FMNIST & EMNIST & CIFAR & ISIC & Heart & Sent-140 & Mimic-III & Avg Rank \\ 
\midrule
FedProx & 40.0 & \textbf{80.0} & 40.0 & 0.0 & 25.0 & 33.3 & 25.0 & 5.1 \\
pFedMe & 40.0 & 40.0 & 0.0 & 0.0 & \textbf{100.0} & 33.3 & 0.0 & 5.7 \\
Ditto & 40.0 & 0.0 & 0.0 & 0.0 & 75.0 & 33.3 & 0.0 & 6.5 \\
LocalAdaptation & 20.0 & 60.0 & 80.0 & 0.0 & 25.0 & 40.0 & 25.0 & 4.8 \\
FedBABU & 80.0 & 40.0 & 60.0 & 0.0 & 0.0 & 40.0 & 25.0 & 4.6 \\
FedLP & 60.0 & \textbf{80.0} & 20.0 & 0.0 & 75.0 & 40.0 & 50.0 & 3.6 \\
FedLAMA & 0.0 & 0.0 & 0.0 & 0.0 & 50.0 & 20.0 &  \textbf{75.0} & 6.7 \\
pFedLA & 0.0 & 0.0 & 0.0 & 0.0 & 0.0 & 6.7 & 0.0 & 8.5 \\
\midrule
PLayer-FL & \textbf{80.0} & 0.0 & \textbf{100.0} & \textbf{50.0} & \textbf{100.0} & 40.0 & 0.0 & \textbf{3.5} \\
PLayer-FL-Random & 0.0 & 0.0 & \textbf{100.0} & \textbf{50.0} & 50.0 & 20.0 & 0.0 & 6.0 \\
\bottomrule
\end{tabular}
\end{table}

\end{document}